\documentclass[10pt,journal,compsoc]{IEEEtran}
\ifCLASSOPTIONcompsoc
  \usepackage[nocompress]{cite}
\else
  \usepackage{cite}
\fi
\ifCLASSINFOpdf
\else
\fi

\usepackage{mathptmx}
\usepackage{graphicx}
\usepackage{times}
\usepackage{caption}
\usepackage{morefloats}
\usepackage[tight,normalsize,sf,SF]{subfigure}

\usepackage{amsmath,amscd}
\usepackage{amsfonts}
\usepackage[ruled,vlined]{algorithm2e}
\usepackage{algorithmic}
\usepackage{epsfig}
\usepackage{enumerate}
\usepackage{multirow}
\usepackage{array}
\usepackage{diagrams}
\usepackage{diagbox}

\usepackage{color,soul} 

\makeatletter
 \def\SOUL@hlpreamble{%
 \setul{}{2.4ex}
 \let\SOUL@stcolor\SOUL@hlcolor
 \SOUL@stpreamble
 }
\makeatother

\begin{document}
%
\title{Radiative Transport Based Flame Volume Reconstruction from Videos}
%
%
%
%

\author{Liang~Shen,~Dengming~Zhu,~Saad~Nadeem,~Zhaoqi~Wang,~and~Arie~E.~Kaufman,~\IEEEmembership{Fellow,~IEEE}
\IEEEcompsocitemizethanks{\IEEEcompsocthanksitem L. Shen is with the University of Chinese Academy of Sciences, China.
\protect\\
E-mail: shenliang@ict.ac.cn.
\IEEEcompsocthanksitem S. Nadeem and A.E. Kaufman are with the Department of Computer Science, Stony Brook University, Stony Brook, NY 11794.
\protect\\
E-mail: \{sanadeem, ari\}@cs.stonybrook.edu.
\IEEEcompsocthanksitem L. Shen, D. Zhu and Z. Wang are with the Institute of Computing Technology, Chinese Academy of Sciences, China.
\protect\\
E-mail: \{shenliang, mdzhu, zqwang\}@ict.ac.cn.}
\thanks{}}
\IEEEtitleabstractindextext{%
\begin{abstract}
We introduce a novel approach for flame volume reconstruction from videos using inexpensive charge-coupled device (CCD) consumer cameras. The approach includes an economical data capture technique using inexpensive CCD cameras. Leveraging the smear feature of the CCD chip, we present a technique for synchronizing CCD cameras while capturing flame videos from different views. Our reconstruction is based on the radiative transport equation which enables complex phenomena such as emission, extinction, and scattering to be used in the rendering process. Both the color intensity and temperature reconstructions are implemented using the CUDA parallel computing framework, which provides real-time performance and allows visualization of reconstruction results after every iteration. We present the results of our approach using real captured data and physically-based simulated data. Finally, we also compare our approach against the other state-of-the-art flame volume reconstruction methods and demonstrate the efficacy and efficiency of our approach in four different applications: (1) rendering of reconstructed flames in virtual environments, (2) rendering of reconstructed flames in augmented reality, (3) flame stylization, and (4) reconstruction of other semitransparent phenomena.
\end{abstract}

}

\maketitle

\IEEEdisplaynontitleabstractindextext

%
\IEEEpeerreviewmaketitle

\section{Introduction}

Capturing and reconstructing fluids has been the subject of considerable research in computer graphics.
Several capture-based methods have been developed for the reconstruction of
flames \cite{hasinoff2007photo}, \cite{ihrke2004image}, \cite{okabe2015fluid}, \cite{wu2014reconstruction},
gases \cite{atcheson2008time}, water surface \cite{li2013water},
and mixing fluids \cite{gregson2012stochastic}.
Information extracted from the reconstruction is valuable for direct re-rendering,
developing data-driven models, and combining with physically-based simulation methods to obtain improved results for a variety of applications \cite{okabe2015fluid}, \cite{gregson2014capture}, 
such as re-simulation and detail enhancement.
More importantly, we can deepen the understanding of the principle underlying specific fluid behavior
through the captured data and model this real-world phenomenon more accurately.
In this paper, we focus on flame volume reconstruction from captured videos.

\begin{figure*}[htp]
\centering
\includegraphics[width=\textwidth]{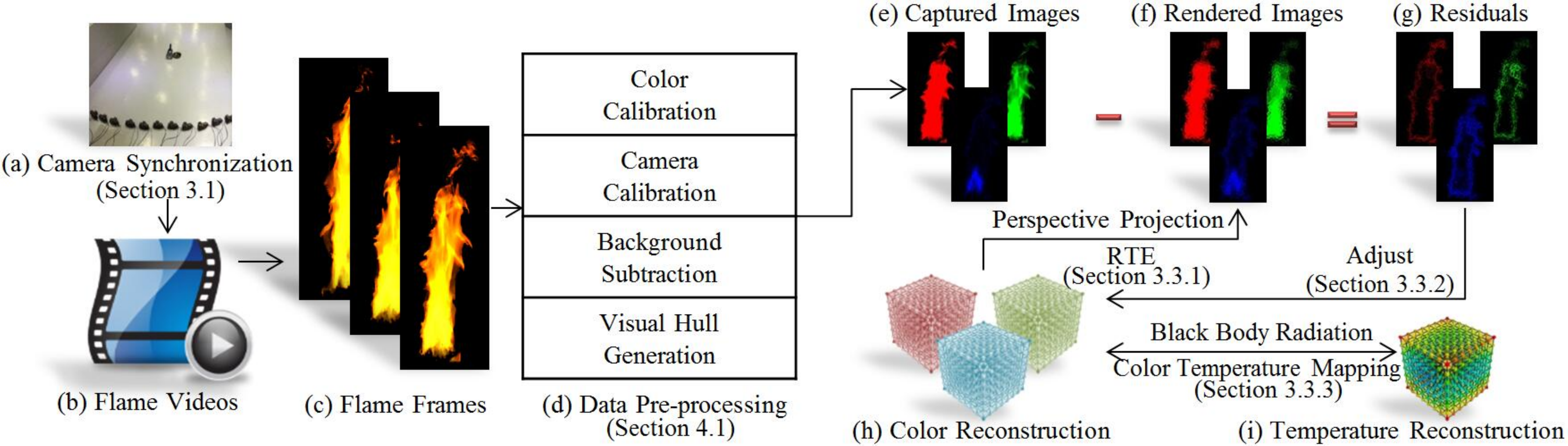}
\vspace{-5mm}
\caption{The pipeline for volume reconstruction of flames from videos. With a smear-based CCD camera synchronization method (a),
flame videos (b) are captured and then converted to frame sequences (c).
For each frame, after data pre-processing (d), the captured images (e) from different views are compared to the rendered images (f),
which are the rendering results of the color volume data (h) using perspective projection and RTE.
The residuals between the captured and rendered images are used to adjust the volume data (h).
The comparison and adjustment are carried out iteratively to finish the color reconstruction.
For the temperature reconstruction (i), we use black-body radiation to build the mapping between the color and the temperature.
Specifically, the temperature reconstruction is directly related to the green channel reconstruction,
as the temperature values are mapped to the green values in our method.
\label{fig_pipeline}}
\vspace{-3mm}
\end{figure*}

Compared to other fluids, flames are more turbulent and noisy, and hence, high-end professional cameras were traditionally required to capture this rapidly changing phenomenon. The cost of this high-end equipment has previously been the bottleneck in the accurate volume reconstruction of flame data. With the rapid development of couple-charged device (CCD) consumer cameras, it is now possible to capture flame data using these relatively inexpensive devices. However, it is challenging to synchronize these CCD consumer cameras for accurate flame volume reconstruction. In this work, we introduce a novel method using a stroboscope to synchronize the CCD cameras (Fig. \ref{fig_pipeline}(a)) and obtain a good reconstruction from our captured flame data.

Previous solutions for flame volume reconstruction have used simplified linear rendering models with parallel projection. The use of a linear rendering model ignores the complex phenomena such as extinction and scattering during the reconstruction process. Parallel projection, on the other hand, is not how our eyes or photography normally work, and hence, this crude approximation leads to considerable errors in the final reconstruction.

In contrast, we use perspective projection and a more accurate rendering formation model based on the radiative transport equation (RTE) \cite{siegel2001thermal}, which directly incorporates emission, extinction, and scattering in our reconstruction process. More specifically, we use perspective projection and RTE (Fig. \ref{fig_pipeline}(f)--(h)) to render different views for our flame volume data. These rendered views are iteratively compared against the synchronized captured frames (Fig. \ref{fig_pipeline}(e)) and refined to adjust the volume data until a certain threshold is reached.

The final flame volume data incorporates both the color intensity and the black-body radiation based temperature information from the captured video data. In essence, the rendered views are used to assign color intensities to the voxels in the volume data.
Simultaneously, we use the black-body radiation model to build the color-temperature mapping and, consequently, to reconstruct the temperature values from the color intensities (Fig. \ref{fig_pipeline}(i)).
In this paper, we focus on providing an accurate visual rendering of our reconstruction in various applications, and thus, the reconstruction of the true temperature will be the focus of future work (as discussed in Section~\ref{sec_conclusion}).

In the implementation, we accelerate our volume reconstruction method using the Compute Unified Device Architecture (CUDA) parallel computing framework to achieve real-time performance, even with the use of the complex rendering model. We evaluate our method with simulated and real captured data, and demonstrate the efficiency and efficacy of our flame volume reconstruction in four different applications.

The contributions of this paper are as follows:
\begin{itemize}
  \item A novel flame volume reconstruction method based on the radiative transport equation which allows modeling of complex phenomena such as emission, extinction, and scattering. Moreover, our method is GPU-accelerated and provides real-time performance.
  \item A consumer CCD camera synchronization technique.
  \item Four applications of our work: (1) rendering reconstructed flames in virtual environments, (2) rendering reconstructed flames in augmented reality, (3) flame stylization, and (4) reconstruction of other phenomena.
\end{itemize}

The paper is organized as follows. We discuss related works in the next section, followed by the details of our algorithm in Section~\ref{sec_algorithm} and its implementation in Section~\ref{sec_implementation}. We evaluate our algorithm and show additional results in Section~\ref{sec_results}. Then, we demonstrate the efficacy of our approach in four different applications in Section~\ref{sec_applications}. Finally, we conclude with the current limitations of the proposed method and the avenues for future work in Section~\ref{sec_conclusion}.


\section{Related Work}
\label{sec_related_work}

The reconstruction and modeling of flames has drawn significant attention in the fields of computer graphics and visualization. Various techniques have been proposed to solve this problem.


\subsection{Computed Tomography}
Computed tomography is widely used in medical imaging and requires thousands of input views for classic methods, such as filtered back-projection and algebraic techniques
\cite{andersen1984simultaneous}, \cite{ehlke2013fast}, \cite{fiddy1985radon}, \cite{natterer1986mathematics}, \cite{ramm1996radon}.
However, for common reconstruction problems, it is difficult to obtain a large number of input views and set up strictly parallel positions for the cameras.
For our flame reconstruction, we focus on the sparse-view tomography problem for semitransparent phenomena which uses limited views as input.
Some reconstruction methods have been proposed from this perspective, including the wavelet graph model \cite{frese2002adaptive}, 
time-resolved Schlieren tomography for gas flows \cite{atcheson2008time}, and stochastic tomography for mixing fluids \cite{gregson2012stochastic}.

Our work is most closely related to the stochastic tomography method for mixing fluids \cite{gregson2012stochastic},
which dynamically changes the samples in the reconstructed volume based on a random walk algorithm.
However, unlike stochastic tomography, our method is based on the radiative transfer equation, which describes the
physical phenomenon of energy transfer in the form of electromagnetic radiation.
We also apply an adaptive sample adjustment method and CUDA-based acceleration to our reconstruction process,
which provides much faster convergence than the random walk method.

\subsection{Flame Color and Temperature Measurement}
Most flame measurement methods for research involve temperature sensors \cite{baum1998examination, correia2001advanced},
lasers \cite{xue2001use}, and special optical systems \cite{albers1999schlieren}, \cite{hossain2014tomographic}, \cite{kang2015three}, \cite{schwarz1996multi}, \cite{toro2014flame}.
However, these methods fail to provide 3D reconstruction at a resolution high enough for realistic rendering of flames.

In recent years, the physical nature of flame illumination has been utilized using the mathematical black-body radiation model to aid in flame temperature reconstruction \cite{zhou20153}, \cite{wang2015image}, \cite{luo2007combustion}.
Unfortunately, the black-body radiation model by itself cannot provide realistic visual flame results, which limits its application in the computer graphics domain.

In terms of flame color intensity reconstruction, density-sheet decomposition \cite{hasinoff2007photo, hasinoff2003photo}, color-temperature mapping \cite{wu2014reconstruction},
and appearance transfer \cite{okabe2015fluid} methods have been introduced. These methods convert the volume reconstruction to parallel slices reconstruction and require a planar configuration of the input views; if the in-plane assumption is removed, then these methods do not work.
The grid optical tomography approach \cite{ihrke2004image, ihrke2006adaptive} was proposed to overcome this in-plane limitation, though it is not applicable to a more realistic rendering model.

\subsection{Camera Synchronization}
Consumer cameras have been used to reconstruct time-varying natural phenomena \cite{atcheson2008time, gregson2012stochastic}.
The key obstacle in the use of consumer cameras is the synchronization problem, due to the lack of synchronization hardware.

Previous works on the synchronization of multiple video sequences are based on feature tracking and geometric constraints
\cite{carceroni2004linear}, \cite{dai2006subframe}, \cite{lei2006tri}.
A different method based on detecting flashes \cite{shrestha2006synchronization} provided frame-level synchronization.
Recently, to solve the issue of rolling shutter shear and the synchronization problem of complementary metal-oxide-semiconductor (CMOS) consumer-grade camcorders,
a method based on strobe illumination and the subframe warp method was proposed \cite{bradley2009synchronization}.

Unfortunately, there are no obvious features that can be tracked in the flame videos. Moreover, the flame changes rapidly and irregularly, and 
therefore the synchronization accuracy of the frame-level or the sub-frame warp \cite{bradley2009synchronization} is unacceptable for the reconstruction of the flame.

\section{Algorithm}
\label{sec_algorithm}

In this paper, we present a method for flame volume reconstruction from videos. Fig.~\ref{fig_pipeline} shows the pipeline of our approach.
We present a smear-based CCD camera synchronization method to capture flame videos using multiple cameras.
For each video frame, we reconstruct color and temperature using an iterative approach.

For color reconstruction, the red, green, and blue (RGB) channels are reconstructed separately.
In essence, the volume data is first initialized with zeros, and for each iteration, the RTE based rendering model is used to render the reconstructed images.
After comparing the captured video frames with the reconstructed images, we adjust the volume data using the residuals of the pixel intensities.
The iteration process ends when the error difference between two successive iterations falls below 0.01.
The reconstruction is implemented using the CUDA parallel computing framework. The CUDA acceleration and the iterative approach provide a real-time visualization of the reconstruction results after every iteration.

For temperature reconstruction, the black-body radiation model is used to build the color-temperature mapping.
Following the same steps in the color reconstruction,
we achieve real-time visualization of the reconstruction results after every iteration by simply looking up the computed color-temperature mapping.

\begin{figure}[h]
\centering
\includegraphics[width=0.9\linewidth]{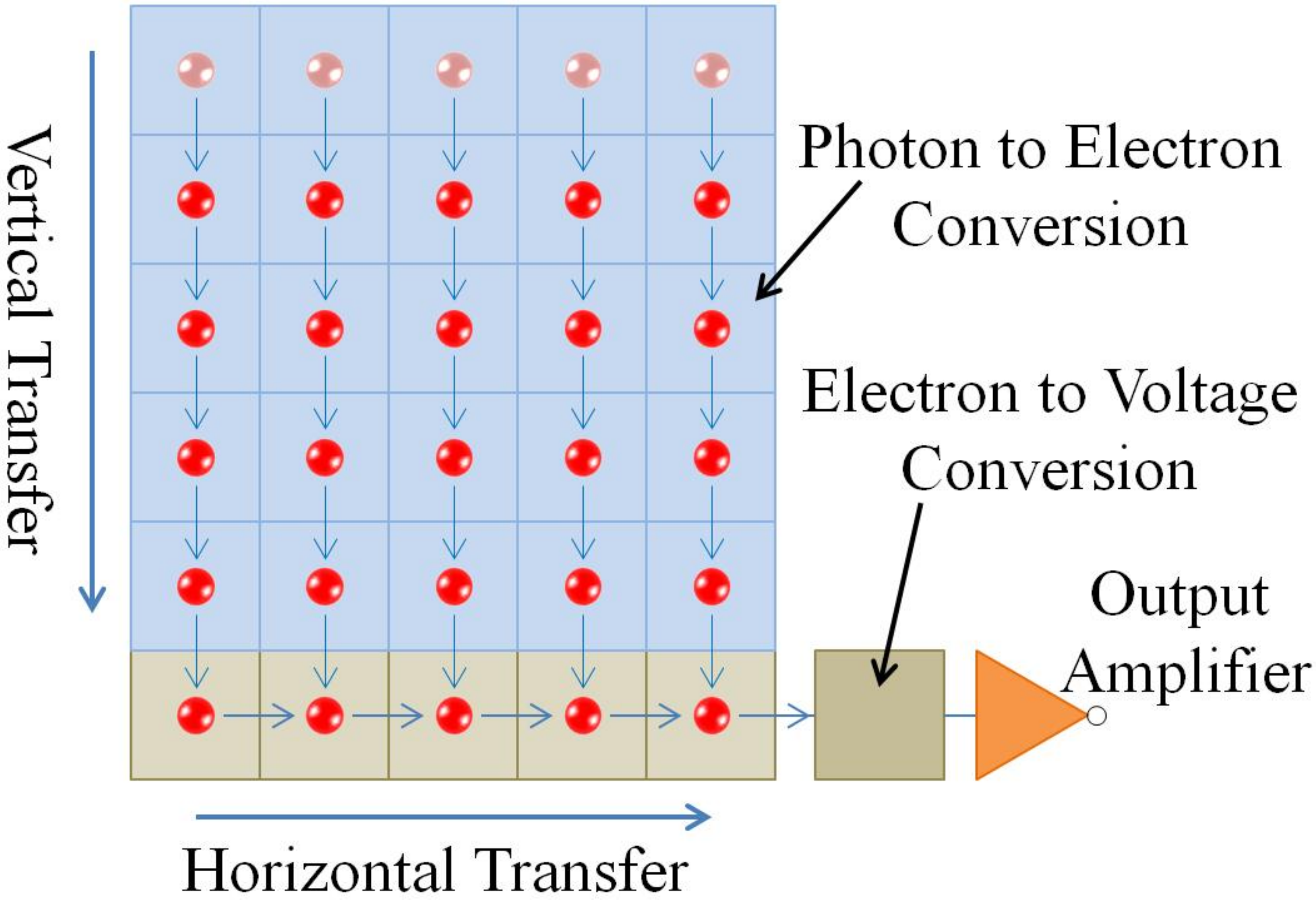}
\caption{CCD structure. The row-by-row vertical transfer leads to the generation of smear.
\label{fig_CCD_structrue}}
\vspace{-5mm}
\end{figure}

\subsection{Smear-Based CCD Camera Synchronization}
\label{sec_sync}

To solve the multi-view reconstruction problem, synchronized flame videos need to first be captured.
Traditionally, high-end industrial cameras containing synchronization hardware were used for this purpose. However, these cameras are expensive, costing at least 700 dollars per camera.
In our work, we utilize cheap consumer CCD cameras (that can cost as low as 100 dollars per camera) to achieve state-of-the-art flame volume reconstruction results in real-time. These cameras, however, do not contain any synchronization hardware. Therefore, we present a method to synchronize these inexpensive consumer CCD cameras using the smear phenomenon in them \cite{teranishi1987smear}. More specifically, we achieve synchronization by resetting the camera shutter until a fixed accuracy is obtained.


\subsubsection{CCD Smear}
\label{ccd_smear}

Common architectures of CCD image sensors include full-frame, frame-transfer, and interline architectures.
All of these architectures follow similar imaging processes as shown in Fig. \ref{fig_CCD_structrue}. More specifically,
the cells of the CCD sensor convert the gathered photons to electrical charges. These charges are vertically transferred to the horizontal readout line row-by-row.
For each row, after the horizontal transfer process, the charges are then converted to voltage information,
and finally the digital data for the image is retrieved through the amplifier.
The final image is generated by the same operations for all rows of the cells in the sensor.

\begin{figure}[!t]
\begin{center}
\setlength{\tabcolsep}{3pt}
\begin{tabular}{cc}
\includegraphics[width=0.23\textwidth]{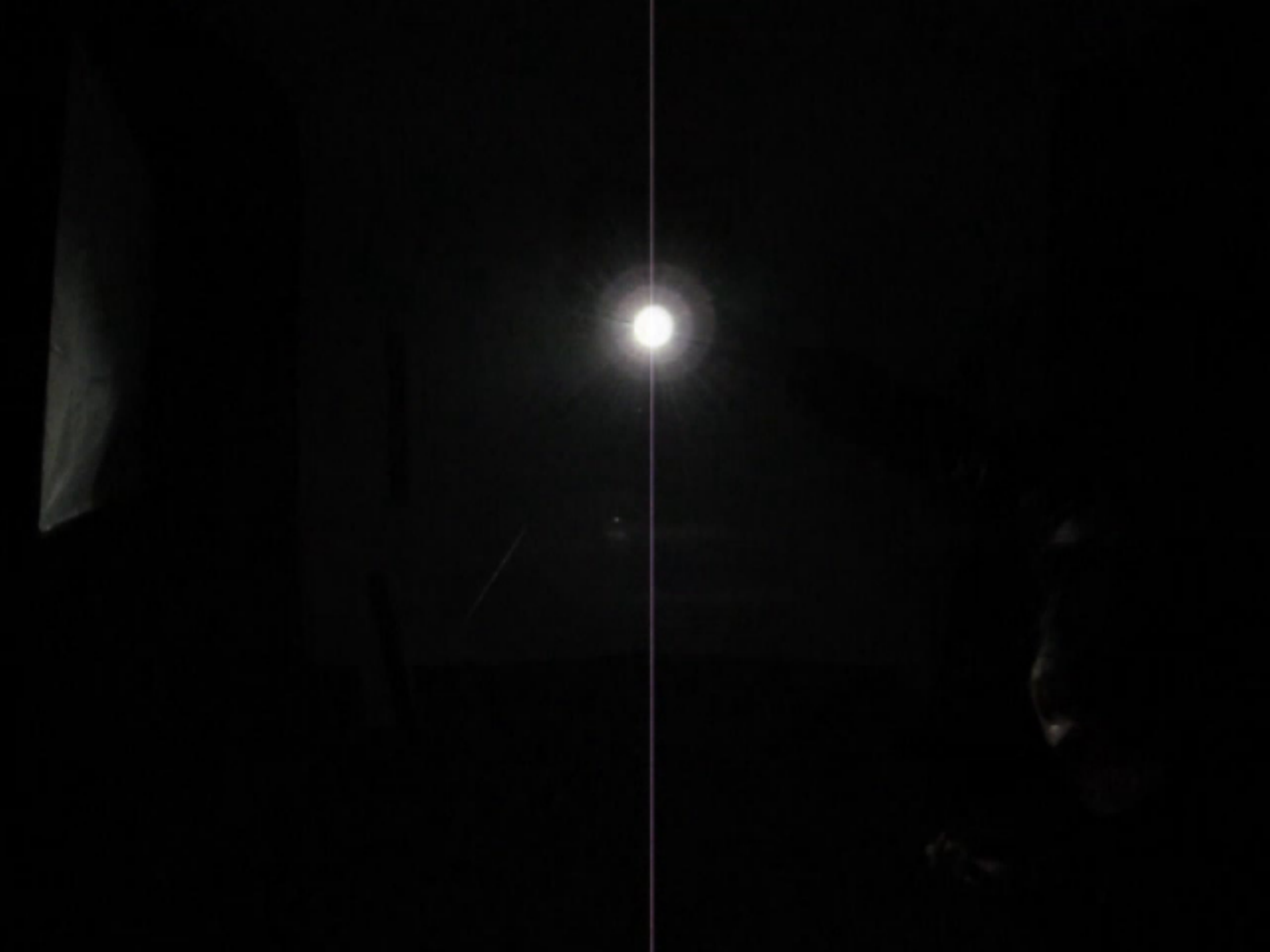}&
\includegraphics[width=0.23\textwidth]{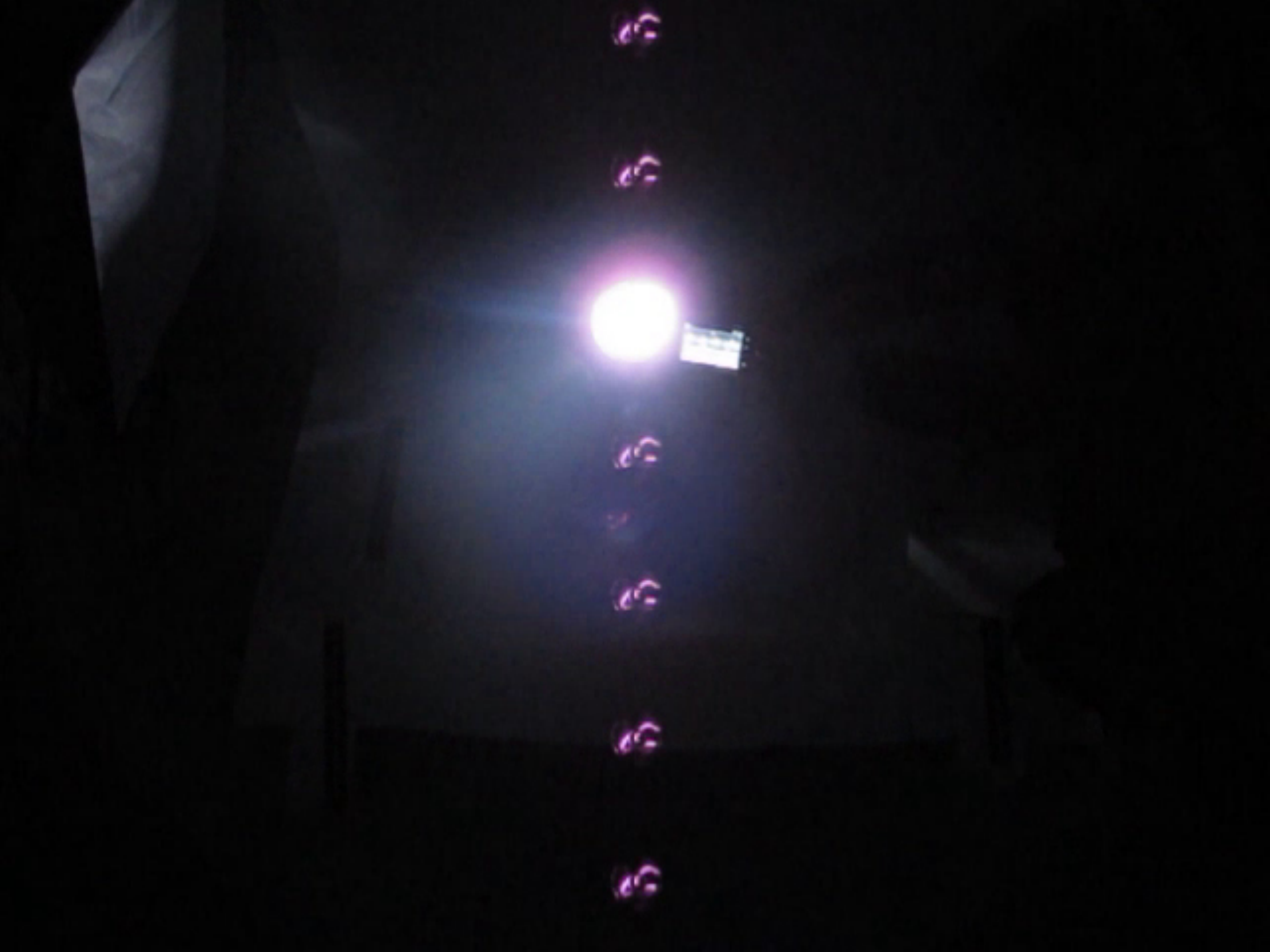}\\
(a) & (b)
\end{tabular}
\end{center}
\vspace{-4mm}
\caption{CCD smear. (a) The smear of a light source with constant lighting. (b) The smear of a strobe light source.
\label{fig_smear}}
\vspace{-1mm}
\end{figure}

If an intense light source is imaged onto the CCD image sensor,
undesired signals appear as a bright vertical (from top to bottom) stripe emanating from the light source location in the image.
The undesired brighter section around the light source which is the result of the overflow of charge is called ``blooming'',
and the vertical brighter section is called ``smear''.
Smear is produced by the incident light during the vertical transfer process.
Fig.~\ref{fig_smear} shows different smear patterns for different light sources.
When the light source emits a light with constant lighting, the cells of the CCD chip continue to capture the intense light during the vertical transfer process,
which leads to the generation of a vertical bright line along the light source position.
Similarly, when a strobe light source is used, the smear line would change to smear dots, since the strobe would only shine intermittently.

\subsubsection{CCD Camera Synchronization}
\label{subsec_ccd_sync}

\begin{figure*}[!t]
\begin{center}
\begin{tabular}{cc}
\includegraphics[width=0.48\textwidth]{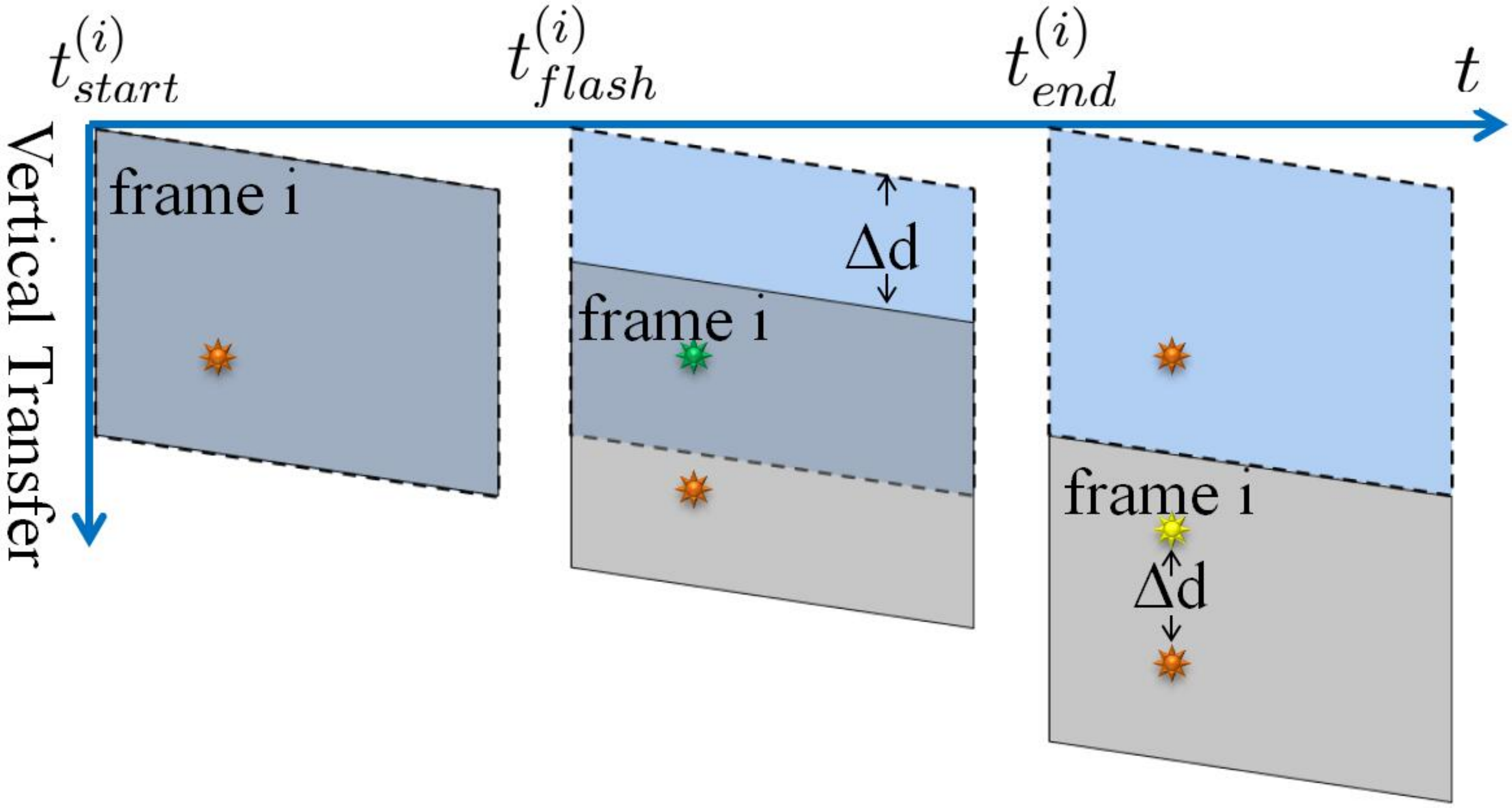}&
\includegraphics[width=0.48\textwidth]{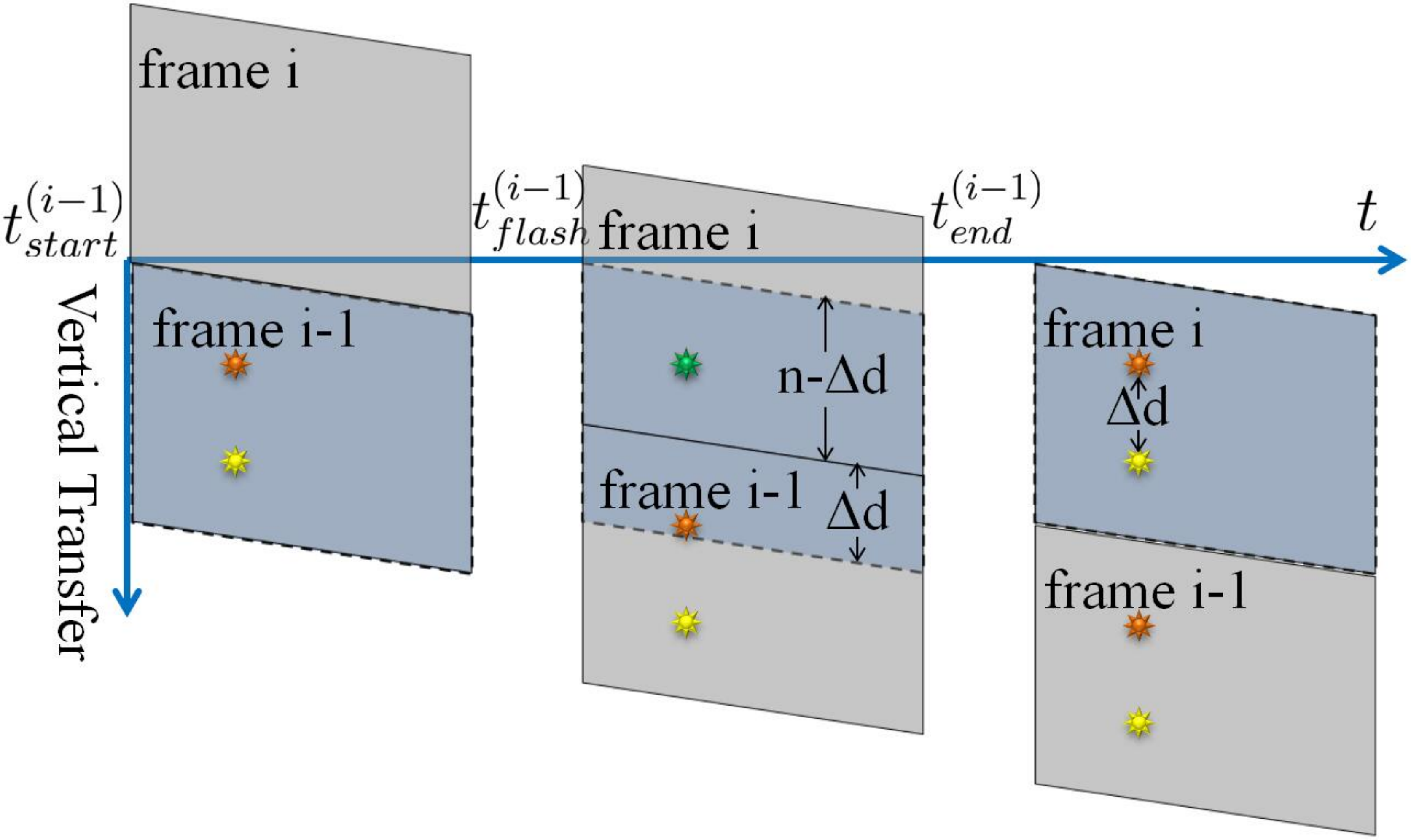}\\
(a) & (b)
\end{tabular}
\end{center}
\vspace{-3mm}
\caption{For different flash times, the smear dots would appear either above or below the light source. Scenarios (a) and (b) present the process of these two situations. An orange dot denotes the position of the light source in the images, and the yellow one represents the smear. When the strobe turns on, the light source is shown as a green dot.
\label{fig_smear_time}}
\end{figure*}

A video camera uses two procedures, acquisition and read, to produce one image frame.
In the acquisition phase, the cells of the CCD sensor convert the gathered photons to electrical charges. Then, in the read phase, the charges are vertically transferred to the horizontal readout line row-by-row.
To describe the synchronization method,
we focus on the time $t_{image\_read}$ taken to transfer the pixels for the final image.
Assuming the resolution of the image is $m\times n$, we get:
\begin{equation}
\label{eqn_t_image_read}
t_{image\_read} = nt_{per\_row}.
\end{equation}
Here, $t_{per\_row}$ denotes the time needed to transfer one row of the image, and is used to measure the error of the synchronization later.

To implement our synchronization method, we use CCD cameras to capture images of a strobe,
which has a controllable flash rate.
When we set a much higher flash rate than the CCD frame rate,
there will be several bright dots (smear) in one image, as shown in Fig. \ref{fig_smear}(b).
We can use two adjacent dots to compute $t_{per\_row}$ as:
\begin{equation}
\label{eqn_t_per_row}
t_{per\_row} = \frac{1}{\Delta d_{smear}f_{flash}}.
\end{equation}
Here, $\Delta d_{smear}$ denotes the distance in rows between two adjacent bright dots
on the same side (above/below) of the light source in the image, and
$f_{flash}$ denotes the flash rate which is given by the strobe instrument.

When the flash rate is lowered, the number of bright dots in the image decreases.
When the flash rate equals the video frame rate, the number of bright dots are either zero (the strobe turns on in the image acquisition phase)
or one (the strobe turns on in the image read phase).
In the single bright dot case, the dot could be either above or below the light source.
If the smear dot is above the light source, the smear is generated during the read phase of the current frame, as shown in Fig. \ref{fig_smear_time}(a).
The distance in rows, $\Delta d^{(i)}$, between the bright dot and the light source for the $i$th frame can be expressed as:
\begin{equation}
\label{eqn_d_src_smear_up}
\Delta d^{(i)} = \frac{t_{flash}^{(i)} - t_{start}^{(i)}}{t_{per\_row}},
\end{equation}

If the smear dot is below the light source, as shown in Fig. \ref{fig_smear_time}(b), the smear is generated during the read phase of the last frame:
\begin{equation}
\label{eqn_d_src_smear_down}
\Delta d^{(i)} = n - \frac{t_{flash}^{(i - 1)} - t_{start}^{(i-1)}}{t_{per\_row}}.
\end{equation}
Here, $t_{start}^{(i)}$ denotes the time needed for the $i$th frame to start the transfer, and $t_{flash}^{(i)}$ is the time needed for the strobe to turn on during the read phase of the $i$th frame, which results in the smear dot.

For the synchronization of multiple video cameras, we need $t_{start}^{(i)}$ for all cameras to be the same.
From Equations \ref{eqn_d_src_smear_up} and \ref{eqn_d_src_smear_down},
$t_{start}^{(i)}$ is determined by $\Delta d^{(i)}$, $t_{flash}^{(i)}$, and $t_{per\_row}$.
We make the cameras capture the same strobe images, so that the $t_{flash}^{(i)}$ is the same.
Inexpensive cameras of the same model still have good accuracy and stability with respect to the frame rate,
so $t_{per\_row}$ stays consistent.

Therefore, synchronization can be achieved by the following settings:
\begin{itemize}
\item Set the flash rate of the strobe to the same value as the frame rate of the cameras.
\item Keep the smear dot on the same side of the light source for all camera images.
\item Adjust the smear dots to make them equi-distant from the light source.
\end{itemize}

We have provided a detailed animation of the synchronization process in the supplementary video.

\subsubsection{Data Capture}
\label{subsec_DCP}

\begin{figure}[t]
\begin{center}
\begin{tabular}{cc}
\includegraphics[width=0.22\textwidth]{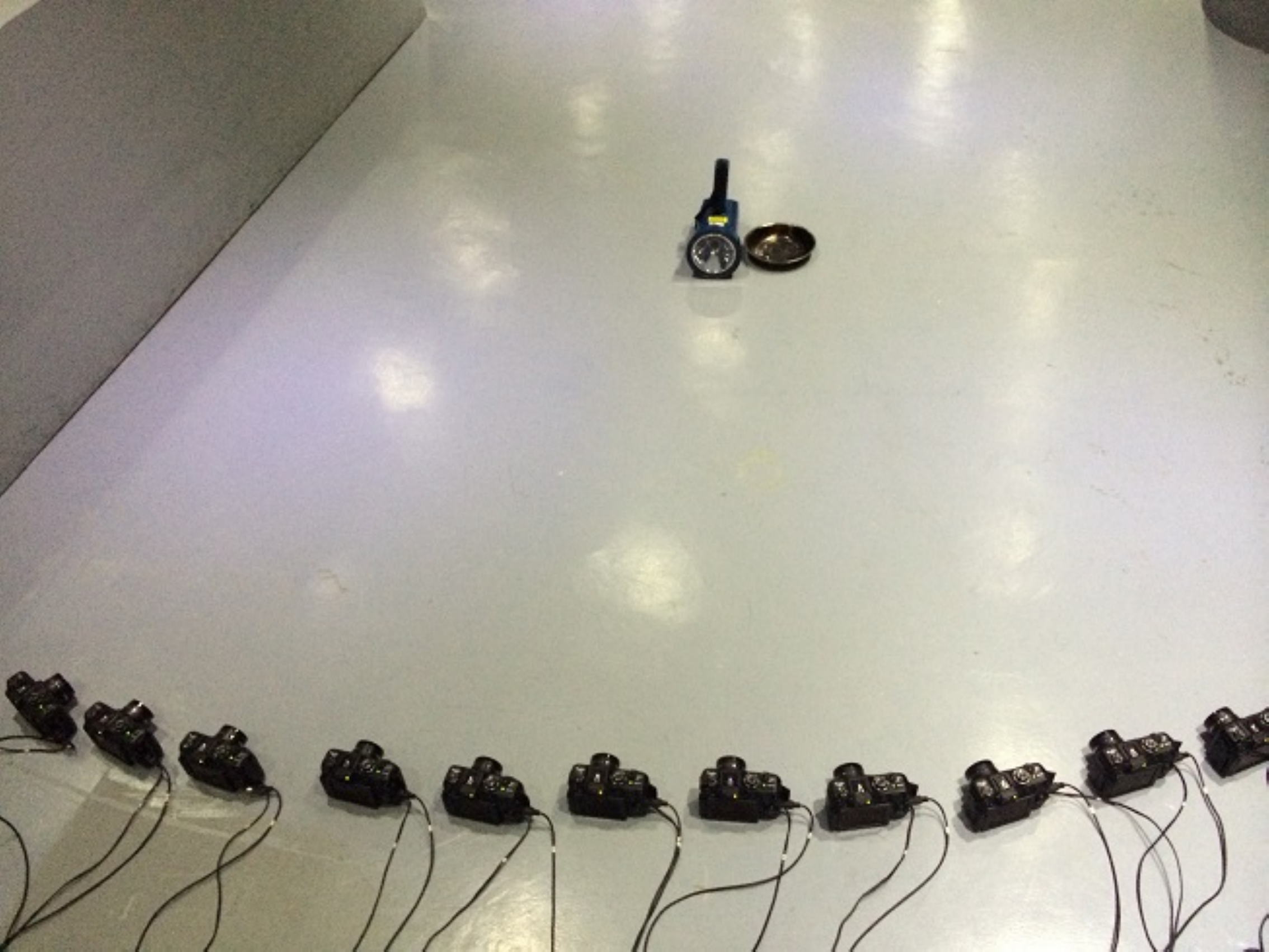}&
\includegraphics[width=0.22\textwidth]{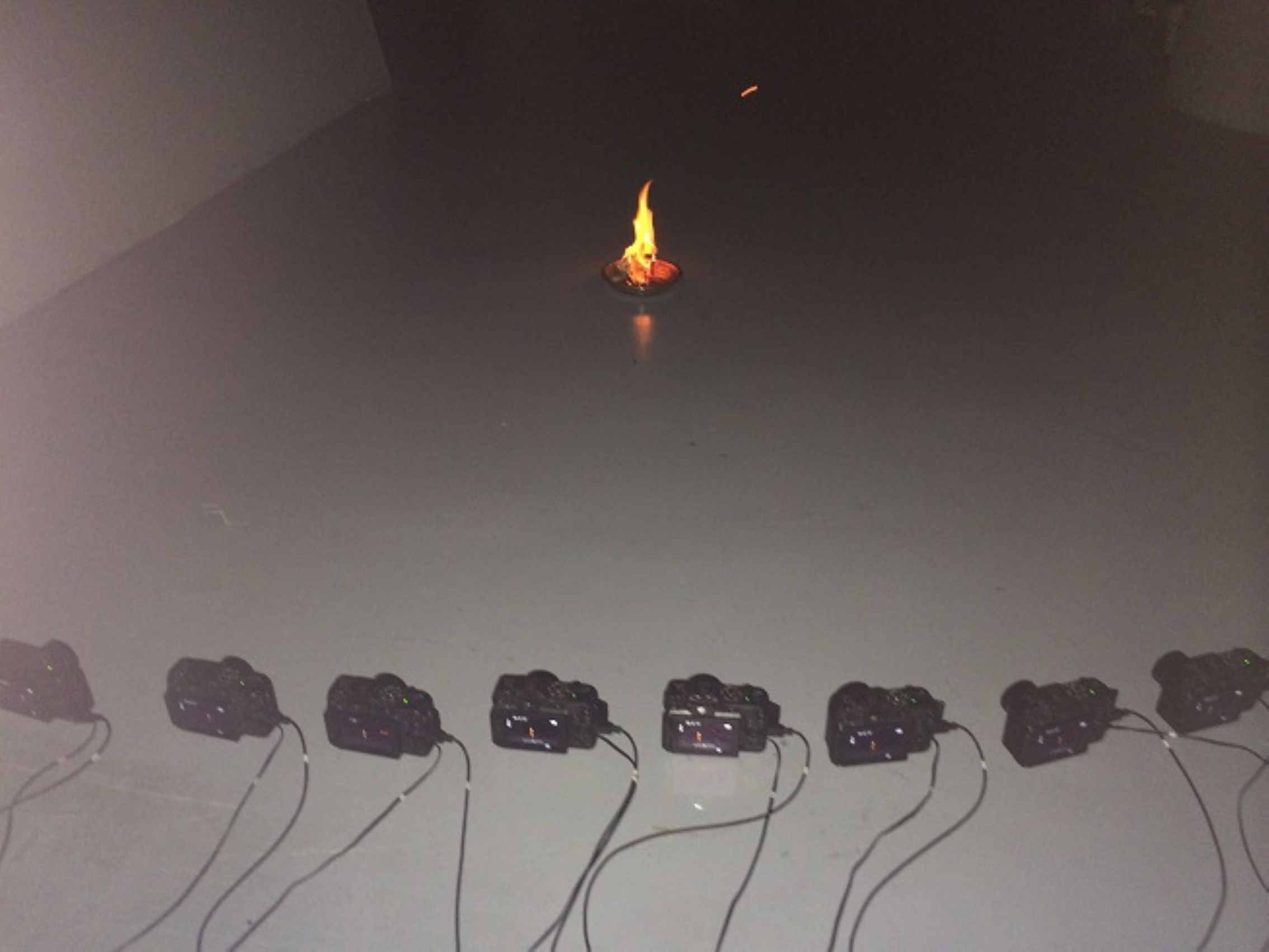}\\
(a) & (b)
\end{tabular}
\end{center}
\vspace{-4mm}
\caption{Data capture setup. (a) The scene of camera synchronization using a stroboscope. (b) A scene of the flame videos capture.
\label{capture_scene}}
\end{figure}

To implement our method for flame reconstruction,
we use eleven Canon PowerShot G12 cameras to capture RGB videos of flames with a resolution of $1280 \times 720$, as shown in Fig. \ref{capture_scene}.
In order to easily calibrate the cameras,
the cameras are set up in an arc-like topology to ensure that all of the cameras see the calibration board simultaneously.
Moreover, this topology prevents cameras from capturing images of each other.
Unlike previous works \cite{hasinoff2007photo}, \cite{wu2014reconstruction}, \cite{hasinoff2003photo}, the camera positions are not restricted to be on a strict plane,
which makes our approach more practical.
The distance between the flame and each camera is approximately one meter.
All of the cameras are set to the same configuration to obtain the same response to the incident light.
During the experiments, we ignite paper and alcohol with different particles to generate different flames.

For the smear-based synchronization method, we use a Monarch Instrument Nova-Strobe dbx stroboscope as the light source.
The dbx has flash rates ranging from 30 to 20000 flashes per minute that are adjustable in 0.1 step increments.
We set the rate to 23.98 flashes per second, which is the same as the video frame rate.

The distance between the smear dot and the light source is determined by the start time of the shutter,
so we adjust the start time of the shutters to make the distances for each camera almost the same.
Consequently, we achieve the synchronization of the cameras as described in Section \ref{subsec_ccd_sync}.
For the G12 camera, the start time of the shutter can be adjusted using the button for switching between different resolutions in the video mode.
Since this adjustment requires manual intervention, we do not expect to obtain the exact same distances for each camera.
However, we can easily set a distance within a 100-pixel offset within 5 trials for each camera.
The resultant time to transfer one row of the pixels is about 54 $\mu s$ using Equation \ref{eqn_t_per_row}.
Therefore, we can easily control the accuracy of our synchronization within $100 \times 54 \mu s = 5.4 ms$,
much less than the frame-level synchronization, $1s \div 23.98 = 41.7ms$.
We can achieve more accurate synchronization if the 100-pixel offset distance for each camera is reduced even further with more trials.

\subsection{Rendering Model}\label{sec_image_model}
In this section, we provide a concise introduction to the rendering model used in this paper.
Instead of the simplified linear optical model \cite{hasinoff2007photo}, \cite{ihrke2004image}, \cite{wu2014reconstruction},
\cite{hasinoff2003photo},
\cite{ihrke2006adaptive}
or parallel projection
\cite{hasinoff2007photo}, \cite{wu2014reconstruction},
\cite{hasinoff2003photo},
we use a radiative transport equation (RTE) based rendering model and perspective projection for the reconstruction.

The RTE is an integro-differential equation which describes the light transport in participating media, as follows:
\begin{equation}
\label{equ_RTE}
\begin{split}
(\vec{\omega} \cdot \nabla )L(\vec{x},\vec{\omega })=
	&  + \sigma_a(\vec{x})L_e(\vec{x},\vec{\omega }) \\
	&  - \sigma_t(\vec{x})L(\vec{x},\vec{\omega }) \\
	&  + \sigma_s(\vec{x})\int_S{L(\vec{x},\vec{\omega}_i) \cdot p(\vec{\omega},\vec{\omega}_i)d\omega_i}.
\end{split}
\end{equation}
Here, the left side of the equation denotes the variation of the spectral radiance in the direction of interest $\vec{\omega}$. $L$ is the spectral radiance, $L_e$ the emitted spectral radiance, $\sigma_a$ the absorption coefficient,
$\sigma_s$ the scattering coefficient, $\sigma_t = \sigma_s + \sigma_a $ the extinction coefficient, and
$p(\vec{\omega},\vec{\omega}_i) $ the phase function, which specifies the spherical distribution of the scattering light at a position.

The optical model \cite{max1995optical} for volume rendering is described as follows:
\begin{equation}
\label{eqn2}
E \left(s_e \right) = \int_{s_b}^{s_e} T (s, s_e) L(s) ds
+ T (s_b, s_e) E(s_b).
\end{equation}
Here, $E \left(s \right)$ denotes the intensity at position $s$,
$T(s_1, s_2)$ is the transparency from $s_1$ to $s_2$,
$s_b$ is the position of the background,
$s_e$ is the position of the eye point, and
$L(s)$ is the radiative contribution at position $s$.

\subsection{Volume Reconstruction}
\label{sec_volume_rec}

In terms of the goal of flame volume reconstruction from video frames,
reconstructing the color intensity \cite{hasinoff2007photo}, \cite{ihrke2004image}, \cite{hasinoff2003photo} is intuitive.
However, some works have also focused on temperature reconstruction \cite{wu2014reconstruction}, \cite{wang2015image}.
In this work, we reconstruct both color and temperature using our method.
Instead of building a linear system \cite{ihrke2004image} or designing a random walk framework \cite{gregson2012stochastic},
we dynamically adjust the volume data in an iterative way. To explain our method clearly, we provide the following definitions:

\textbf{Key points}: Vertices of voxels in the reconstructed volume.
Each key point contains a value and all key points constitute the reconstructed volume.
The goal of our method is to reconstruct the values of the key points.

\textbf{Sample points}: Points used to render the flames and to adjust the value of their eight neighboring key points during the adjustment phase.

%

\subsubsection{Volume Rendering}
\label{subsec_rendering}

Given the volume data,
we use the rendering model described in Section \ref{sec_image_model} to render the reconstructed images.
When traveling through the flame, traced rays would be curved \cite{seron2005implementation, zhao2007visual}.
As we deal with small scale flames at present, we ignore this effect for simplification.
Rays are cast into the volume,
and sample points on each ray are used to determine the final intensity of the corresponding pixels.
We calculate the radiance of the sample points, and then integrate them to obtain the pixel color.
In order to integrate the radiance of sample points on one ray,
we apply under blending \cite{bavoil2008order} in front-to-back order
using the following equations:
\begin{equation}
\label{equ_blending_color}
L_{dst} = T_{dst}(T_{src}L_{src}) + L_{dst},
\end{equation}
\begin{equation}
\label{equ_blending_alpha}
T_{dst} = ( 1 -  T_{src})T_{dst}.
\end{equation}

The radiance of sample points is subject to the emission, extinction, and scattering phenomena,
corresponding to the three summands in the right side of Equation \ref{equ_RTE}.
The description of these phenomena is as follows.

\textbf{Emission}: During the process of combustion, some photons are emitted by the hot soot particles and the emission results in the light of flames \cite{max1995optical}.
We directly model the volume data as color intensities which are used for the emission phenomenon.

\textbf{Extinction}: While traveling through the flame, some photons are absorbed or scattered in different directions. This is called extinction.
The extinction coefficient $\sigma_t$ in Equation \ref{equ_RTE} is relative to the number of particles in a voxel \cite{max1995optical}.
The sophisticated optical models specify the extinction coefficient as a finite, continuously varying function of the volume data.
Here, we assume $\sigma_t$ to be a constant for simplification.

\textbf{Scattering}: The scattering phenomenon is defined as the process by which photons travel from the direction of interest to different directions (out-scattering), or from different directions to the direction of interest (in-scattering).
The distribution of scattered light is specified by the phase function

\begin{equation}
\label{equ_phase_func}
p(\vec{\omega},\vec{\omega}_i) = \frac{1}{4\pi} \cdot
\frac{ 1 - g^2}{\left( 1 + g^2 + 2g (\vec{\omega} \cdot \vec{\omega}_i) \right)^{\frac{3}{2}}}.
\end{equation}
Here, $g$ is a value in the range $(-1, 1)$, and
$g < 0$, $g = 0$, and $g > 0$ correspond to backward, isotropic, and forward scattering, respectively. Here, we use an isotropic phase function for simplification \cite{klehm2014property}.
As described in \cite{pegoraro2006physically}, in a low-albedo medium such as the flames of interest, extinction usually dominates largely over scattering which contributes very little to the final image. Moreover, we opt for isotropic scattering because of the computational efficiency.

\subsubsection{Volume Adjustment}
\label{subsec_adjustment}

Tomographic reconstruction approaches \cite{hasinoff2007photo, ihrke2004image, wu2014reconstruction} minimize the residuals of intensities between the input and the reconstructed images to drive the reconstruction process. We follow a similar residual intensity minimization approach for our reconstruction.

After the reconstructed image is generated by the volume rendering process,
a ray then travels through the reconstructed volume and several sample points are generated.
Each sample point is used to adjust the value of its eight neighboring key points
based on the sample point transparency, the distance between the sample point and the key point,
and the residuals between the captured and the reconstructed images.
The adjustment is done as follows:

\begin{equation}\label{formula_adjust_value}
{A}_{j,s}  =
\alpha_{l}\Delta L(x,y)
T(s, s_e) /
e^{\alpha_{d}D(j,s) },
\end{equation}
\begin{equation}\label{formula_temperature_adjustment}
V_{j,k+1} = V_{j,k} + \sum_{s} G(s) {A}_{j,s}.
\end{equation}

Here, $V_{j,k}$ denotes the volume data of the key point $j$ after $k$ iterations, and
${A}_{j,s}$ denotes the value of adjustment for the key point $j$ influenced by the sample point $s$.
As shown in Equation \ref{formula_temperature_adjustment},
for each iteration, we adjust the volume data of the key point $j$
by adding each ${A}_{j,s}$ weighted by a random number $G(s)$, which is generated from a Gaussian distribution with mean equal to 1.
For ${A}_{j,s} $, we consider three aspects, as shown in Equation \ref{formula_adjust_value}:
\begin{itemize}
\item $\Delta L(x,y)$: the residual of color intensity between the source and the reconstructed image at pixel $(x,y)$.
\item $T (s, s_e)  $: the transparency from the sample $s$ to the eye position.	This is used to balance the adjustment of different regions in the reconstructed volume.
\item $D(j,s)$: the distance between key point $j$ and sample point $s$.
\end{itemize}

$\alpha_{l} $ and $\alpha_d $ are scale factors employed
to describe the quantity of each part which contributes to the adaptive adjustment value.

We discuss each part in detail in the following sections.

\textbf{Color Constraint}: The ultimate goal of the reconstruction is to produce flame frames with the same color as the captured frames.
Therefore, the differences of pixel intensity between the captured frames and the reconstructed frames
are used to reduce the residuals between the input and reconstructed images.
The residual is given as $\Delta L(x,y)   = L_{src}(x,y)   - L_{rec}(x,y)  $,
where $L_{rec}(x, y)$ denotes the intensity of pixel $(x, y)$ in the reconstructed image, and
$L_{src}(x, y)$ the intensity of pixel $(x, y)$ in the captured source frame.

\textbf{Transparency}: Given that we use the under operator to blend the samples in front-to-back order,
assuming the transparency for all samples to be a constant $\tau \in (0, 1)$
and $k$ is the order of the sampling for a ray,
the blending transparency of the $k$th sample is computed as follows:
\begin{equation}
	\label{formula_trans}
	T (s, s_e) = (1-\tau)^{k - 1},
\end{equation}
which is in the range $(0, 1)$ and decreases exponentially with the sample order $k$.
The longer the distance between the sample and the eye,
the less contribution the sample makes towards the final intensity.
Therefore, $T(s, s_e) $ is used to scale the adjustment value.

\textbf{Distance Constraint}: It is apparent that the samples closer to the key points would cause greater changes to the volume data than the values of the key points.
We use the Euclidean distance between key point $j $ and sample point $s $ to evaluate the adjustment value for the distance constraint part. The transformation $1 / e^{D(j,s)} $ constrains this part in the range $(0, 1]$.

\subsubsection{Temperature Reconstruction}
\label{subsec_tem_rec}

In order to reconstruct the black-body radiation based flame temperature,
we first build the color-temperature mapping \cite{nguyen2002physically},
and then follow the same steps as the color reconstruction for the green channel, as described above.
Meanwhile, to provide the visualization results after every iteration,
we look up the color-temperature mapping on the fly for rendering.

We select the green channel for temperature reconstruction because of the following reasons:
\begin{itemize}
\item When judging the relative luminance of different colors in well-lit situations, humans tend to perceive light within the green parts of the spectrum as brighter than the red or blue lights of equal power \cite{CIEcolor}.
\item Based on the color mapping methods \cite{nguyen2002physically}, \cite{CIEcolor}, we found that the intensity of the green channel grows with the temperature within a certain range.
\end{itemize}
Therefore, the relationship between the green channel and temperature is confirmed based on the observation and the computational results.
The color-temperature mapping is created as follows.
In the local thermodynamic equilibrium (LTE) system, the black-body radiation law provides
that the electromagnetic spectral radiance emitted by the medium (black-body) is directly indicated
by the temperature of the medium of a given distribution of wavelength \cite{rushmeier1995volume},
which fits the situation in most fires.
Ideally, we assume these kinds of fire act as a black-body and compute the emitted spectral radiance using Planck's formula:

\begin{equation}
L \left( \lambda , T \right) = \frac {2 hc^2} {\lambda^5} \frac {1} {e^{\frac {hc} {\lambda k T}} - 1}.
\end{equation}
Here, $T$ denotes the temperature, $k$ is the Boltzmann constant, $h$ is the Plank constant, and $c$ is the speed of light in the medium.
This spectrum can then be converted to RGB \cite{wu2014reconstruction}, \cite{pegoraro2006physically}, \cite{nguyen2002physically},
building the color-temperature mapping.

\section{Implementation}
\label{sec_implementation}

In this section, we present the details of the data processing and acceleration techniques used in our algorithm.

\subsection{Data Pre-processing}\label{subsec_DCP}

The flame videos are captured in a dark environment.
For the data pre-processing, we split the videos into individual frames. Pixels with intensities greater than 30 are set as flame pixels, and the non-flame pixels are assigned the color black.
After the color calibration \cite{ihrke2007reconstruction},
the processed frames are used as input to our algorithm.
To map the camera coordinates to the object (flame) coordinates,
the standard techniques for camera calibration \cite{zhang1999flexible}
are used to compute the intrinsic and extrinsic parameters of cameras.

\subsection{Sampling Step and Volume Dimension}\label{subsec_vdass}

The 3D volume consists of voxels and an appropriate sampling step needs to be set to access all these voxels.
Since we use perspective projection in the reconstruction, only a few rays vertically intersect with the reconstructed volume cube. Besides, we try to deal with small scale flames, hence few rays intersect with the volume with large angles.
If the sampling step is large, then some volume data will be missed along the ray.
In contrast, if the sampling step is small, then duplicate data will be accessed.
Therefore, we set the sampling step equal to the voxel edge length, so that samples on the same ray would evenly access one voxel ideally.

To determine the volume dimensions, we first measure the position and size of the flames in the world coordinate system, which is determined by a calibration board \cite{zhang1999flexible}.
Then, given the camera parameters, we calculate the projection pixels of the eight vertices of the volume.
Ideally, we obtain twelve projected lines $l_I$ corresponding to the twelve edges of the volume cube for each input image $I$.
The lines $l_I$ are then projected to the image $uv$ coordinate, and each line $l_{(I, uv)}$ will occupy $N_{(l, I, u)}, N_{(l, I, v)}$ of pixels along the $u$ and $v$ dimensions, respectively.
Let $n_{(max, d)}$ denote the maximum number of $N_{(l, I, u)}$ and $ N_{(l, I, v)}$, and $l_d, (d \in \{x, y, z\})$ the length of the volume in dimension $d$.
We set $\frac{n_{(max, d)}}{\alpha}$ as the number of voxels in dimension $d$.
Then, the edge length of each voxel is $ \frac{\alpha l_d}{ n_{(max, d)}}$, and the dimensions of the other two directions can similarly be calculated.
A large dimension leads to a waste of memory and some voxels might not be accessed by any ray. In contrast, a small dimension leads to low resolution of the reconstructed results.
In experiments, we use $\alpha = 1.5$ to access all the voxels in our flame volume data.

\subsection{CUDA-Based Acceleration}\label{subsec_CBA}
In order to accelerate our algorithm, we implemented the visual hull computation, volume rendering, and adjustment using CUDA.
\subsubsection{Visual Hull}
To reduce the number of volume values to be reconstructed,
we use a visual hull \cite{laurentini1994visual} in our reconstruction method.
We exploit the bitwise operations in the CUDA kernel to achieve good performance
and obtain the visual hull within 200 milliseconds for ten input views and $128^3$ voxels.
As shown in Algorithm \ref{alg_visual_hull}, we first initialize the visual hull tags of the volume with 0.
Then, we assign the $imageIndex$ from 0 to the number of input views ($cameraNumber - 1$),
and the visual hull tags are updated with the fast bitwise operation of $imageIndex$.
Only the rays cast from the flame pixels are used in this process.
Finally, if the last $cameraNumber$ bits of the voxel tag are all equal to 1, the voxel is inside the visual hull.
Our reconstruction method only reconstructs the volume data inside the visual hull.
We simply assign a small negative number to the key points outside the visual hull, which will not visually affect the final reconstructed results.
\begin{algorithm}
\small
\caption{Visual Hull Computation on GPU}
\label{alg_visual_hull}
\begin{algorithmic}[1]
\STATE Initialize all visual hull tags of the volume with 0
\FOR {each input image}
	\FOR { flame rays in parallel}
    		\FOR { each visualHullTag around the ray}
	       		\STATE visualHullTag $|=$ (1 $<<$ imageIndex)
    		\ENDFOR
	\ENDFOR
\ENDFOR
\end{algorithmic}
\end{algorithm}

\subsubsection{Rendering and Adjustment}
To visualize the reconstruction results after every iteration in real time,
we apply GPU acceleration with CUDA in rendering and adjustment processes, as shown in Algorithm~\ref{alg_render_adjust}.
The computation task for rendering is independent for each ray.
Also, the process of adjusting the temperature field is highly parallel,
except for writing the adjusted value to the same key point from some rays,
which could be solved using atomic operations.

In the implementation, the volume data is loaded in the 3D texture memory of the GPU,
and the visual hull tags and the source frame data are loaded in the global memory of the GPU.
In terms of the temperature reconstruction, the color-temperature mapping is loaded in the 1D texture memory of the GPU for fast access.
In the rendering process, every ray cast from the camera optical center is computed by a CUDA kernel.
Every kernel computes the positions of sample points,
queries the sample values from the texture memory,
and calculates the final intensity of the corresponding pixels.
In the adjustment process,
residuals of intensity, the transparency, and the distance between the key point and the sample point
are applied to adjust the volume data using atomic operations.

\begin{algorithm}
\small
\caption{Volume Data Rendering and Adjustment on GPU}
\label{alg_render_adjust}
\begin{algorithmic}[1]
\STATE // Rendering
\STATE totalrays $\leftarrow$ image size
\FOR { ray \textless \text{ totalrays} in parallel}
    \FOR { each sample point on the ray}
        \STATE calculate values from texture memory
    \ENDFOR
    \STATE integrate the values
\ENDFOR
\STATE // Adjustment
\FOR { flames ray in parallel}
    \FOR { each sample point on the ray}
        \STATE compute the adjustment value A of 8 key points
	    \STATE add the As to the key points using atomic operation
    \ENDFOR
\ENDFOR
\end{algorithmic}
\end{algorithm} 
\section{Results}
\label{sec_results}

To evaluate our reconstruction algorithm,
we analyze the performance of our approach on simulated and real captured data for both the color intensity and temperature reconstruction.
Experiments are performed on a laptop with an Intel Core i7-4710MQ CPU 2.50GHz, with 8GB of memory, and an NVIDIA GeForce GTX 850M graphics card.
\subsection{Color Intensity Reconstruction}

We perform a quantitative evaluation of our method using simulated data, where the fire is modelled by a physically-based method \cite{hong2010geometry}.
Using the simulated data, we render multi-view flame images with the method, discussed in Section \ref{sec_image_model},
and use the images as input to our reconstruction method.

We use ten input images from different views with intervals of 36$^\circ$ in a plane.
In reality, the cameras can not strictly be constrained to a horizontal plane,
(i.e., some cameras might look up at the flame and some might look down) and thus we set up the virtual cameras in such a manner.
We render the flame images with a resolution of  $640\times480$.
The simulated volume data is assumed to have a 20cm$\times$20cm$\times$20cm volume
and the center of the volume is the origin.
The cameras are set 30cm away from the origin with a field of view angle of 60$^\circ$.
In the reconstruction process,
the coefficients in Equation \ref{formula_adjust_value}
are set experimentally as $\alpha_{l} = 0.001$ and
$\alpha_{d} = 1$. The value of $\tau$ in Equation \ref{formula_trans} is chosen as 0.05.
For flames of different sizes, we change the value of $\alpha_{d}$ proportionally with the length of the volume edge.
%

The difference between the input images and the reconstructed results is computed as pixel intensity error
for the red, green, and blue channels, within the range [0, 255].
To compute the pixel intensity errors,
we first specify a rectangle in the flame image as a bounding box (which contains the whole flame),
and then we compute the root-mean-square error (RMSE) of all pixels in the bounding box.
The difference in terms of the volume data between the ground truth and the reconstructed one
is computed as the volume data error for the red, green, and blue channels.
As we focus on the visual effect of the reconstruction,
only the data in the visual hull is compared during the RMSE computation,
and regions outside the visual hull are ignored.

\begin{figure}[t!]
\begin{center}
\begin{tabular}{c}
\includegraphics[width=0.4\textwidth]{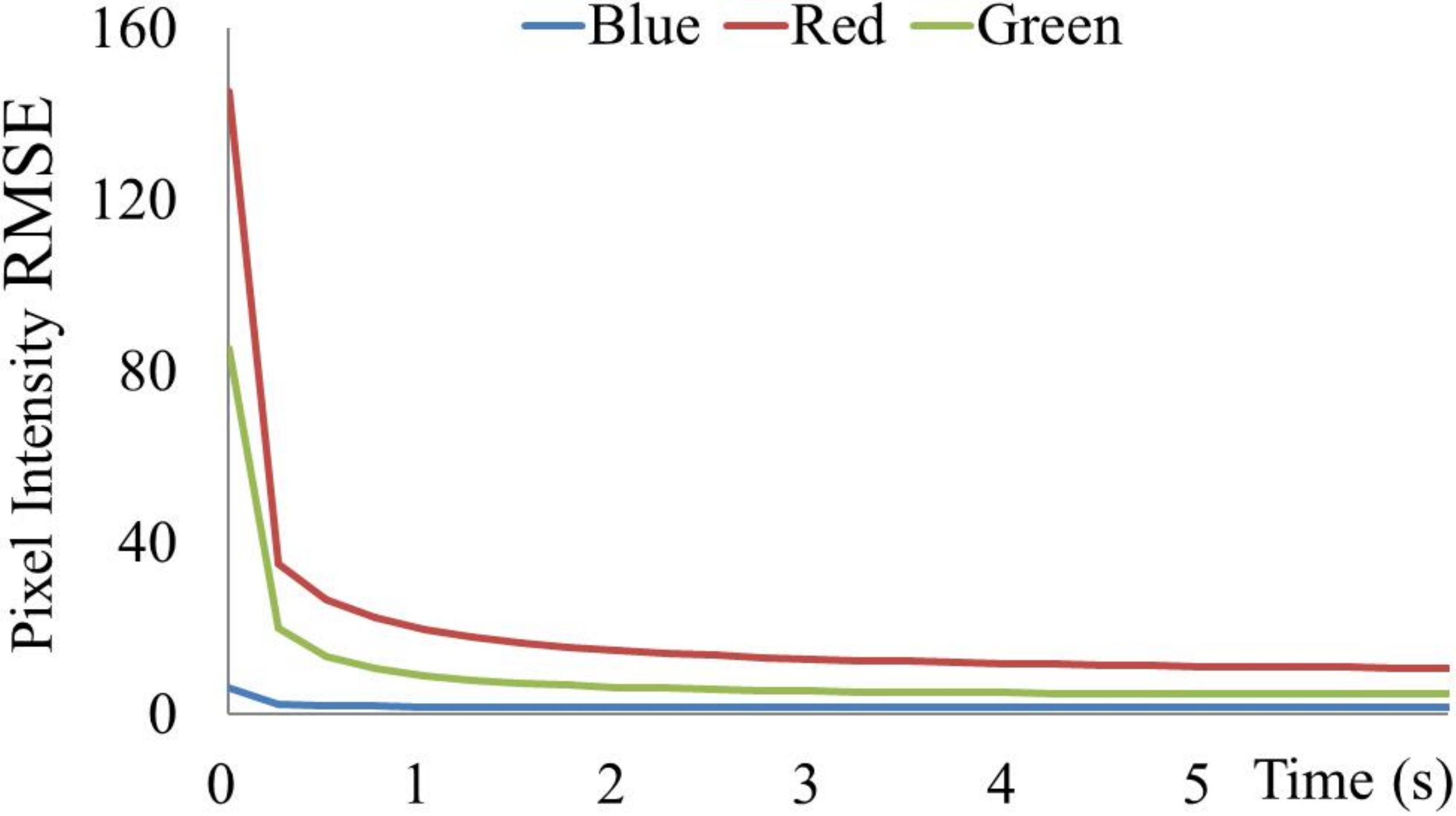}\\
(a)\vspace{1mm} \\
\includegraphics[width=0.4\textwidth]{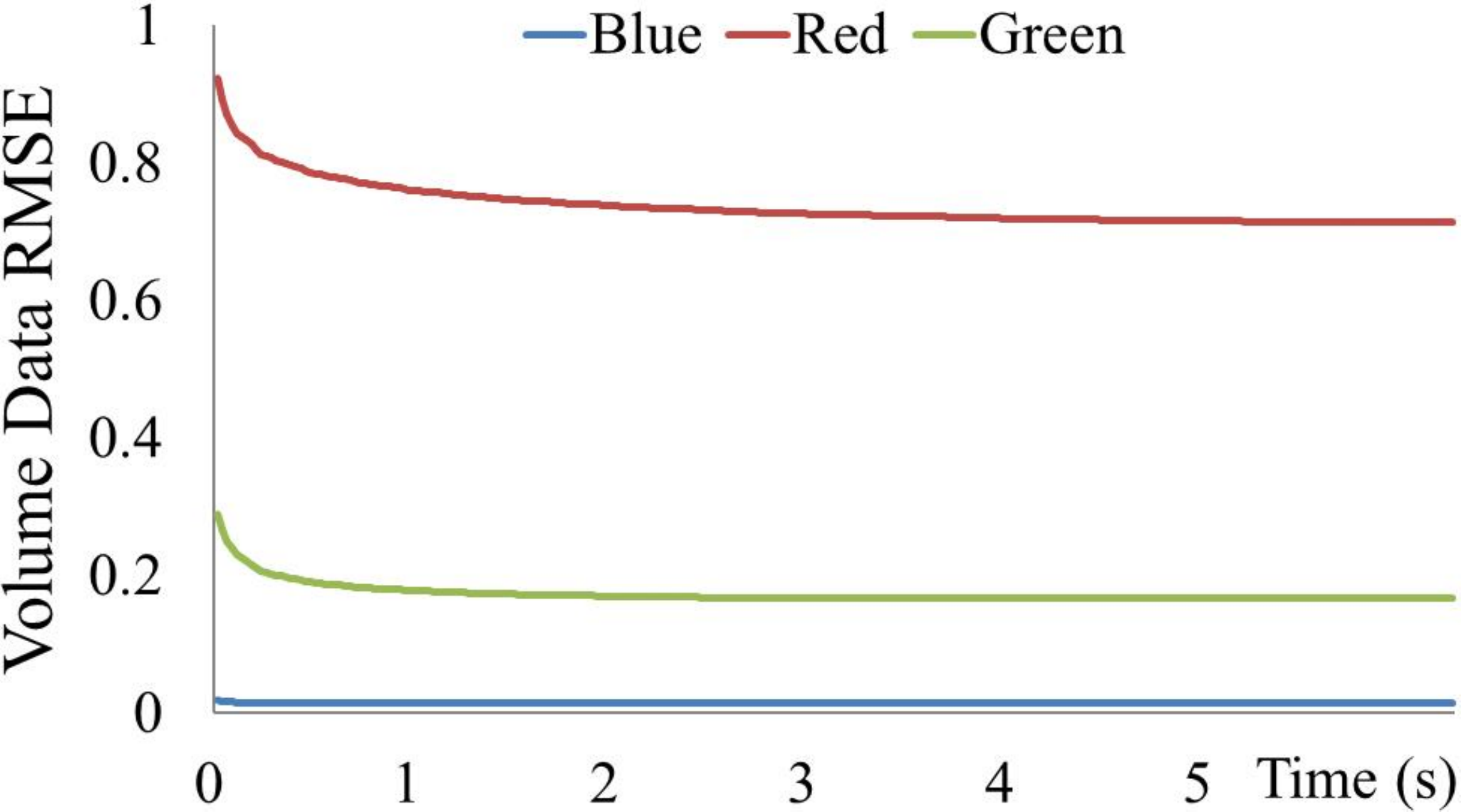}\\
(b) \\
\end{tabular}
\end{center}
\vspace{-4mm}
\caption{(a) Relation between pixel intensity and the runtime of our method. (b) Relation between the volume data error and the runtime. Pixel intensity error and volume data error are efficiently reduced within two seconds.
\label{fig_result_sim_iteration}}
\vspace{-1.5mm}
\end{figure}

\begin{figure}[t!]
\begin{center}
\begin{tabular}{cccc}
\setlength{\tabcolsep}{3pt}
\includegraphics[width=0.09\textwidth]{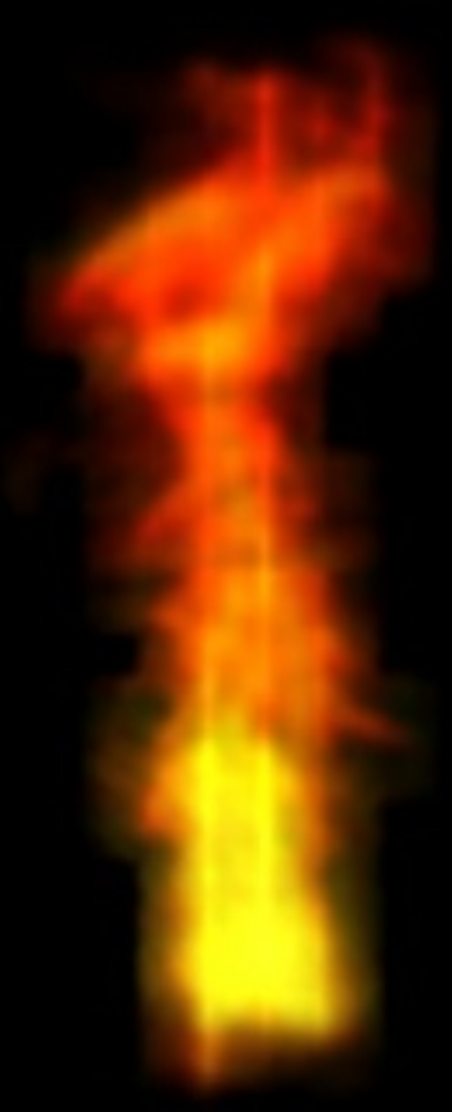}&
\includegraphics[width=0.09\textwidth]{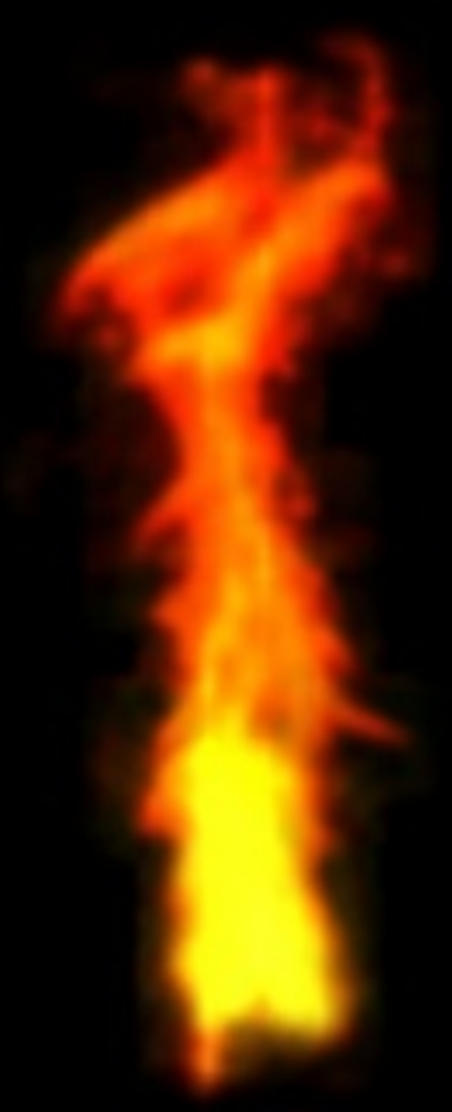}&
\includegraphics[width=0.09\textwidth]{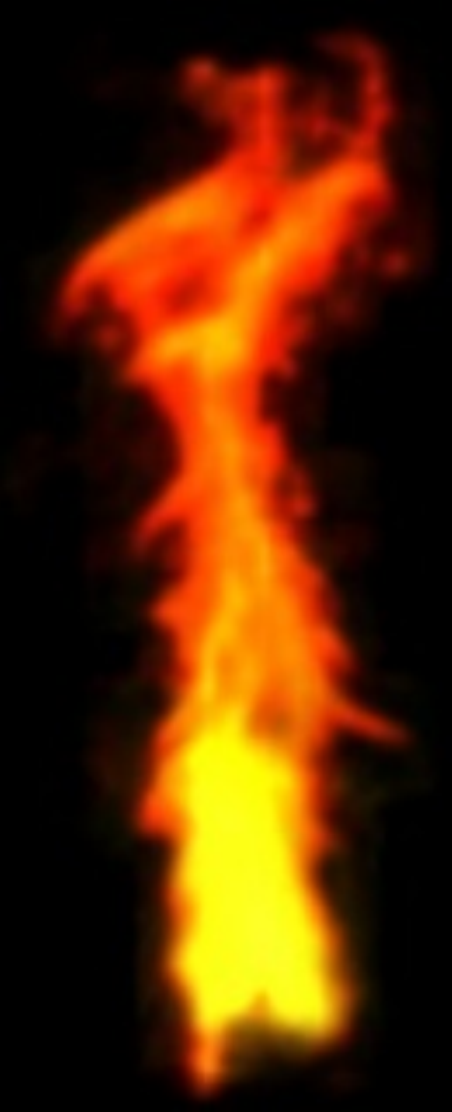}&
\includegraphics[width=0.09\textwidth]{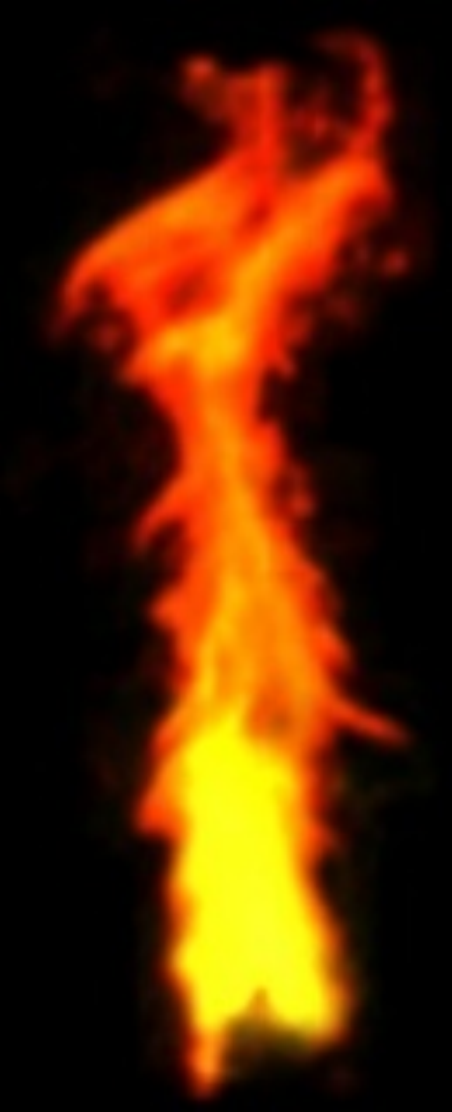}\\
(a) & (b) & (c) & (d)\\
\includegraphics[width=0.09\textwidth]{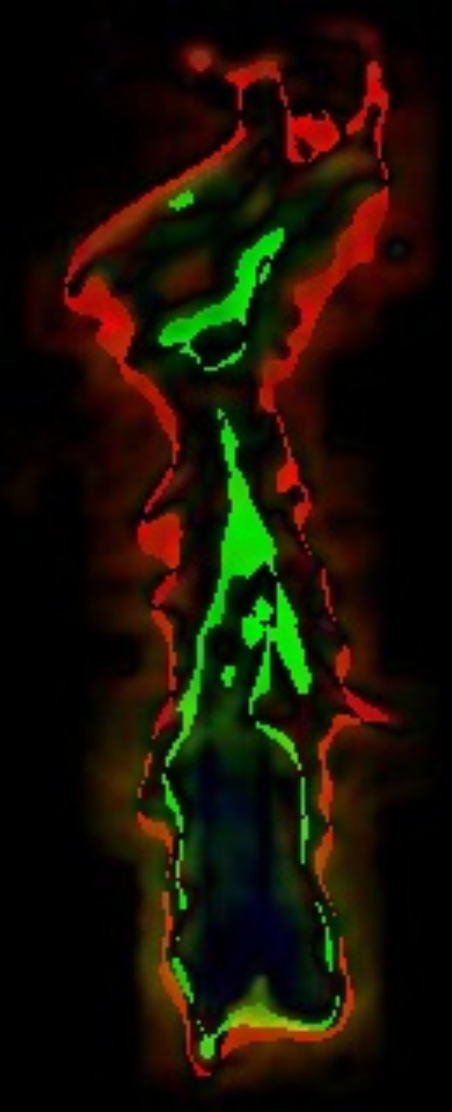}&
\includegraphics[width=0.09\textwidth]{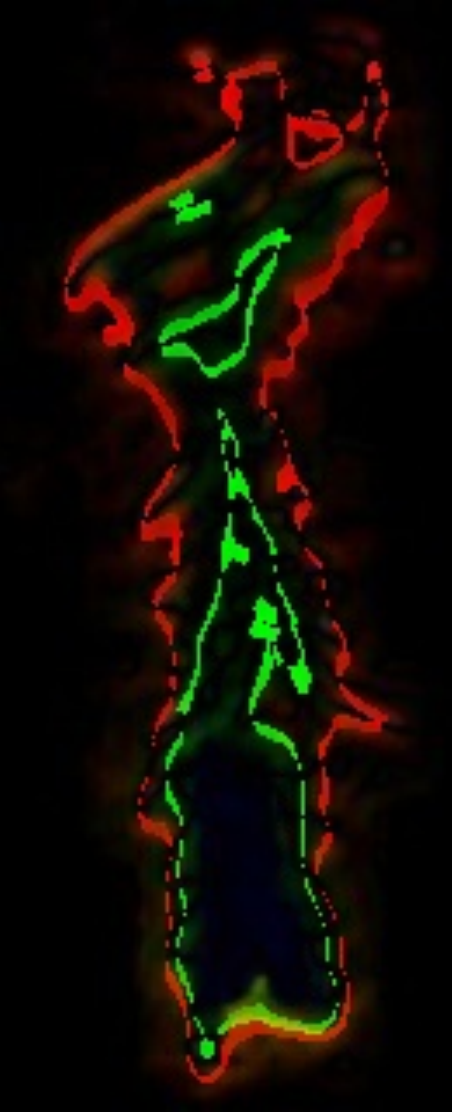}&
\includegraphics[width=0.09\textwidth]{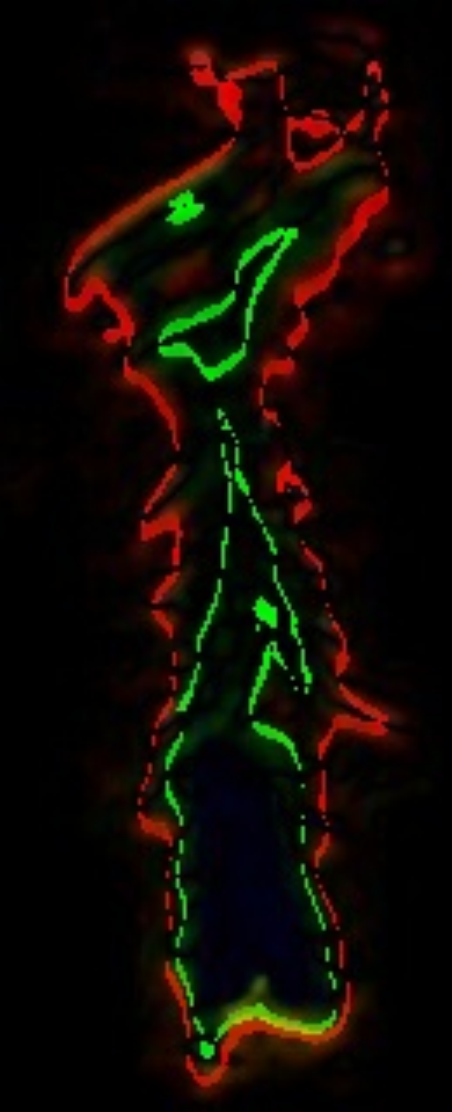}&
\includegraphics[width=0.09\textwidth]{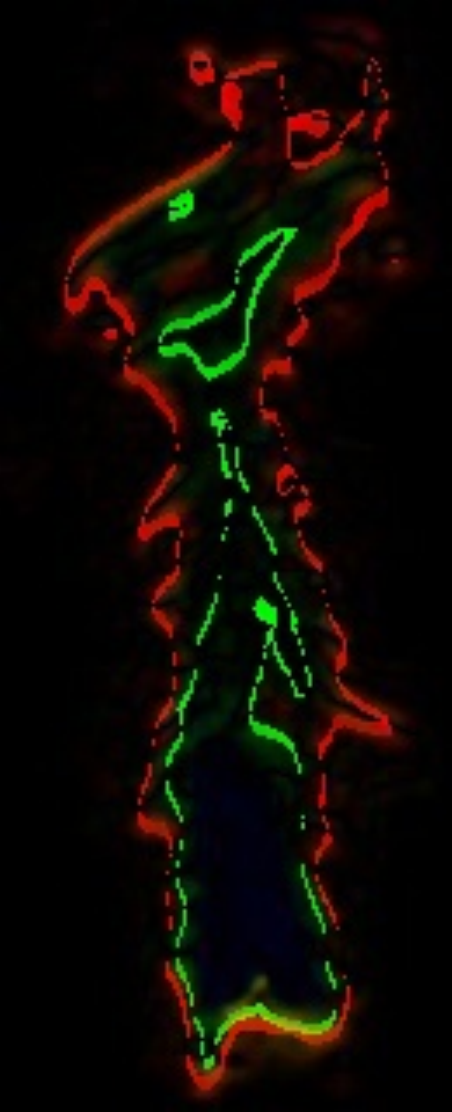}\\
(e) & (f) & (g) & (h)\\
\multicolumn{4}{c}{\includegraphics[width=0.45\textwidth]{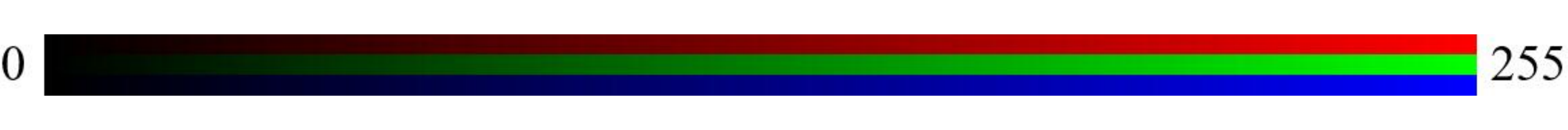}}
\end{tabular}
\end{center}
\vspace{-3mm}
\caption{Results of synthesizing the red, green, and blue channels after (a) 10, (b) 30, (c) 50, and (d) 70 iterations, and the corresponding errors (e), (f), (g) and (h). The errors for all channels are shown in a single image (such as (e), (f), (g) or (h)); the red, green and blue intensities represent the corresponding errors for the red, green and blue channels.
\label{fig_sim_color_iteration_img}}
\end{figure}

We analyze the pixel intensity RMSE and the volume data RMSE with time as shown in Fig. \ref{fig_result_sim_iteration}.
For each color channel, we visualize the results after each iteration with an average frame rate of 40.2 frames per second (fps).
Our method can obtain acceptable reconstruction results within two seconds. Fig. \ref{fig_sim_color_iteration_img} shows the visual results with the three channels synthesized after a series of iterations.

\begin{figure}[t]
\begin{center}
\setlength{\tabcolsep}{2pt}
\begin{tabular}{cccccc}
\includegraphics[width=0.078\textwidth]{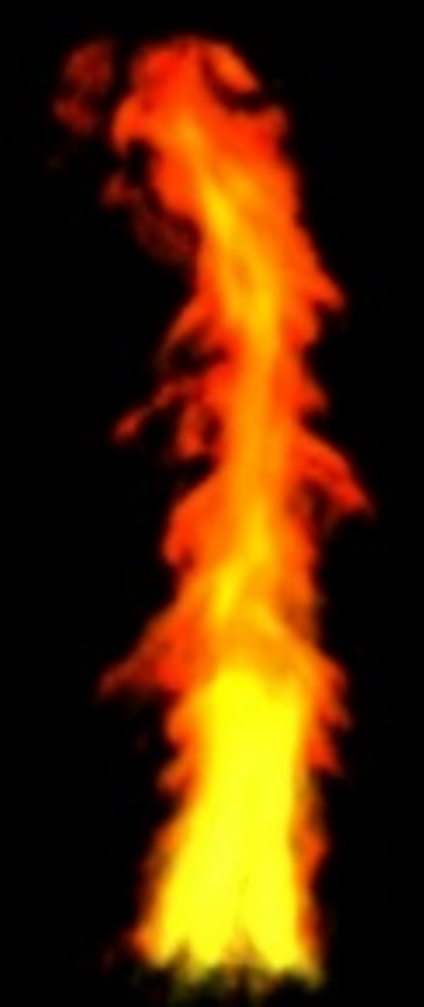}&
\includegraphics[width=0.078\textwidth]{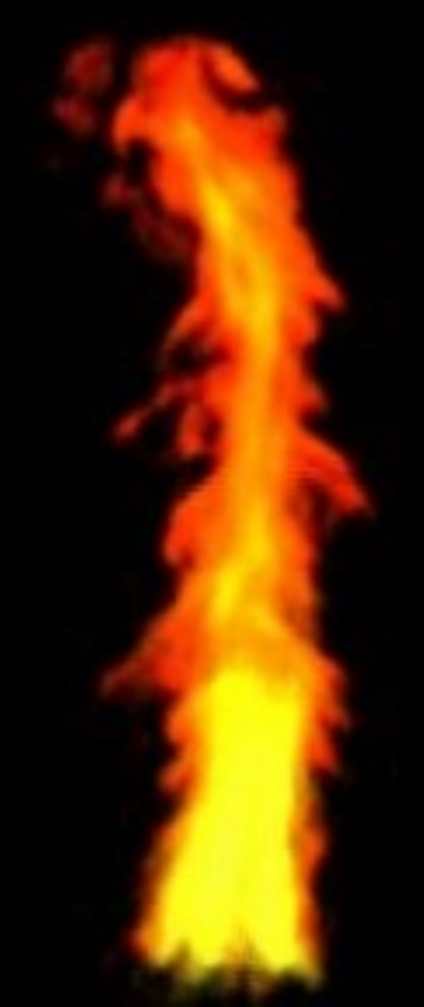}&
\includegraphics[width=0.078\textwidth]{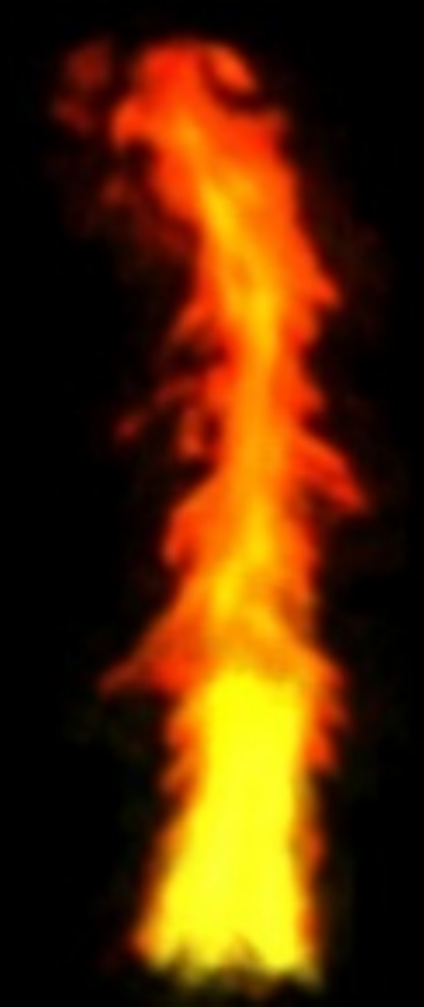}&
\includegraphics[width=0.0705\textwidth]{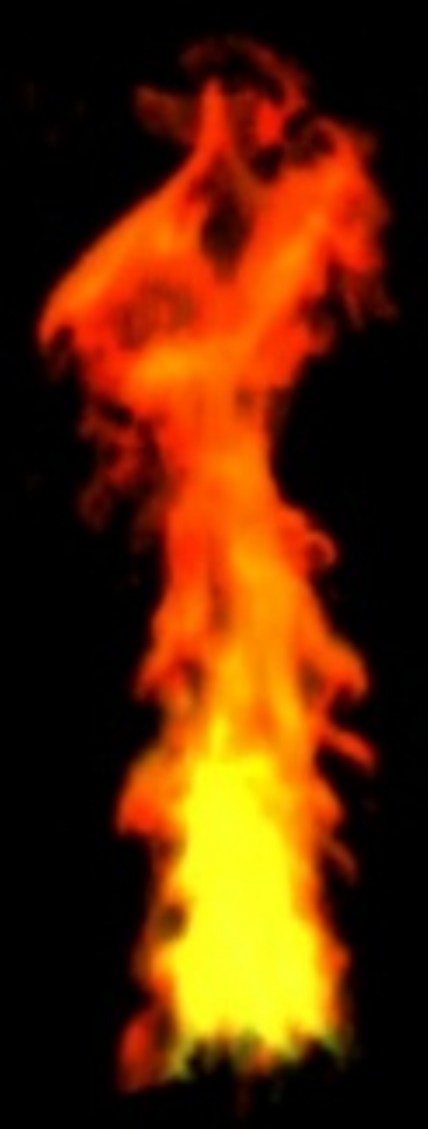}&
\includegraphics[width=0.0705\textwidth]{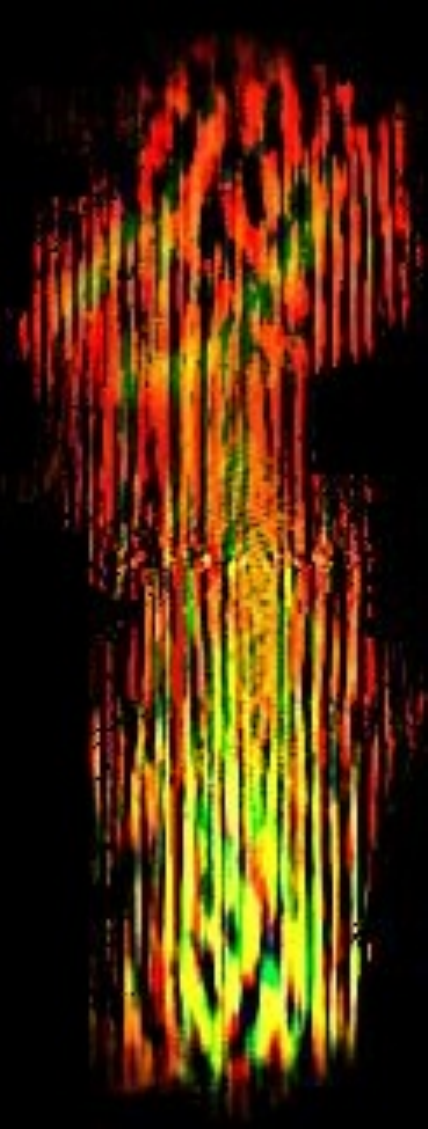}&
\includegraphics[width=0.0705\textwidth]{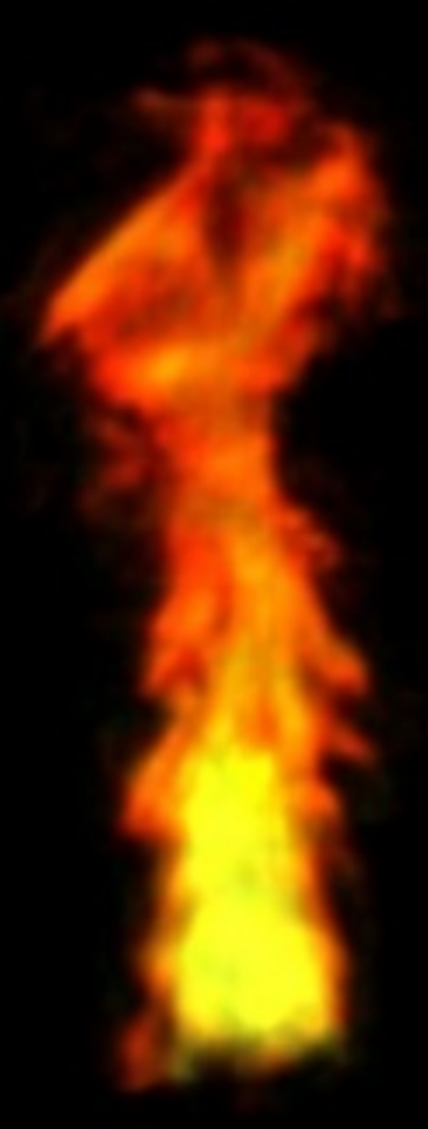}\\
(a) & (b) & (c) & (d) & (e) & (f)
\end{tabular}
\end{center}
\vspace{-3mm}
\caption{Extra evaluation. (a) One of the input images,
(b) and (c), reconstructed images by the algebraic tomography method and our method, respectively. (d) the ground truth of the test view,
(e) and (f), results of the algebraic and our method, respectively.
\label{fig_result_sim_extra}}
\end{figure}

To perform extra evaluation, we use nine views (input views) of the rendered images from the ten total views as input to reconstruct the flame.
After obtaining the reconstructed volume data, the image for the left view (test view) is rendered.
We compare the results generated by our method and the algebraic tomography method of Ihrke and Magnor \cite{ihrke2004image}.
Fig.~\ref{fig_result_sim_extra} and Table~\ref{table_err_of_extra_eva} show that our method achieves acceptable results for the input views, and for the test view, our approach achieves much better visual quality.
The scattering improves the smoothness of the reconstructed  results. Through the comparison, the superiority of the RTE model over the simplified linear optical imaging model is justified.

\begin{table}[t!]
\normalsize
\caption{Reconstruction errors of extra evaluation}
\label{table_err_of_extra_eva}
\vspace{-1mm}
\centering
\setlength{\tabcolsep}{4pt}
\begin{tabular}
{c|c|c|c|c|c|c}
\hline
\multirow{2}{*}{} & \multicolumn{3}{c|}{Input View} & \multicolumn{3}{c}{Test View} \\
\cline{2-7}
& Red & Green & Blue & Red & Green & Blue \\
\hline
Algebraic
& \multirow{2}{*}{4.2} & \multirow{2}{*}{2.3} & \multirow{2}{*}{3.0}
& \multirow{2}{*}{79.3} & \multirow{2}{*}{47.9} & \multirow{2}{*}{28.9} \\
Tomography & & & & & & \\
\hline
Our Method & 12.1 & 5.2 & 2.5 & 17.6 & 8.0 & 3.0 \\
\hline
\end{tabular}
\end{table}

\begin{figure}[t]
\begin{center}
\setlength{\tabcolsep}{1pt}
\begin{tabular}{cccccc}
\includegraphics[width=0.09\textwidth]{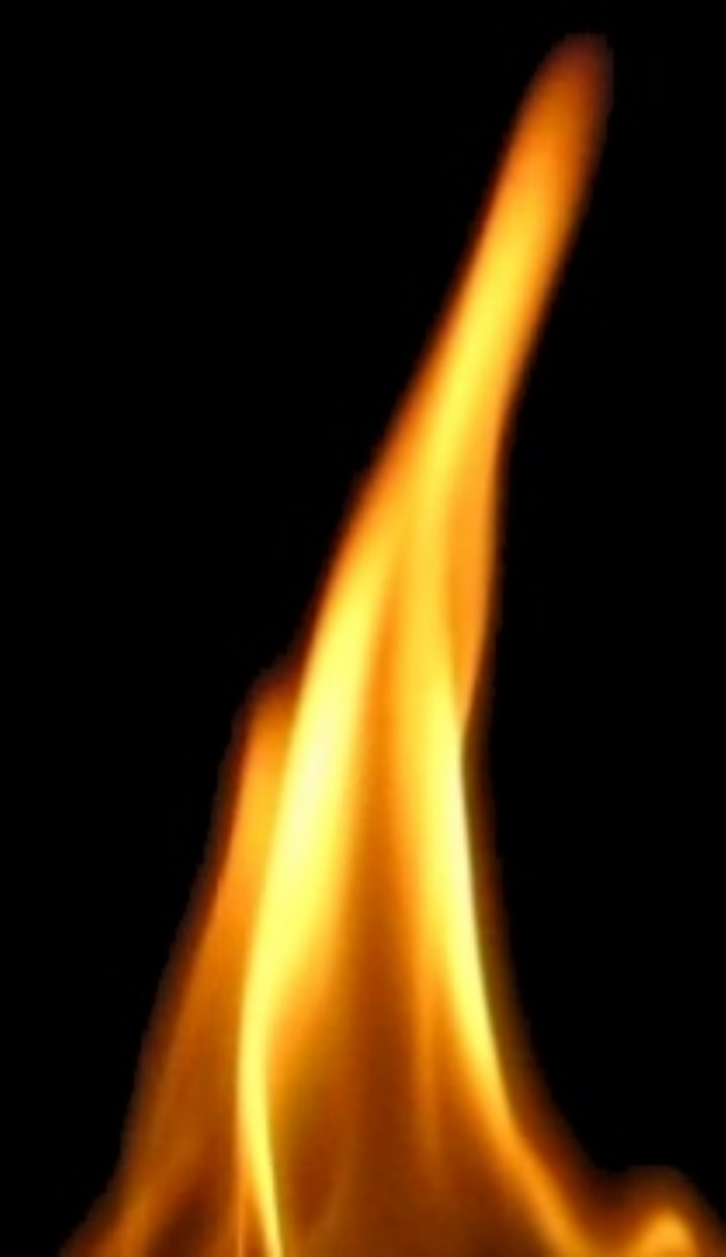}&
\includegraphics[width=0.09\textwidth]{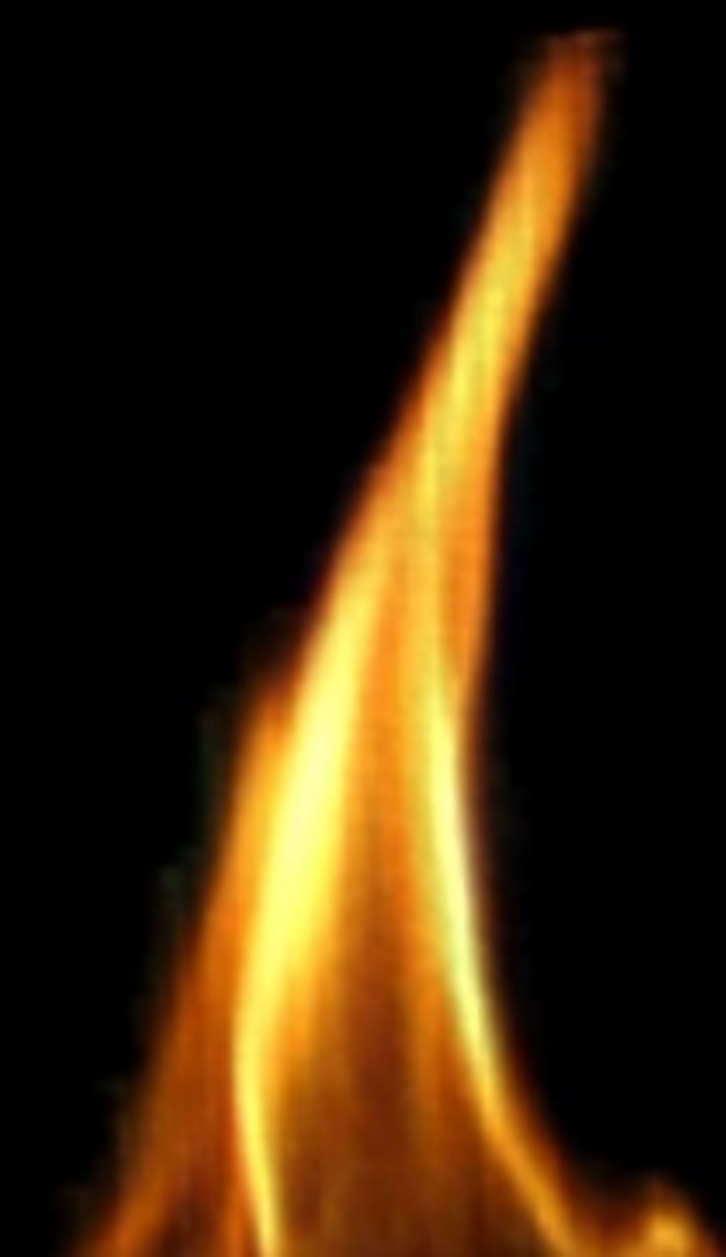}&
\includegraphics[width=0.0575\textwidth]{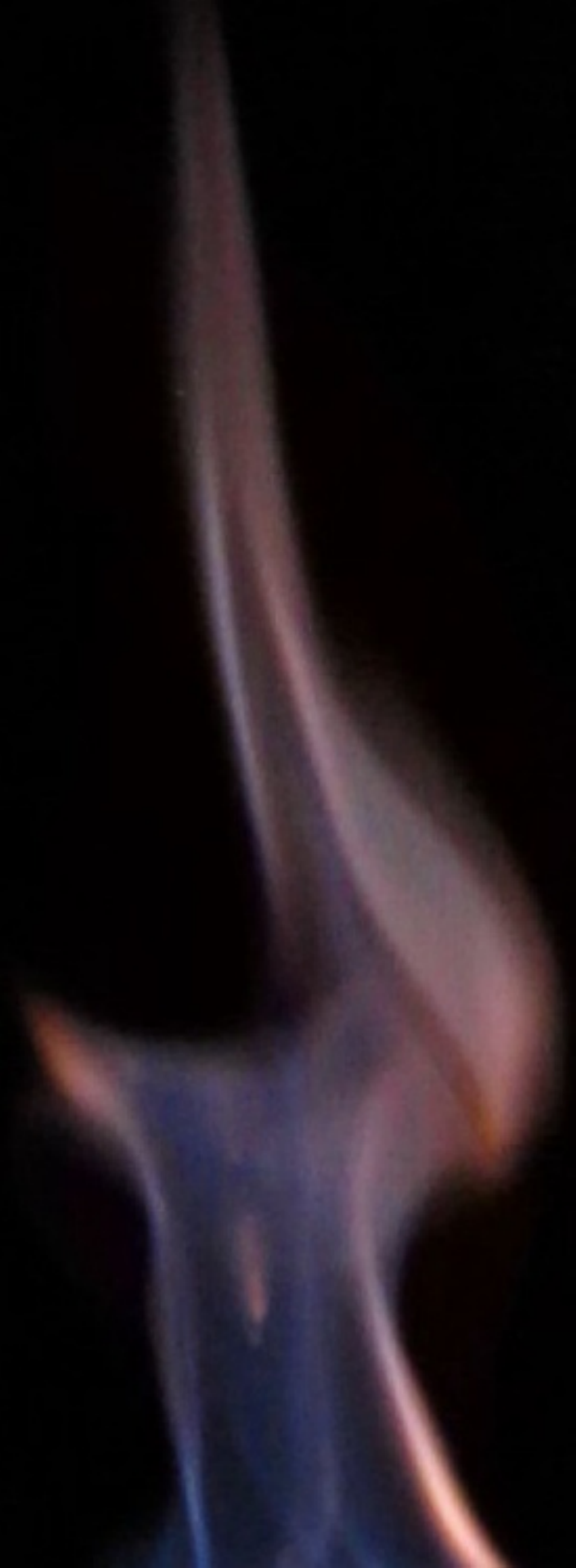}&
\includegraphics[width=0.0575\textwidth]{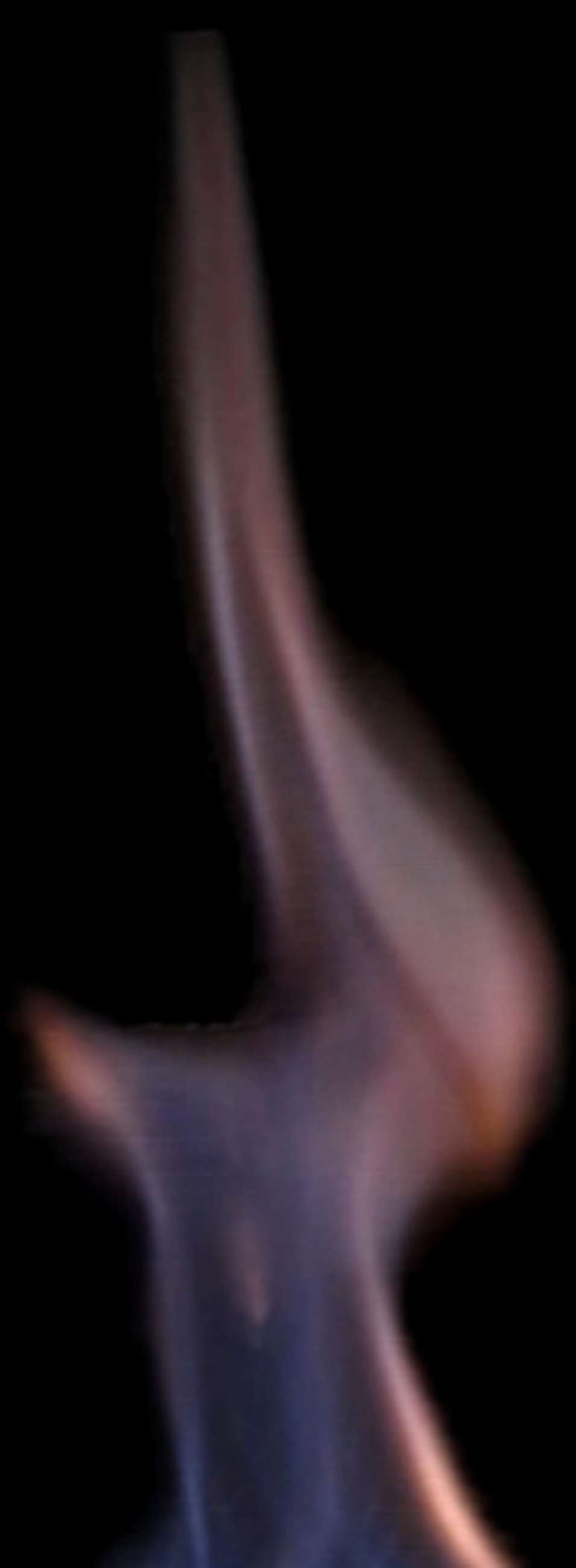}&
\includegraphics[width=0.079\textwidth]{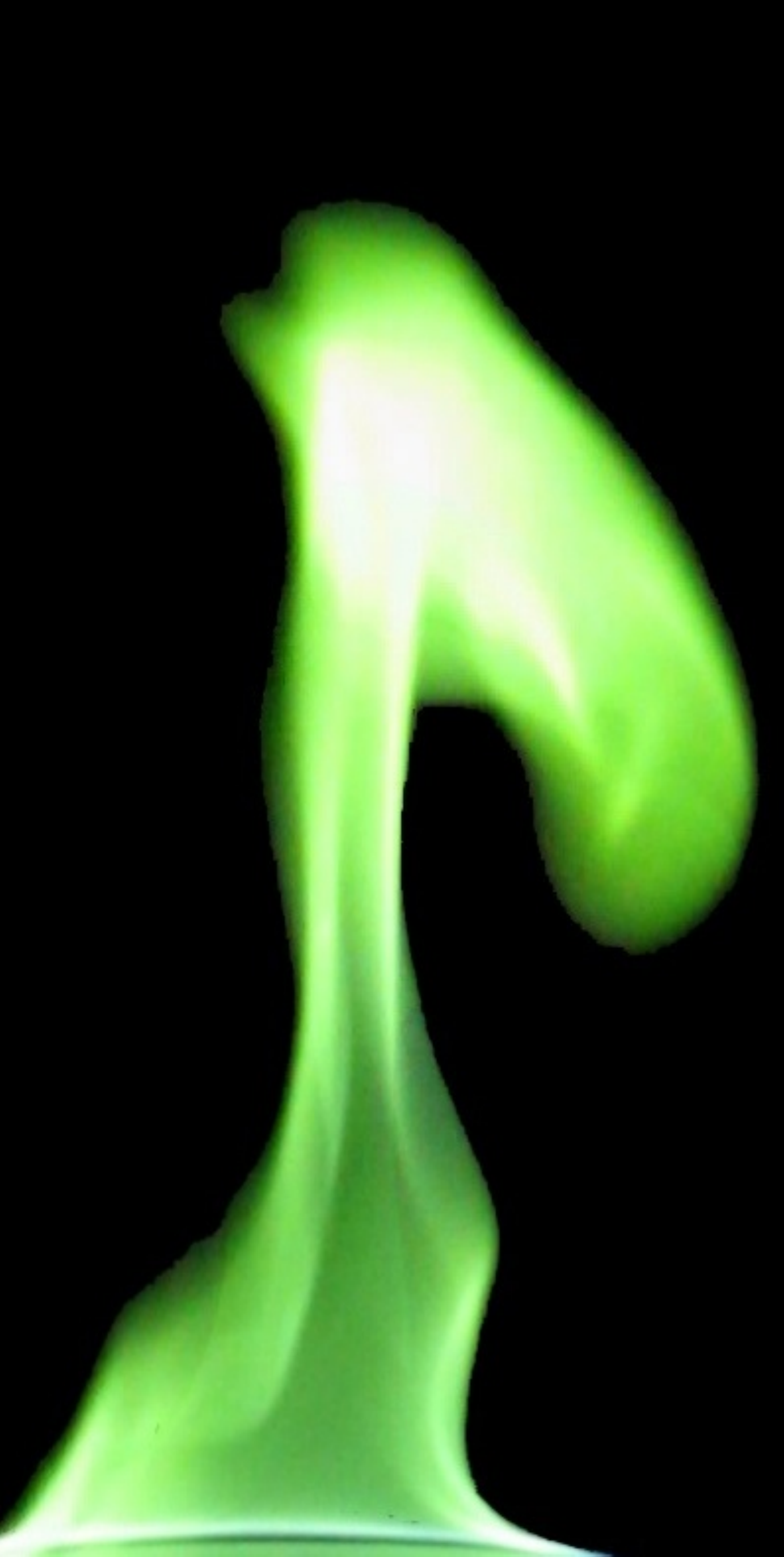}&
\includegraphics[width=0.079\textwidth]{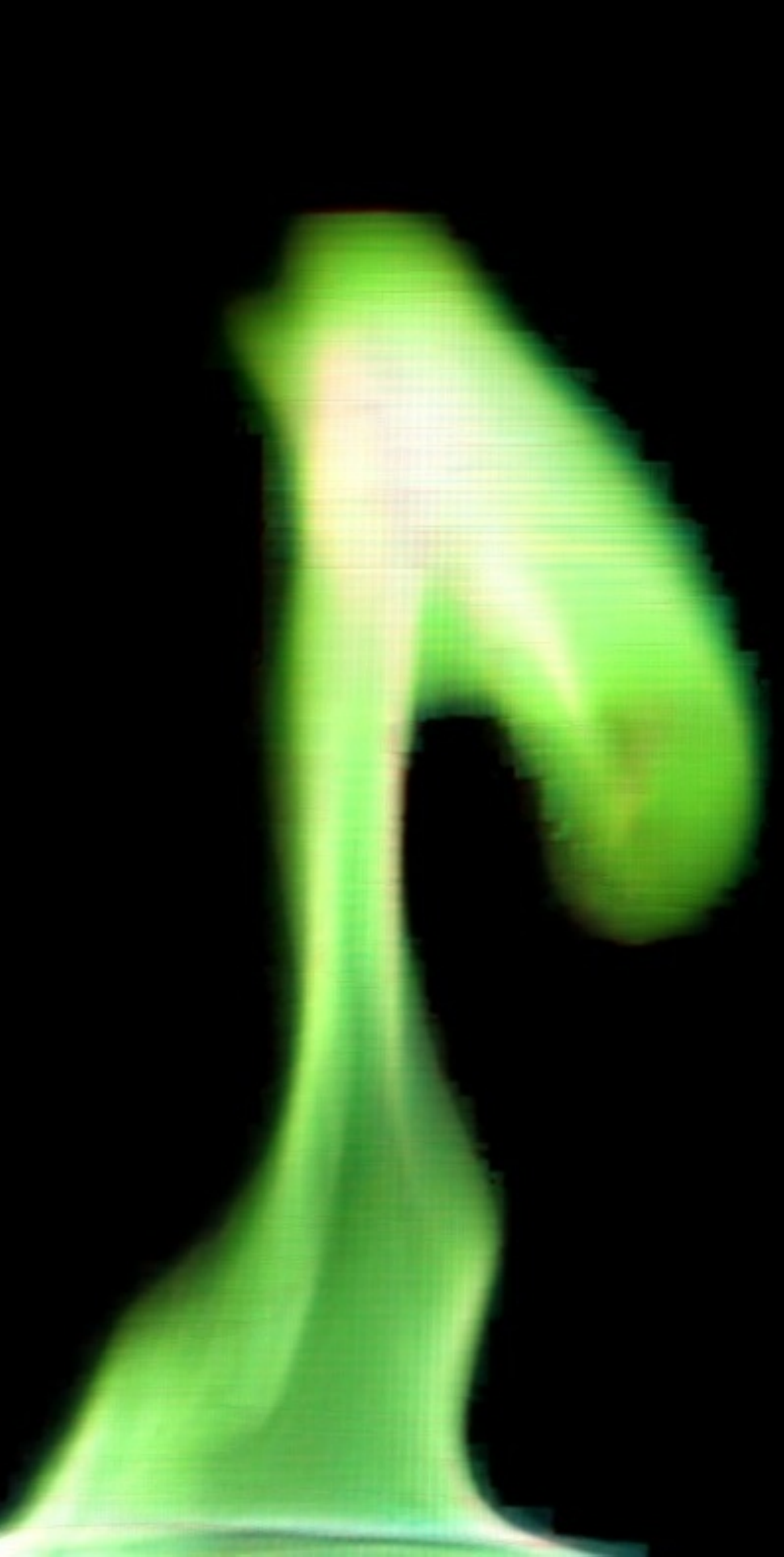}\\
(a) & (b) & (c) & (d) & (e) & (f)
\end{tabular}
\end{center}
\vspace{-3mm}
\caption{(a), (c) and (e): input images; (b), (d) and (f): corresponding reconstructed images.
(a) is generated by lighting paper fuel. (c) and (e) are generated by alcohol mixed with potassium chloride and boric acid, respectively.
\label{fig_result_capture_color}}
\vspace{-4mm}
\end{figure}

With the data captured from the setup described in Section \ref{subsec_DCP},
we use different fuels to generate different flames.
We reconstructed these flames in a 152mm$\times$152mm$\times$152mm volume.
Fig.~\ref{fig_result_capture_color} illustrates the reconstructed results for the captured flames.

\subsection{Temperature Reconstruction}
\begin{figure}[t]
\begin{center}
\setlength{\tabcolsep}{1pt}
\begin{tabular}{cccccccccc}
\includegraphics[width=0.065\textwidth]{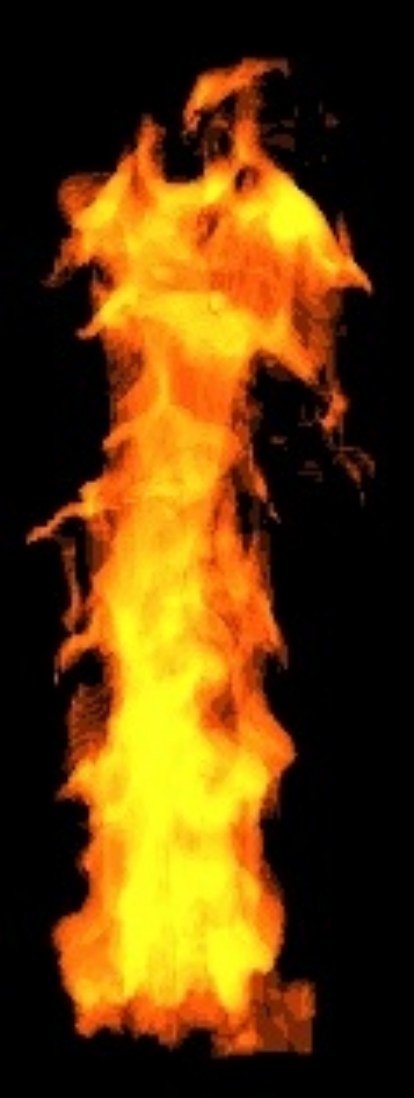}&
\includegraphics[width=0.065\textwidth]{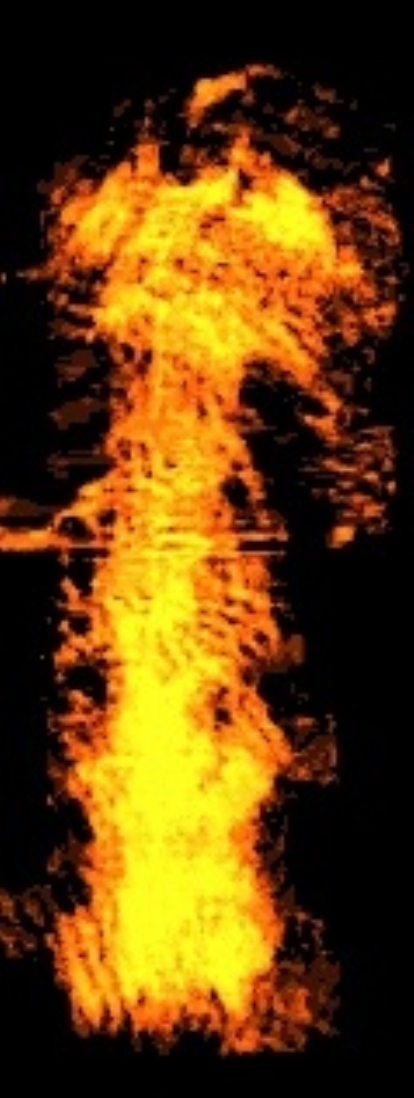}&
\includegraphics[width=0.065\textwidth]{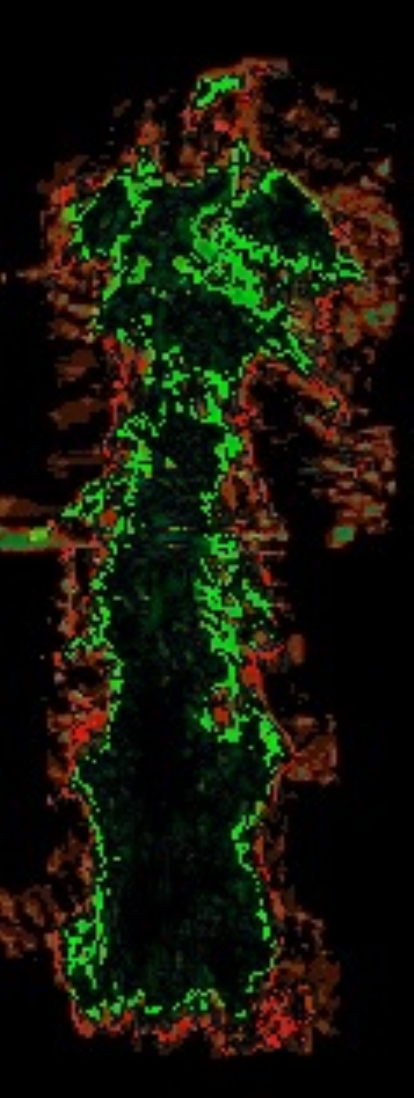}&
\includegraphics[width=0.065\textwidth]{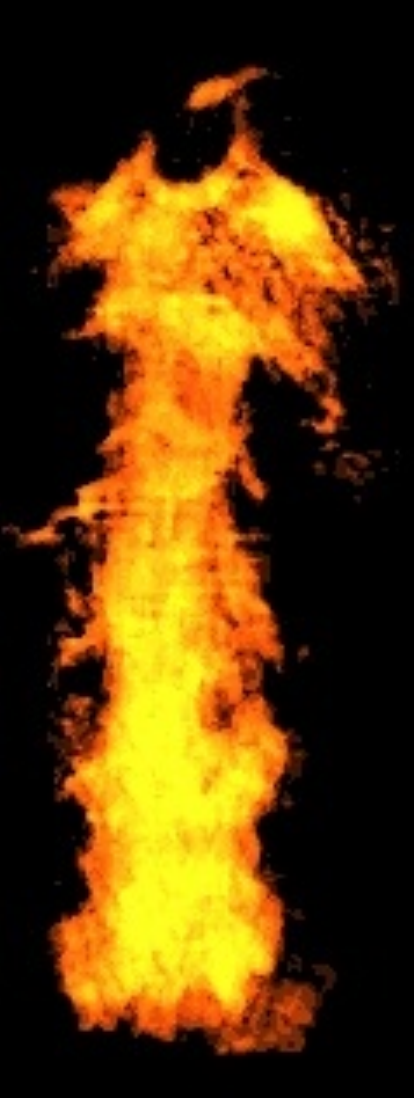}&
\includegraphics[width=0.065\textwidth]{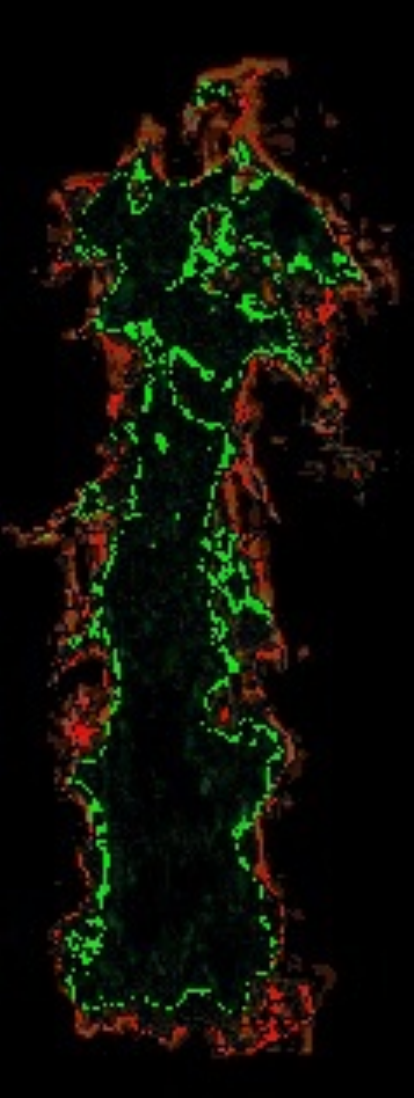}&
\includegraphics[width=0.065\textwidth]{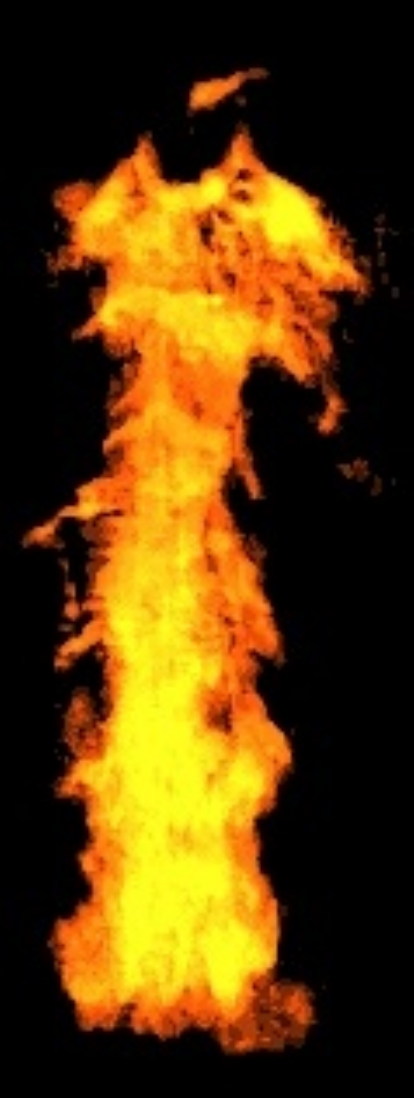}&
\includegraphics[width=0.065\textwidth]{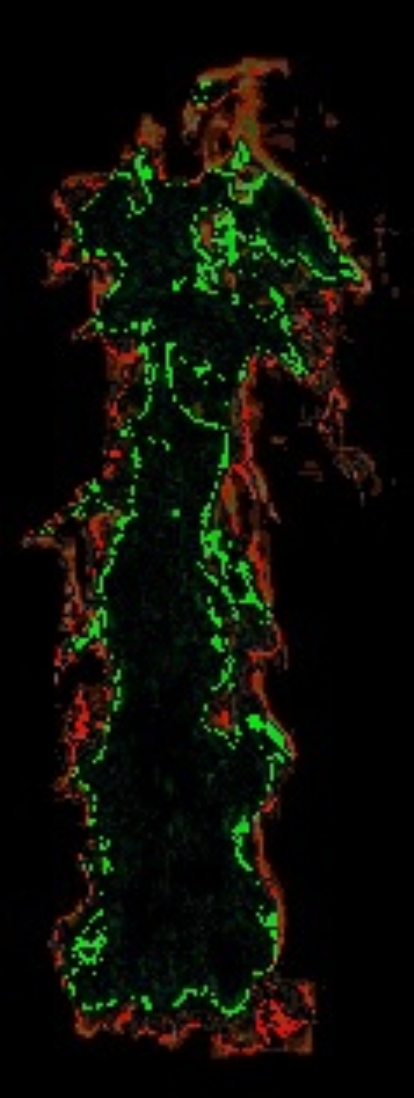} \\
(a) & (b) & (c) & (d) & (e) & (f) & (g)\\
\multicolumn{7}{c}{\includegraphics[width=0.48\textwidth]{figures/results/bar}}
\end{tabular}
\end{center}
\vspace{-3mm}
\caption{Reconstructed results for simulated data using different numbers of input cameras. (a) The simulated ground truth image, the reconstructed results with (b) 4, (d) 16, (f) 28 input cameras, and the corresponding errors when using (c) 4, (e) 16, (g) 28 input cameras.
\label{fig_different_num_camera}}
\end{figure}

\begin{figure}[h]
\begin{center}
\begin{tabular}{c}
\includegraphics[width=0.4\textwidth]{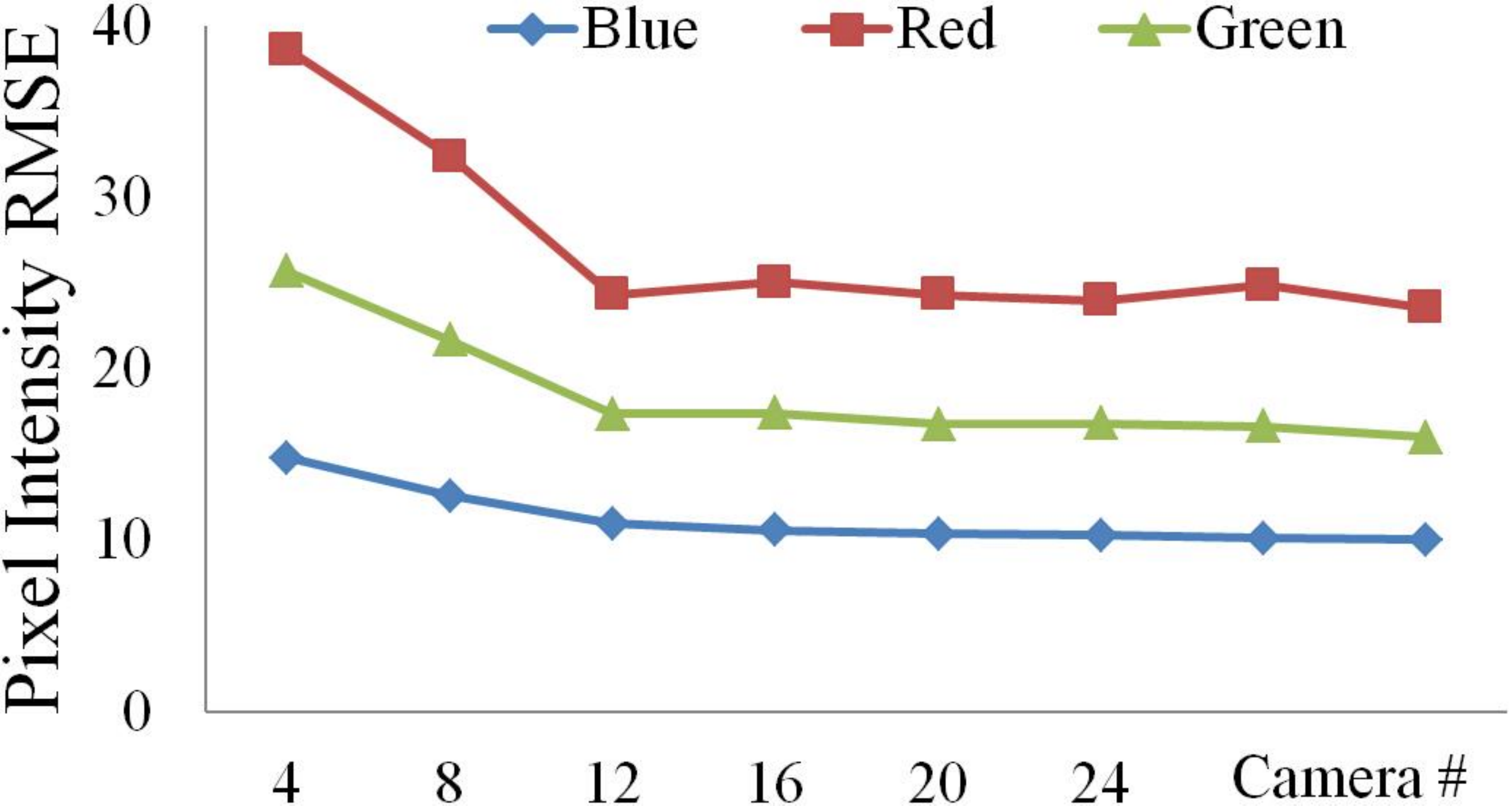} \\
(a)\vspace{1mm}\\
\includegraphics[width=0.4\textwidth]{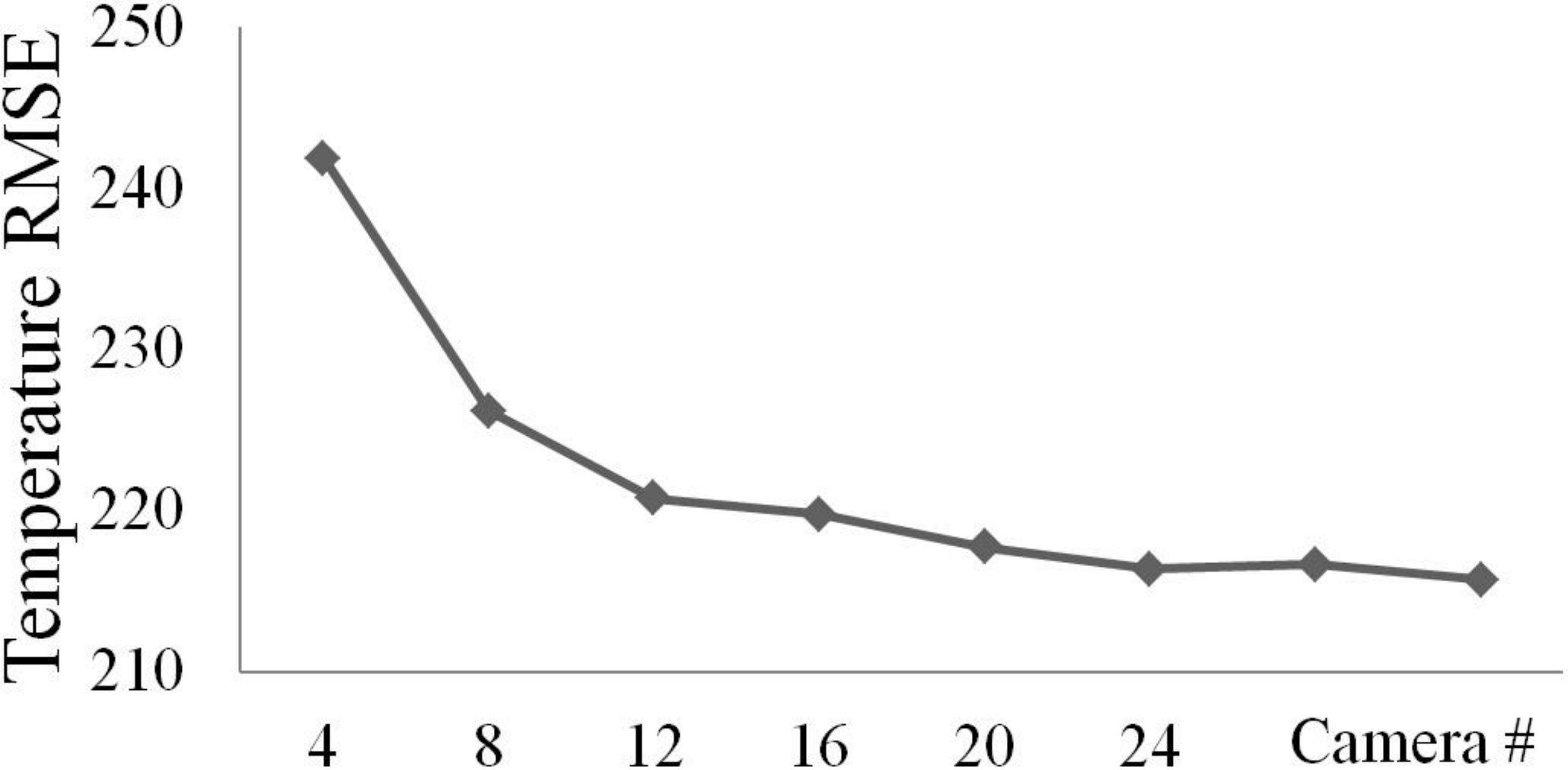}\\
(b)\\
\end{tabular}
\end{center}
\vspace{-3mm}
\caption{Reconstruction errors with different numbers of cameras.
(a) The intensity errors between Fig. \ref{fig_different_num_camera}(a) and reconstructed images with different numbers of input views.
(b) The temperature errors between the ground truth and reconstructed result.
\label{fig_camera_error}}
\end{figure}

Since the temperature of radiators with observable color
(minimum visible red) is more than 800K \cite{pardo2012influence},
and the highest temperature achieved by burning our current fuels is limited,
the temperature is assumed to be in the range (1000K, 2300K).
Using simulated data, we analyze the reconstruction results achieved from different numbers of input cameras.
The difference between the input image and the reconstructed image is computed as the pixel intensity error
for the red, green, and blue channels, as shown in Figs. \ref{fig_different_num_camera} and \ref{fig_camera_error}(a).
The difference in terms of the temperature field between the ground truth and the reconstructed result
is computed as temperature field errors, as shown in Fig. \ref{fig_camera_error}(b).
As the number of input cameras increases, the errors decrease.
The increase in the number of cameras leads to a tighter visual hull, which means that the region to be reconstructed becomes smaller.

\begin{figure}[t]
\begin{center}
\setlength{\tabcolsep}{2pt}
\begin{tabular}{ccc}
\includegraphics[width=0.112\textwidth]{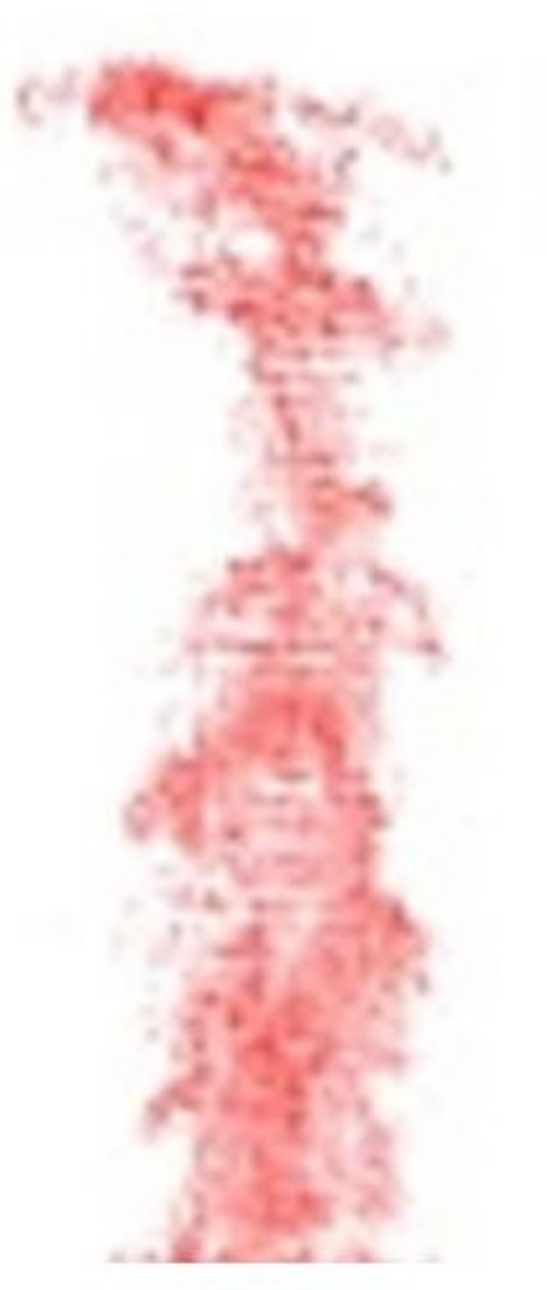}&
\includegraphics[width=0.185\textwidth]{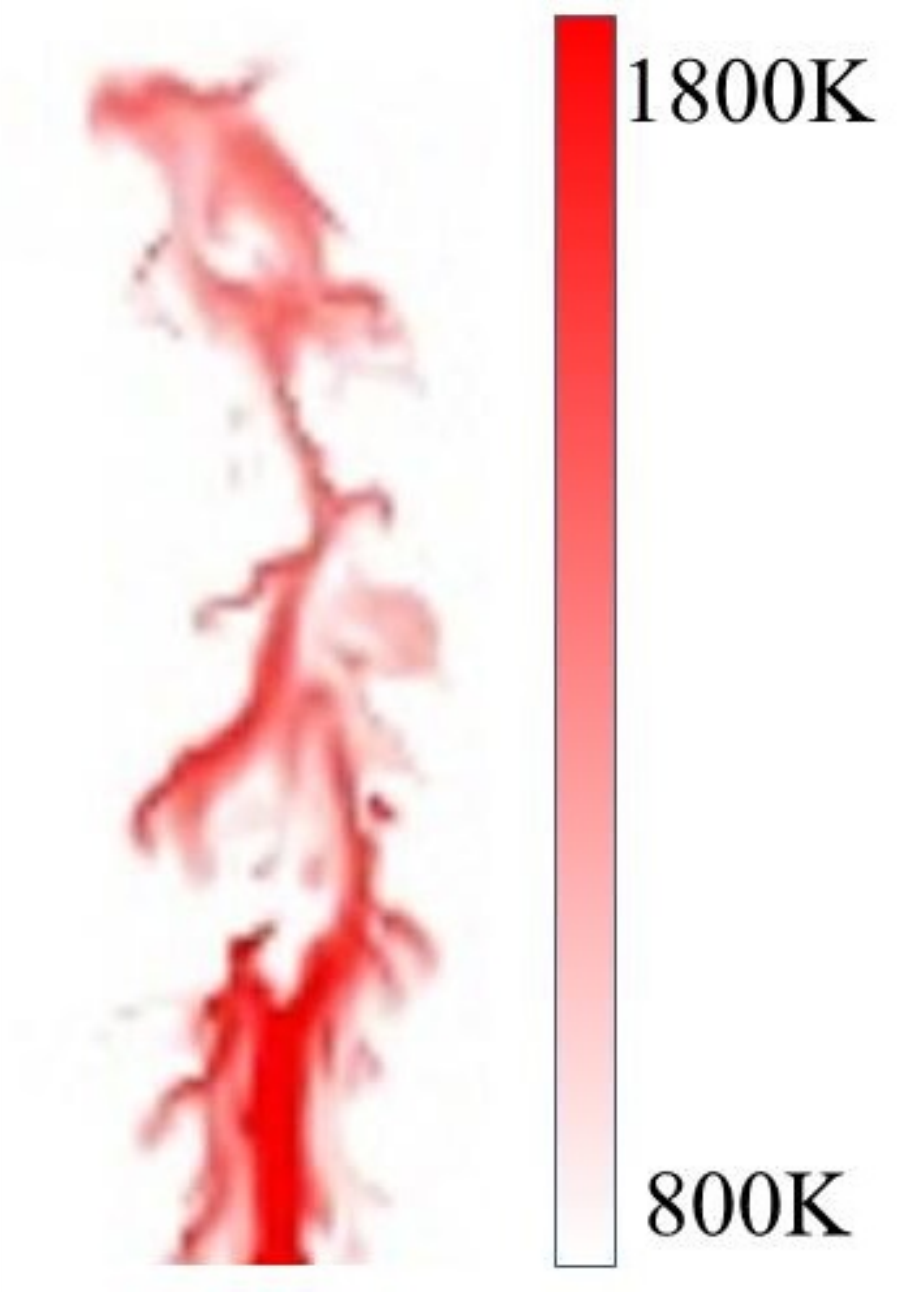}&
\includegraphics[width=0.16\textwidth]{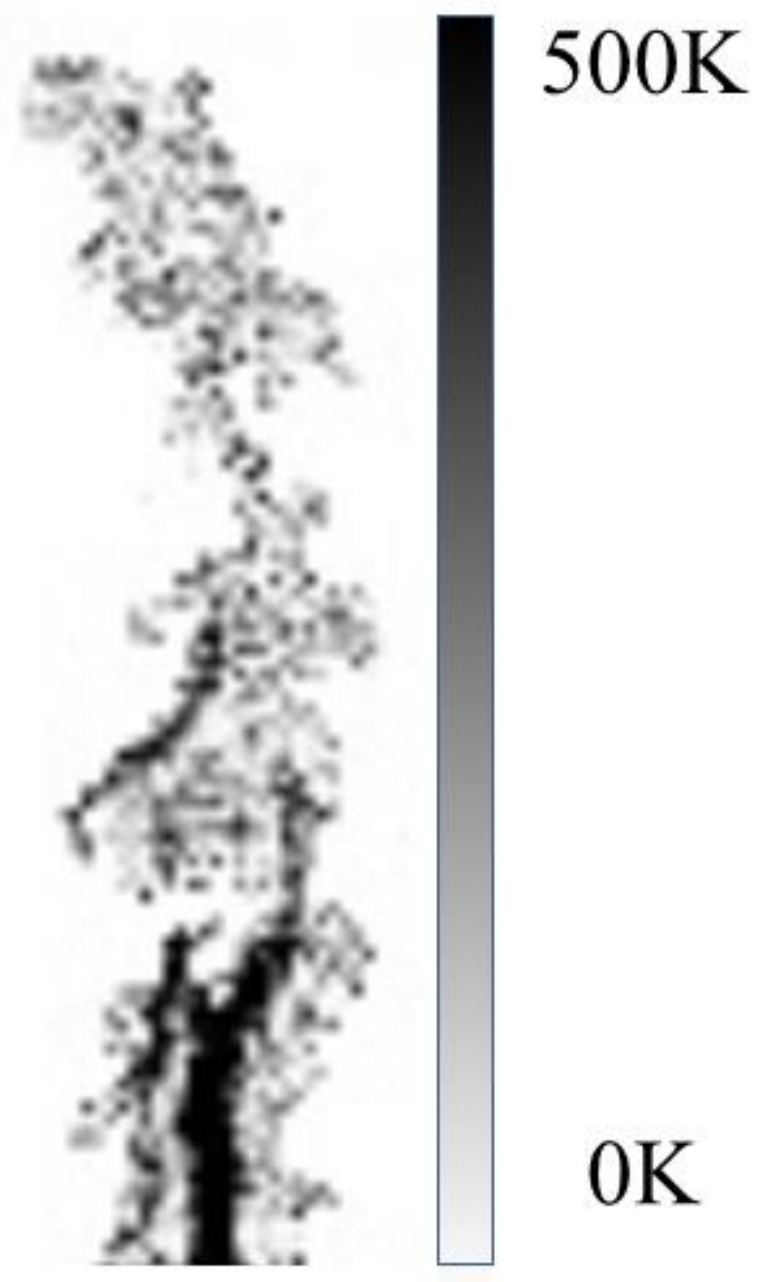}\\
(a) & (b) & (c)
\end{tabular}
\end{center}
\vspace{-3mm}
\caption{Cross section of reconstructed result. (a) The temperature values of one cross section plane in the reconstructed volume, (b) the ground truth,
and (c) the error plot between them.
\label{fig_cross_sec}}
\vspace{-5mm}
\end{figure}

With sixteen input views, the result of one cross section of the volume data is shown in Fig. \ref{fig_cross_sec}.
Without any constraints from fluid dynamics, the temperature values in the cross section seem coarse.
However, we can still obtain visually plausible flame reconstruction results, as shown in Fig. \ref{fig_different_num_camera}(d).

For the captured flames, the black-body radiation based temperature reconstruction can only be applied on the flames with yellow colors.
The reconstruction results are illustrated in Fig. \ref{fig_capturedata}. As shown in Table~\ref{table_sim_cap},
the pixel intensity errors for the captured data are higher than the errors for the simulated data.
The reasons for this include the accuracy of camera calibration, the limitation of the black-body radiation based color-temperature mapping,
and the limited number of input cameras.

\begin{table}[h]
\normalsize
\caption{Color intensity errors of captured data.}
\vspace{-1mm}
\label{table_sim_cap}
\centering
\setlength{\tabcolsep}{4pt}
\begin{tabular}
{c|c|c|c}
\hline
\diagbox{Data Type}{Channel} & Red & Green & Blue \\
\hline
Simulated & 25.2 & 15.2 & 6.5 \\
\hline
Captured & 25.8 & 18.6 & 23.9 \\
\hline
\end{tabular}
\end{table}

\begin{figure}[t]
\begin{center}
\setlength{\tabcolsep}{2pt}
\begin{tabular}{cccccc}
\small{Captured} & \small{Reconstructed} & \small{Interpolated} & \small{Reconstructed} & \small{Captured}\\
\includegraphics[width=0.08\textwidth]{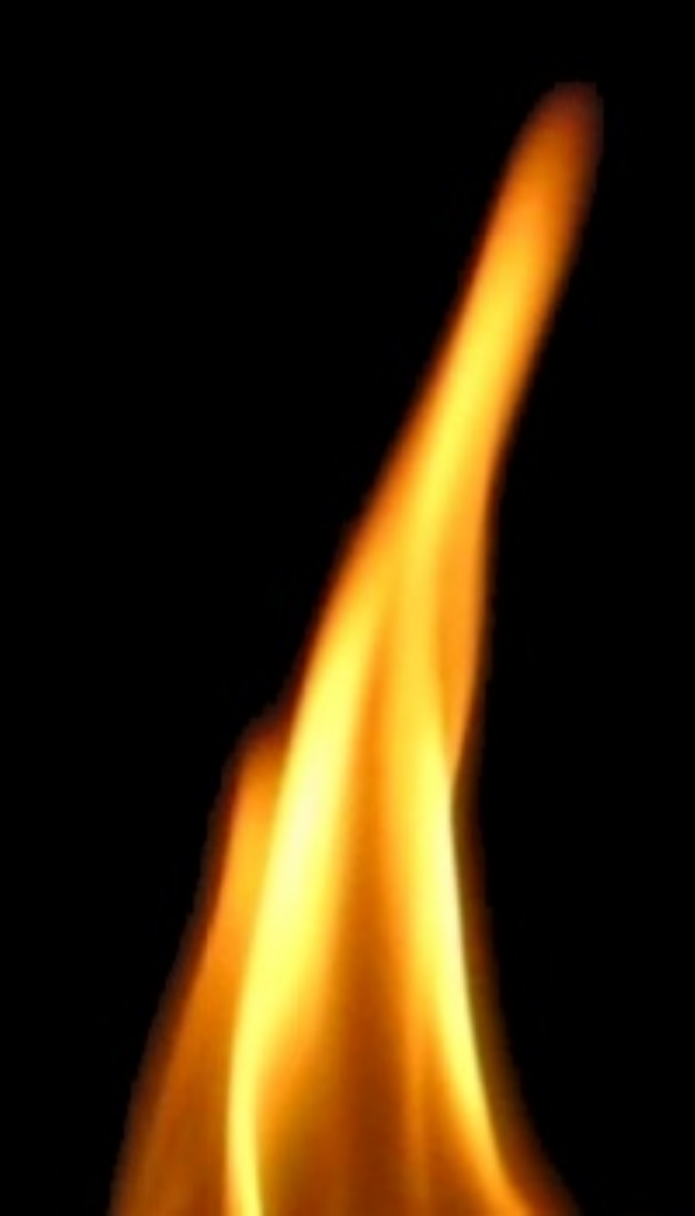}&
\includegraphics[width=0.08\textwidth]{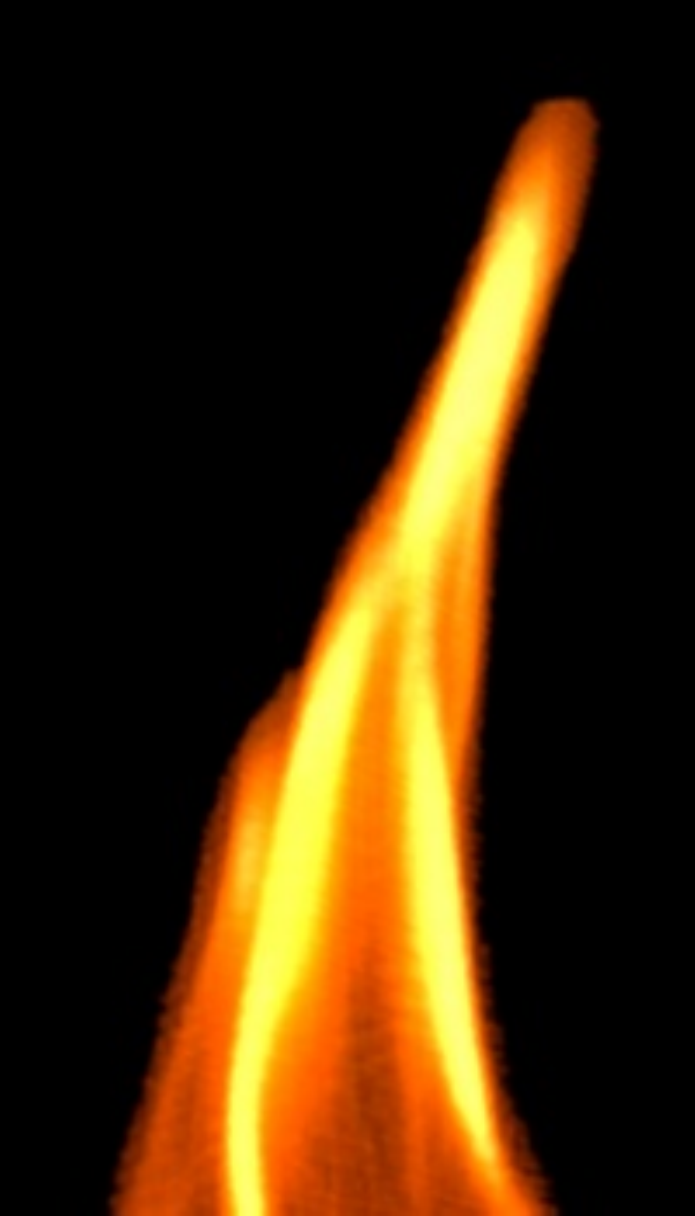}&
\includegraphics[width=0.08\textwidth]{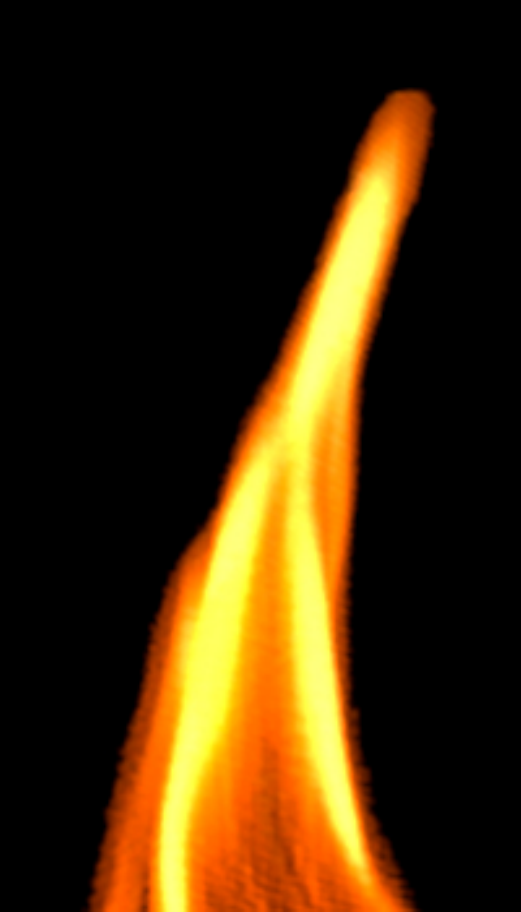}&
\includegraphics[width=0.075\textwidth]{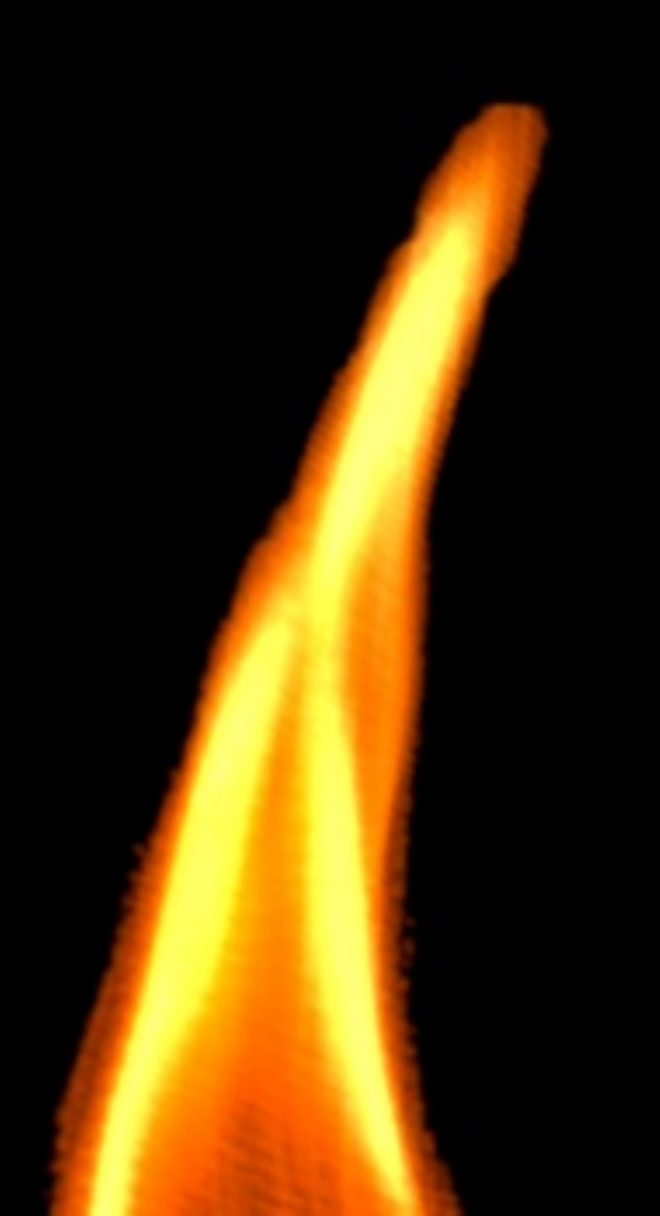}&
\includegraphics[width=0.075\textwidth]{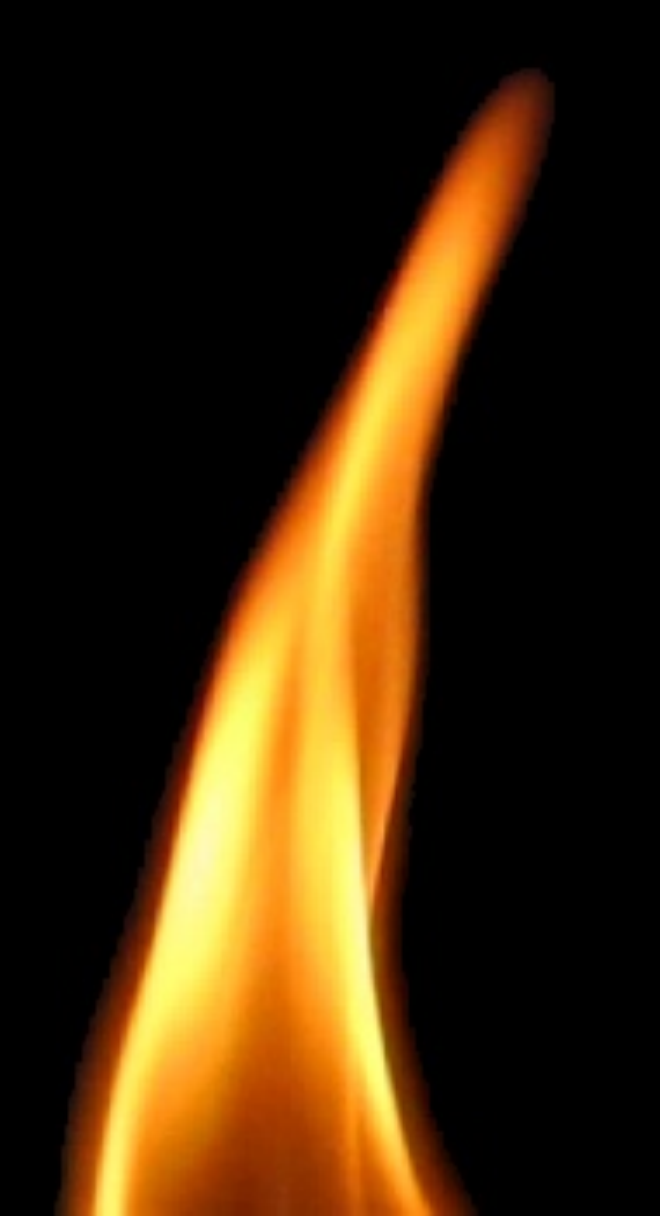}\\
(a) & (b) & (c) & (d) & (e)
\end{tabular}
\end{center}
\vspace{-3mm}
\caption{Reconstructed results from captured images. (a) and (e) are the captured frames, and (b) and (d) are the corresponding reconstructed images. (c) is the intermediate reconstructed view between (b) and (d).
\label{fig_capturedata}}
\vspace{-1mm}
\end{figure}

\begin{figure}[!h]
\begin{center}
\begin{tabular}{ccccc}
\multirow{2}{*}{Input} & Our & \multirow{2}{*}{Error} & Wu et al. & \multirow{2}{*}{Error}\\
& Approach & & ~\cite{wu2014reconstruction} & \\
\includegraphics[width=0.07\textwidth]{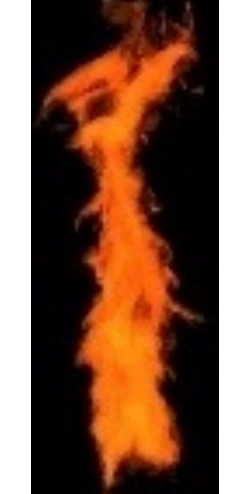}&
\includegraphics[width=0.07\textwidth]{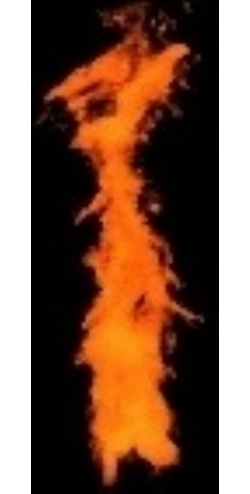}&
\includegraphics[width=0.07\textwidth]{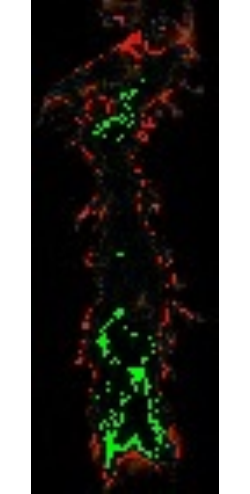}&
\includegraphics[width=0.07\textwidth]{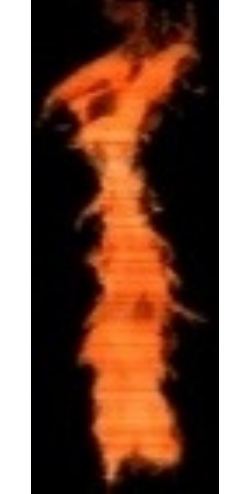}&
\includegraphics[width=0.07\textwidth]{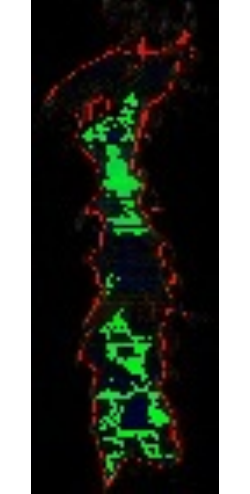}\\
(a) & (b) & (c) & (d) & (e)\\
\includegraphics[width=0.07\textwidth]{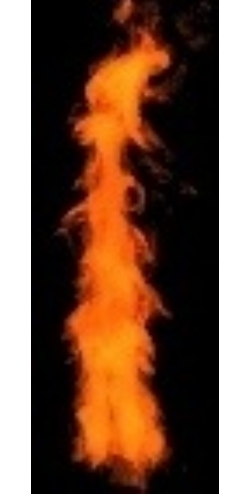}&
\includegraphics[width=0.07\textwidth]{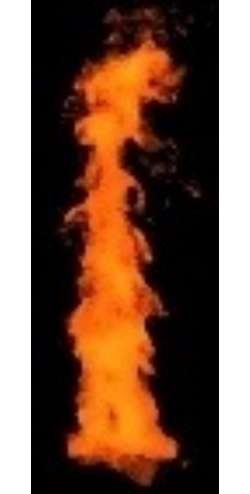}&
\includegraphics[width=0.07\textwidth]{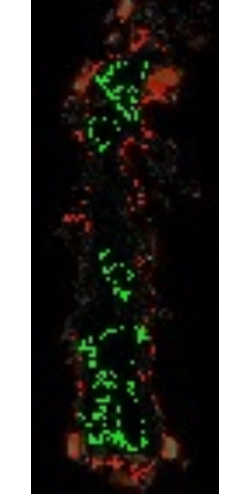}&
\includegraphics[width=0.07\textwidth]{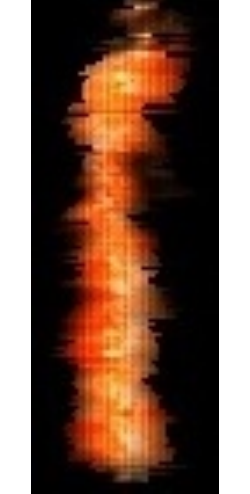}&
\includegraphics[width=0.07\textwidth]{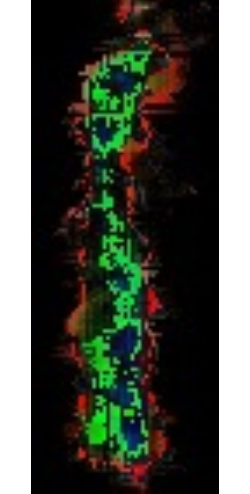}\\
(f) & (g) & (h) & (i) & (j)\\
\multicolumn{5}{c}{\includegraphics[width=0.48\textwidth]{figures/results/bar}}
\end{tabular}
\end{center}
\vspace{-3mm}
\caption{Comparison with Wu et al.~\cite{wu2014reconstruction}. Similar to their method, we use four images as input, corresponding to $0^\circ$, $45^\circ$, $90^\circ$, and $135^\circ$ angle views. Here we show the (a) $0^\circ$ and (f) $45^\circ$ angle input views. The corresponding reconstruction results for their method are shown in (d) and (i), and the respective errors are shown in (e) and (j). The results for our method are shown in (b) and (g) with the respective errors shown in (c) and (h).
\label{fig_comp_2014_paper}}
\end{figure}

\captionsetup[table]{justification=raggedright}
\begin{table*}[!ht]
\normalsize
\caption{Comparison of different methods for flame reconstruction.}
\vspace{-1mm}
\centering
\label{table_time_cost}
\centering
\resizebox{\textwidth}{!}{
\begin{tabular}
{c|c|c|c|c|c|c|c}
\hline
\multirow{2}{*}{Method} &Camera & In-plane
& Projection & Rendering
& \multirow{2}{*}{Iterative}  &GPU &\multirow{2}{*}{ Time(s)***}  \\
& Type & Constraint & Model & Model  & & Accelerated & \\
\hline
\hline
Flame-sheet
& Industrial & \multirow{2}{*}{Yes}
&\multirow{2}{*}{Parallel} &\multirow{2}{*}{ Linear}
&\multirow{2}{*}{No}  &\multirow{2}{*}{No} & \multirow{2}{*}{$>$60} \\
2003~\cite{hasinoff2003photo}, 2007~\cite{hasinoff2007photo}
&  CCD &  & &  & & & \\
\hline
Algebraic Tomography
& Industrial & \multirow{2}{*}{No}
&\multirow{2}{*}{Perspective} & \multirow{2}{*}{Linear}
&\multirow{2}{*}{Yes}  &\multirow{2}{*}{No} & \multirow{2}{*}{$>$60} \\
2004~\cite{ihrke2004image}
 & CCD & & & &  &  &  \\
\hline
Stochastic Tomography
&   Consumer  & \multirow{2}{*}{No}
&\multirow{2}{*}{Perspective} & Linear \&
&\multirow{2}{*}{Yes}   &\multirow{2}{*}{No} & \multirow{2}{*}{$>$60} \\
2012 \cite{gregson2012stochastic}
&  CMOS* & & &   Splatting&  &  & \\
\hline
Color Temperature
 & Industrial & \multirow{2}{*}{Yes}
&\multirow{2}{*}{Parallel} & \multirow{2}{*}{Linear}
&\multirow{2}{*}{No}  &\multirow{2}{*}{No} & \multirow{2}{*}{0.91} \\
2014 \cite{wu2014reconstruction}
& CCD & & & &  &  &  \\
\hline
Two-Color Pyrometric
& Industrial & \multirow{2}{*}{No}
&\multirow{2}{*}{Perspective} & \multirow{2}{*}{Linear}
&\multirow{2}{*}{No}  &\multirow{2}{*}{Yes} & \multirow{2}{*}{$>$10} \\
2015 \cite{zhou20153}
& CCD & & & &  &  & \\
\hline
Refractive Reconstruction
& Industrial& \multirow{2}{*}{No}
&\multirow{2}{*}{Perspective} & \multirow{2}{*}{Refractive}
&\multirow{2}{*}{Yes}   &\multirow{2}{*}{No} & \multirow{2}{*}{$>$60} \\
2015 \cite{wang2015image}
 & HDR CMOS & & & &  &   & \\
\hline
Appearance Transfer
&  Consumer & \multirow{2}{*}{Yes}
&\multirow{2}{*}{Parallel} & \multirow{2}{*}{Linear}
&\multirow{2}{*}{Yes}  &\multirow{2}{*}{No} & \multirow{2}{*}{$>$60} \\
2015 \cite{okabe2015fluid}
& CMOS** & & & &  &  &  \\
\hline
\hline
\textbf{Our}
& \textbf{Consumer}  & \multirow{2}{*}{\textbf{No}}
&\multirow{2}{*}{\textbf{Perspective}} & \textbf{Radiative}
&\multirow{2}{*}{\textbf{Yes}}   &\multirow{2}{*}{\textbf{Yes}} & \multirow{2}{*}{\textbf{0.56}} \\
\textbf{Method}
& \textbf{CCD} & & & \textbf{Transfer}  &  & & \\
\hline
\end{tabular}}
\caption*{\small{*The consumer CMOS cameras were used to capture mixing fluids \cite{gregson2012stochastic}, which does not require very precise synchronization.\\
**Only two cameras are needed in this method, so the synchronization can be achieved by trial and error. The response time of the same trigger signal may be different for each consumer camera. The purpose of this method is to create a fluid animation using the volume sequence modeled from a sparse set of videos as a reference. The precise reconstruction of the volume is not required.\\
***The time for reconstructing one frame of the video.}}
\vspace{-3mm}
\end{table*}

We also compare our method with the approach of Wu et al. \cite{wu2014reconstruction}.
We used four input images with a resolution of 200$\times$310 and viewing angles of $0^\circ$,  $45^\circ$,  $90^\circ$, and  $135^\circ$.
As shown in Fig. \ref{fig_comp_2014_paper}, the results from the Wu et al. approach for the $0^\circ$ angle view are rather poor.
For the  $45^\circ$ angle view, since their method uses the image from this view to generate the visual hull, and
no color information is exploited for the reconstruction, the reconstructed result is more noisy.
Even though we exploit both color and temperature information, we are still able to achieve better time efficiency than their method, as shown in Table \ref{table_time_cost}.

\subsection{Performance Analysis}

Table~\ref{table_time_cost} shows an overview of different methods that have been proposed for flame reconstruction.
Our method provides an economical way to capture flame data, and unlike other methods does not suffer from the in-plane constraint for the cameras.
Moreover, we exploit a more accurate rendering model in our reconstruction process. Finally, though our approach visualizes the reconstruction results after each iteration,
our method can still achieve real-time performance during the reconstruction process (approximately 40fps on our laptop).

\section{Applications}
\label{sec_applications}


Our proposed flame reconstruction method can be used in a variety of applications.

\subsection{Reconstructed flames in virtual environments}
Flame effects are widely used in movies and video games.
Physically-based flame simulations are time-consuming due to the use of computationally expensive solvers for the Navier Stokes equations.
Our approach provides an effective way for modeling a 3D flame from several input images, and the flame volume data can then be reused in virtual environments.
Since we capture real flame videos, the flames modeled by our method are more realistic than the simulations.

\begin{figure*}[!t]
\begin{center}
\setlength{\tabcolsep}{3pt}
\begin{tabular}{cccc}
\includegraphics[width=0.22\textwidth]{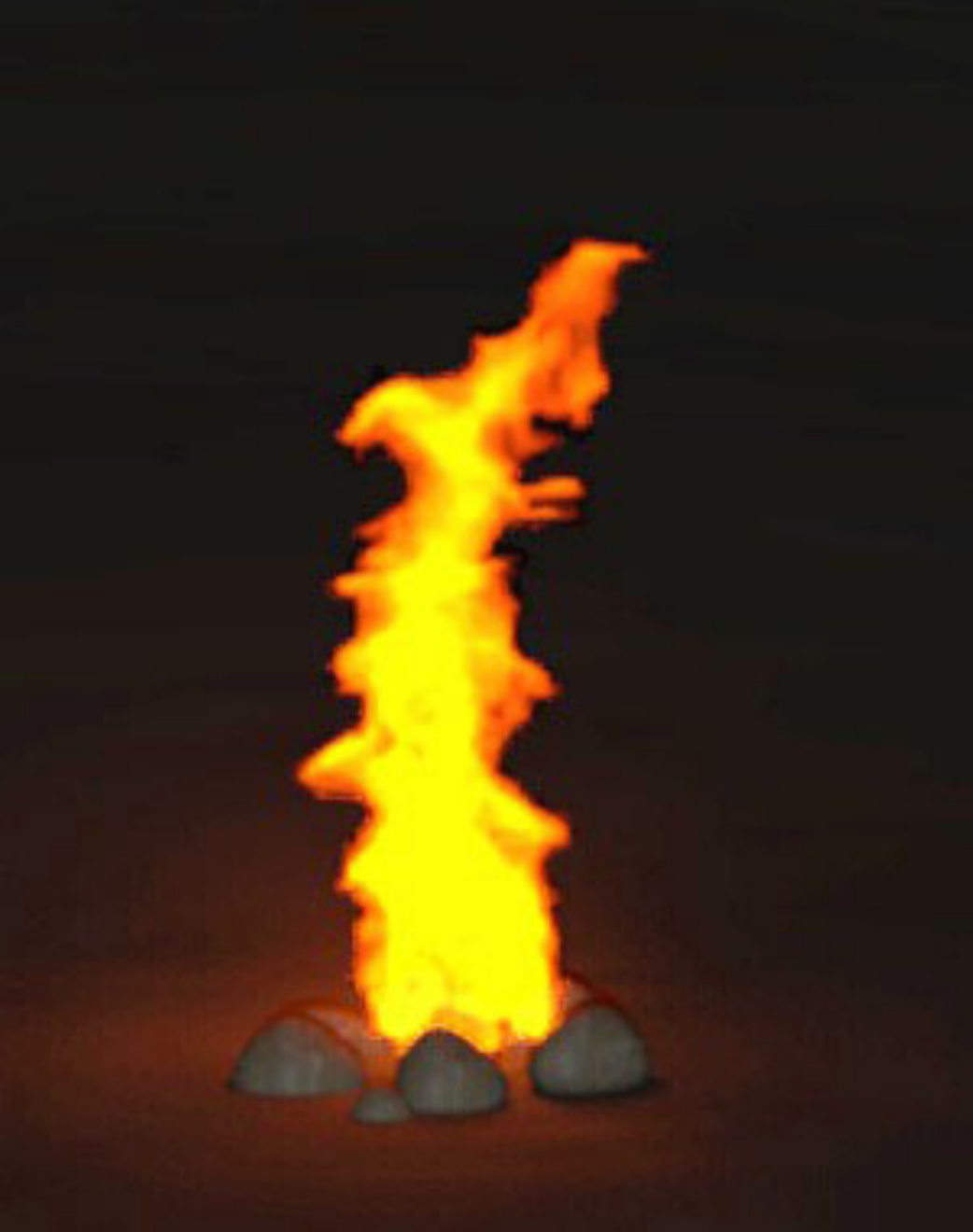}&
\includegraphics[width=0.225\textwidth]{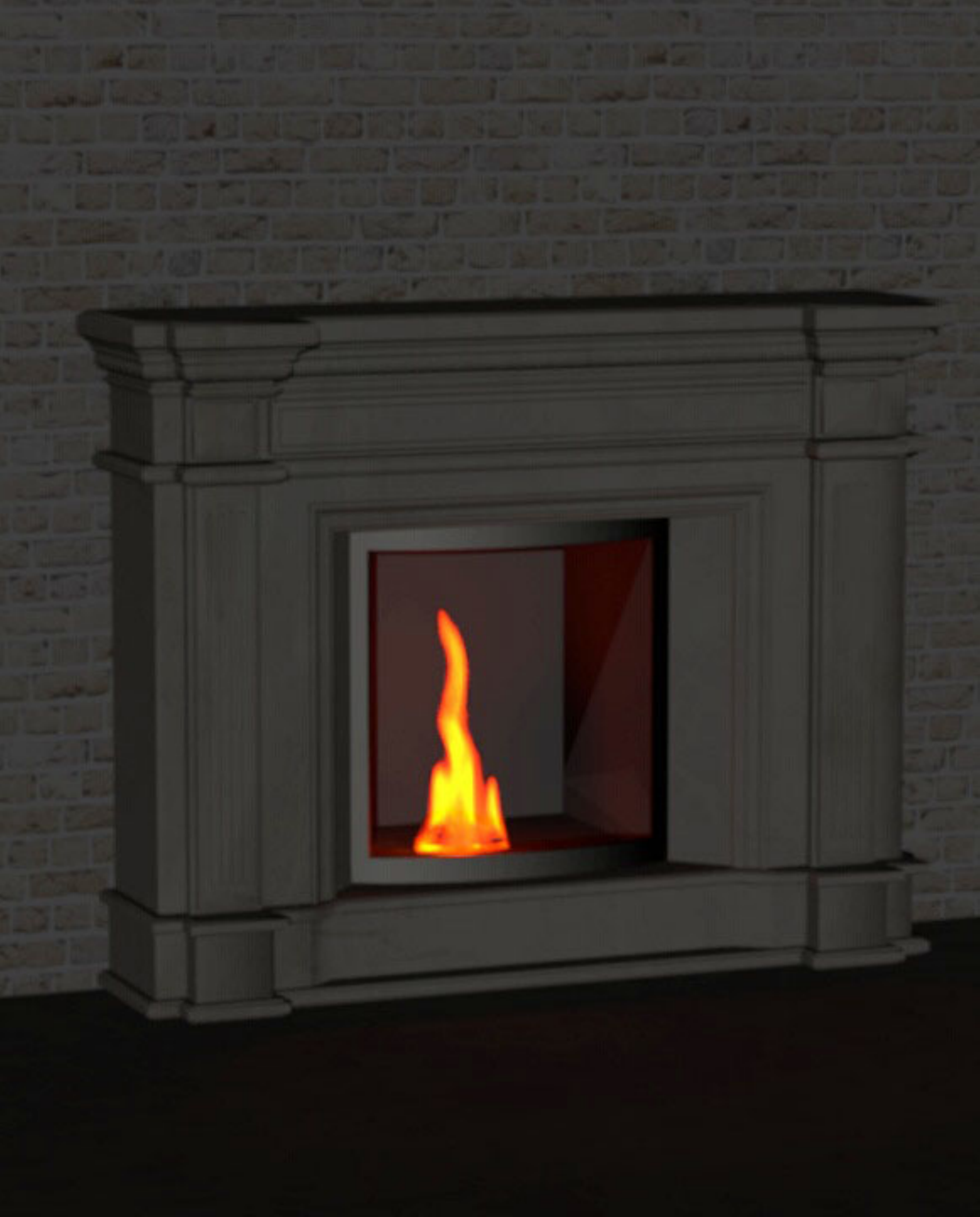}&
\multicolumn{2}{c}{\includegraphics[width=0.498\textwidth]{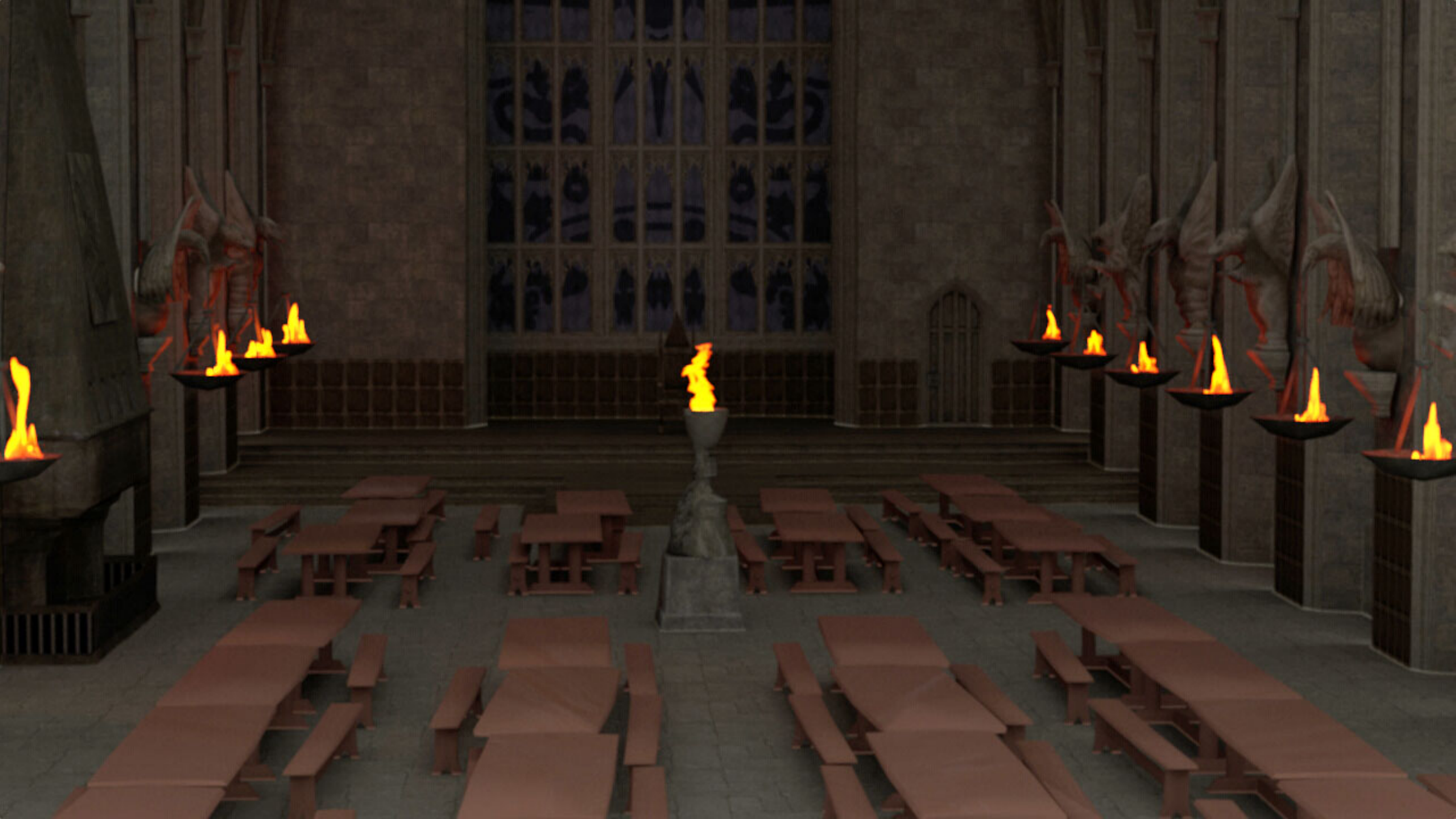}}\\
(a) & (b) & \multicolumn{2}{c}{(c)}\\
\end{tabular}
\end{center}
\vspace{-3mm}
\caption{Reconstructed flames in a virtual environment.
\label{fig_app_vir}}
\vspace{-2.1mm}
\end{figure*}

Given a virtual environment, reconstructed flames can be added into the scene using our method.
Fig. \ref{fig_app_vir}(a) shows a frame from a bonfire scene using reconstructed data from simulated flames.
Fig. \ref{fig_app_vir}(b) shows a frames from a fireplace scene using reconstructed data from a captured flame.
Multiple flames can be combined as a flame group and set into the same scene as shown in Fig. \ref{fig_app_vir}(c). A complete animation of the reconstructed flames in virtual environment settings can be seen in the accompanying supplementary video.

\begin{figure}[t]
\begin{center}
\begin{tabular}{cc}
\includegraphics[width=0.23\textwidth]{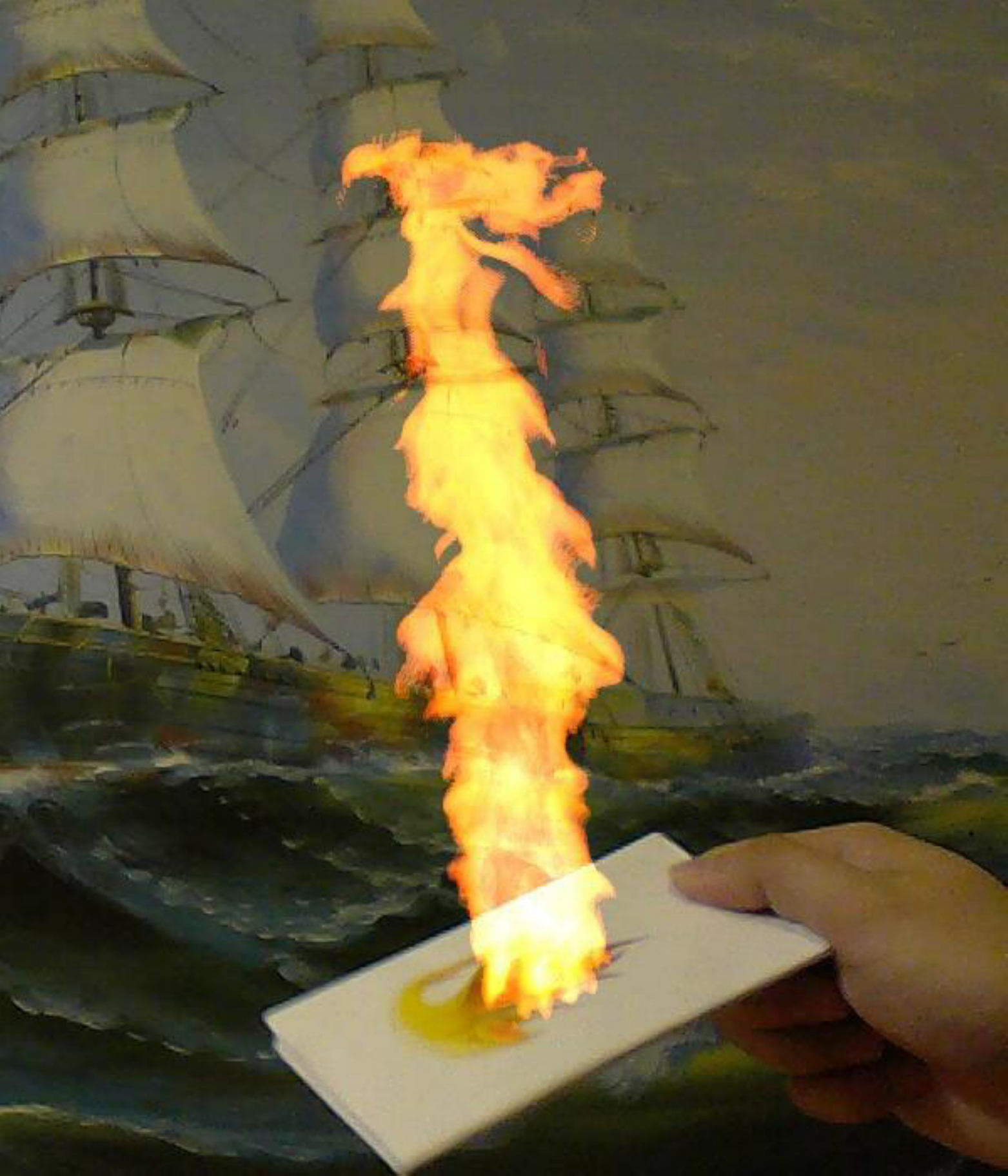}&
\includegraphics[width=0.23\textwidth]{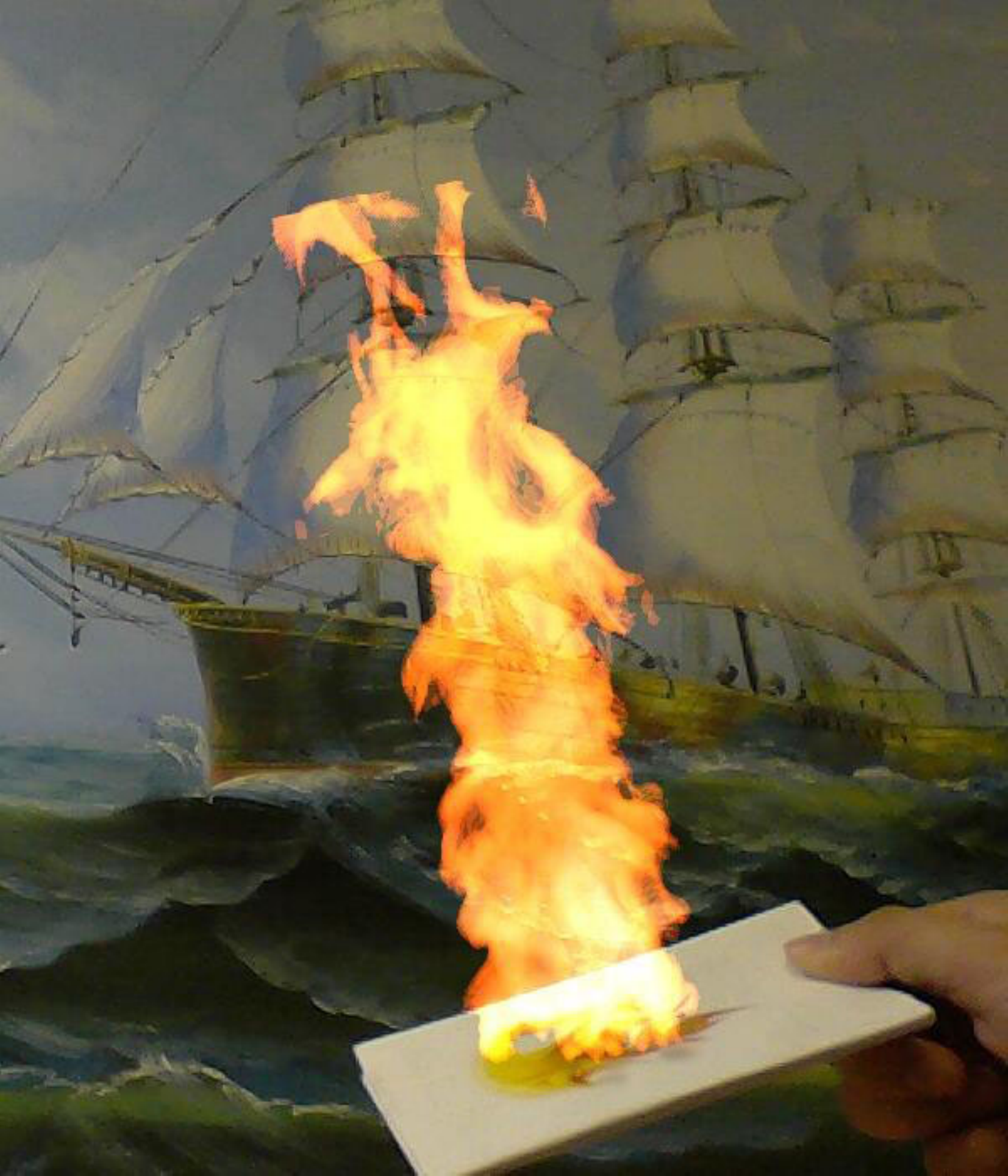}\\
\end{tabular}
\end{center}
\vspace{-3mm}
\caption{Reconstructed flames in augmented reality.
\label{fig_app_ar}}
\vspace{-3mm}
\end{figure}

\begin{figure}[t]
\begin{center}
\setlength{\tabcolsep}{3pt}
\begin{tabular}{ccccc}
\includegraphics[width=0.08\textwidth]{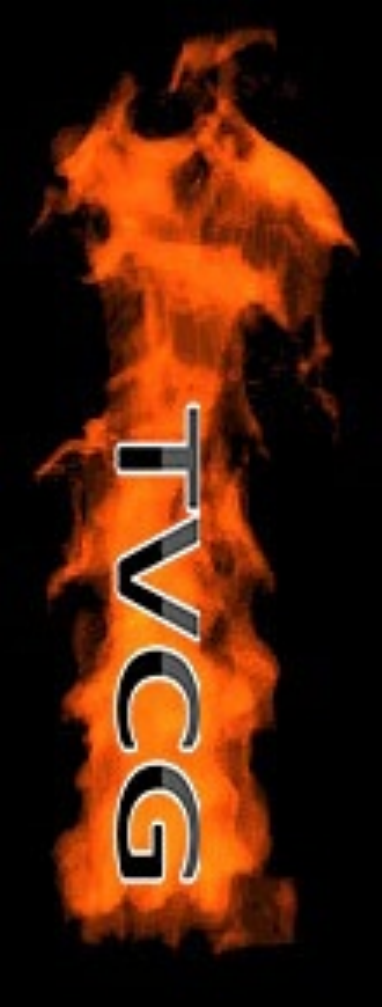}&
\includegraphics[width=0.08\textwidth]{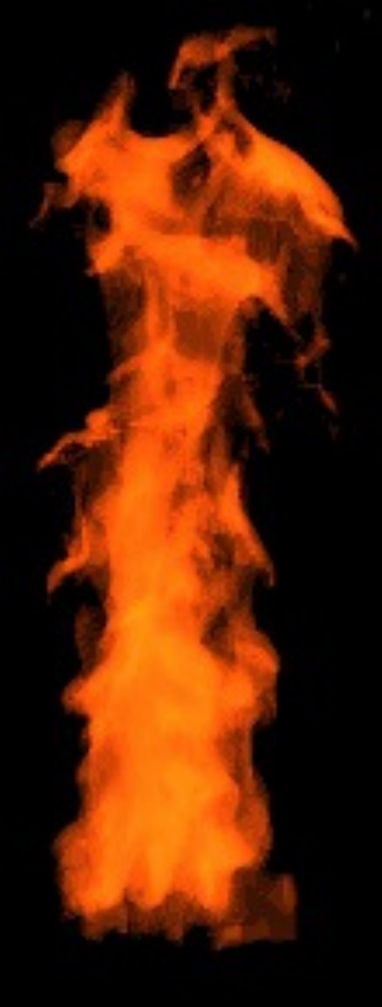}&
\includegraphics[width=0.08\textwidth]{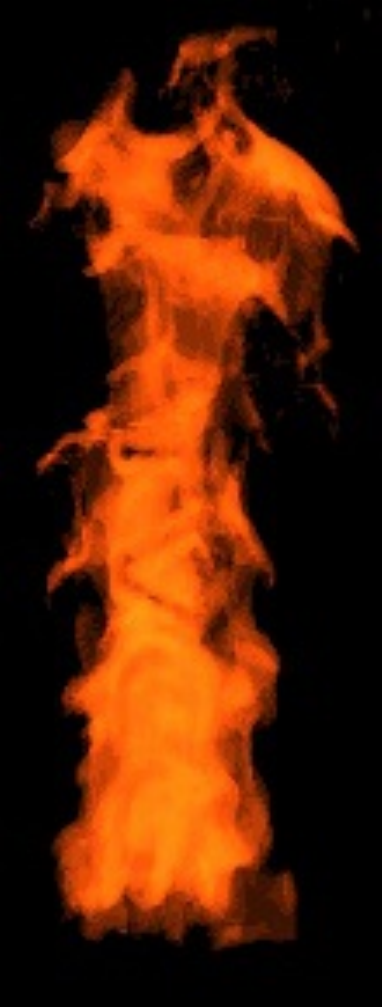}&
\includegraphics[width=0.08\textwidth]{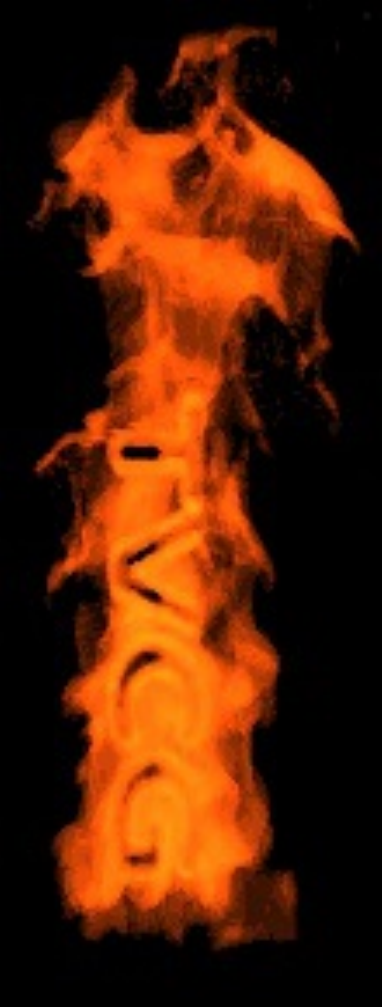}&
\includegraphics[width=0.08\textwidth]{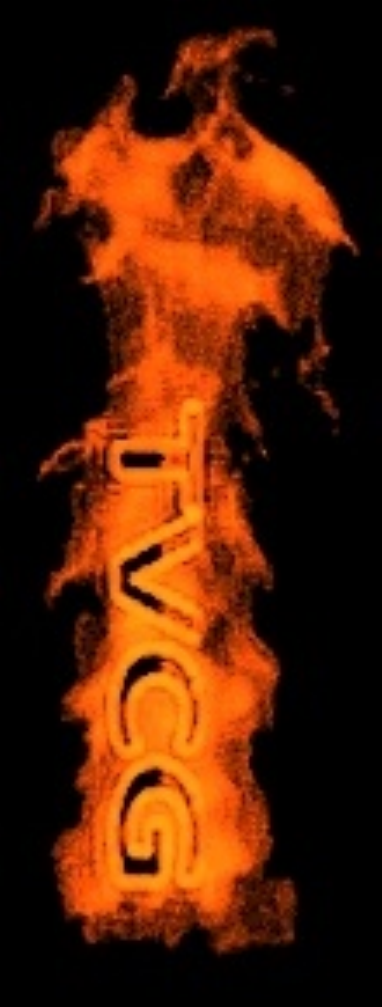}\\
(a) & (b) & (c) & (d) & (e)
\end{tabular}
\end{center}
\vspace{-3mm}
\caption{(a) Modified input image and (b)--(d) show the flame-stylization process.
\label{fig_app_edit}}
\vspace{-2mm}
\end{figure}

\begin{figure}[t]
\begin{center}
\setlength{\tabcolsep}{3pt}
\begin{tabular}{cc}
\includegraphics[width=0.193\textwidth]{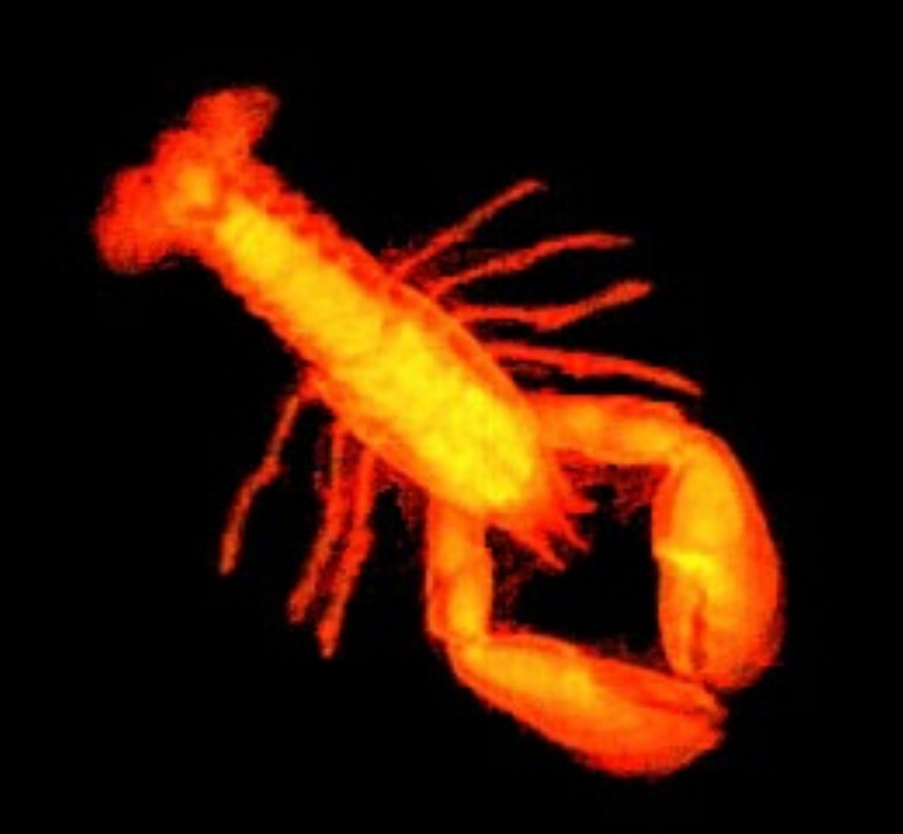}&
\includegraphics[width=0.27\textwidth]{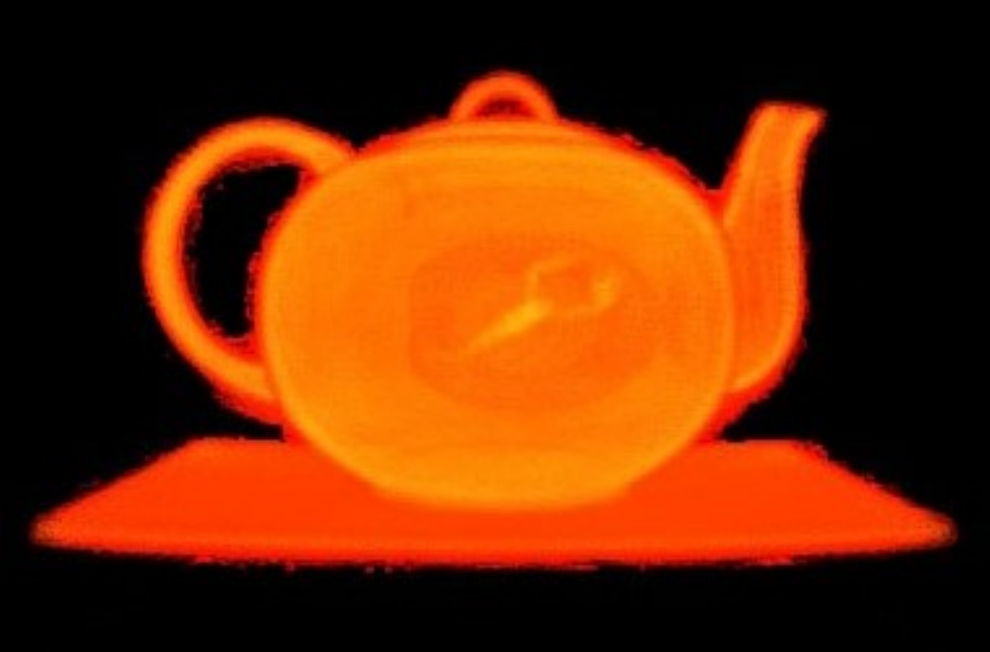}\\
\end{tabular}
\end{center}
\vspace{-3mm}
\caption{The flame stylization of a lobster (left), and the flame stylization of a teapot with a small lobster inside (right).
\label{fig_obj_sty}}
\vspace{-2mm}
\end{figure}

\subsection{Reconstructed flames in augmented reality}
Augmented reality (AR) platforms have become prevalent in the last few years for training and educational purposes. More recently, with the interest generated by games such as Pok\'{e}mon Go~\cite{pokemongo}, the future of AR games seems bright. To cater for AR games and displays in the future, we have also demonstrated our flame reconstruction results in augmented reality settings. Given an augmented reality setting, we can render our reconstructed flames in the scene, as shown in Fig. \ref{fig_app_ar}. A complete animation of the reconstructed flames in an AR setting can be seen in the accompanying supplementary video.

\subsection{Flame Stylization}
In some flame effects, special objects may gradually come out from the flames, which could be treated as a case of volume stylization \cite{klehm2014property}.
Color intensity reconstruction methods cannot achieve these effects since the objects will maintain their original color.
However, with the temperature reconstruction workflow of our method, we can easily capture the flame stylization effects.

For all pixels of the object image, we first set the green channel values equal to the gray intensity,
and then combine this ``green'' image with one of the input flame images.
Since our method runs iteratively and visualizes the results after every iteration,
the users can visualize the complete process of the object appearance in a flame style in the volume. Fig. \ref{fig_app_edit} shows some stylization results.
Since our method achieves real-time performance, the artists can observe the volume stylization effects immediately
after changing the appearance of the input images. Moreover, given images from different views of an object, the flame stylization effects would make the object seem to be overheated or burned, as shown in Fig. \ref{fig_obj_sty}.

\subsection{Reconstruction of Other Phenomena}
\begin{figure}[t]
\begin{center}
\setlength{\tabcolsep}{3pt}
\begin{tabular}{ccc}
\includegraphics[width=0.15\textwidth]{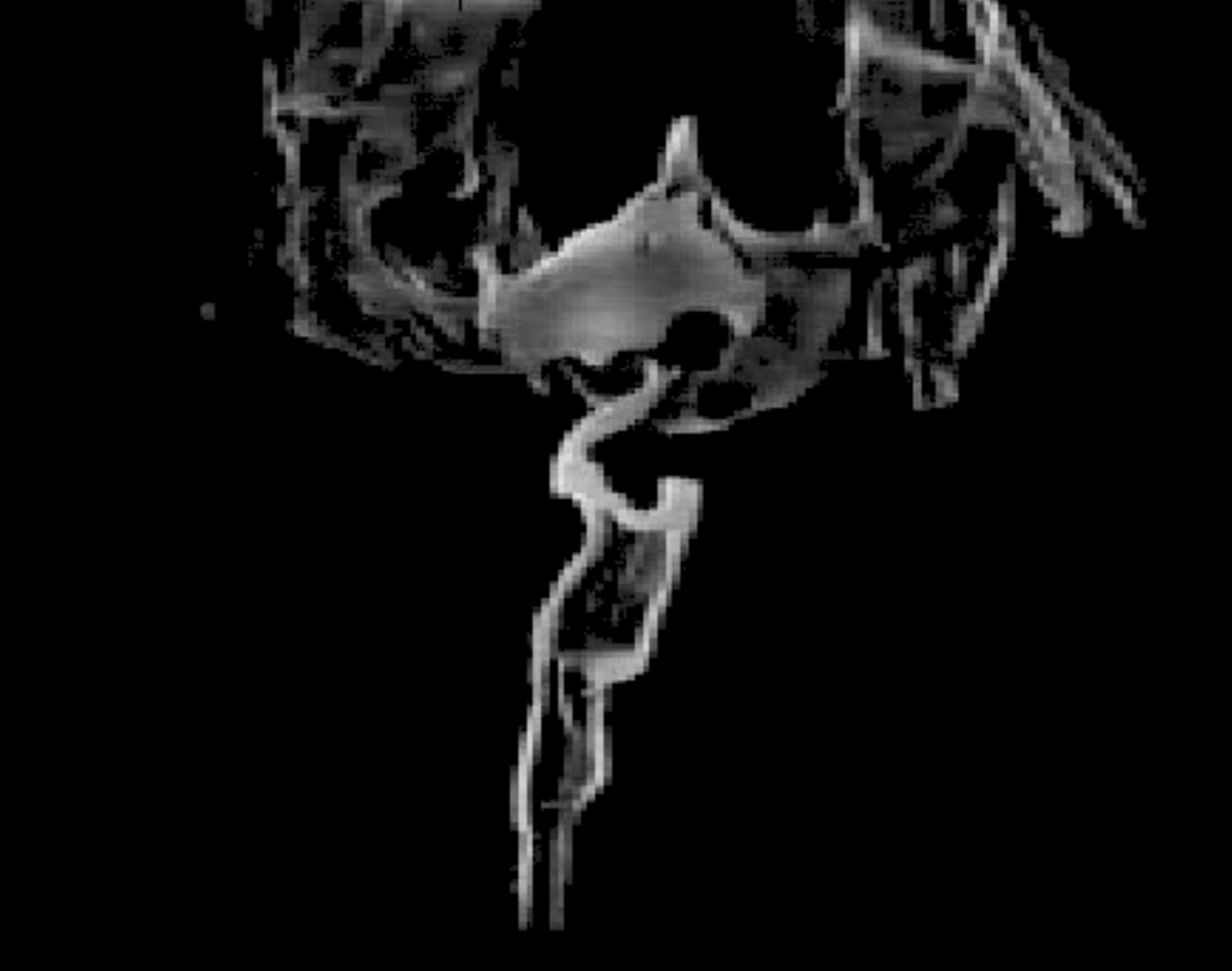}&
\includegraphics[width=0.15\textwidth]{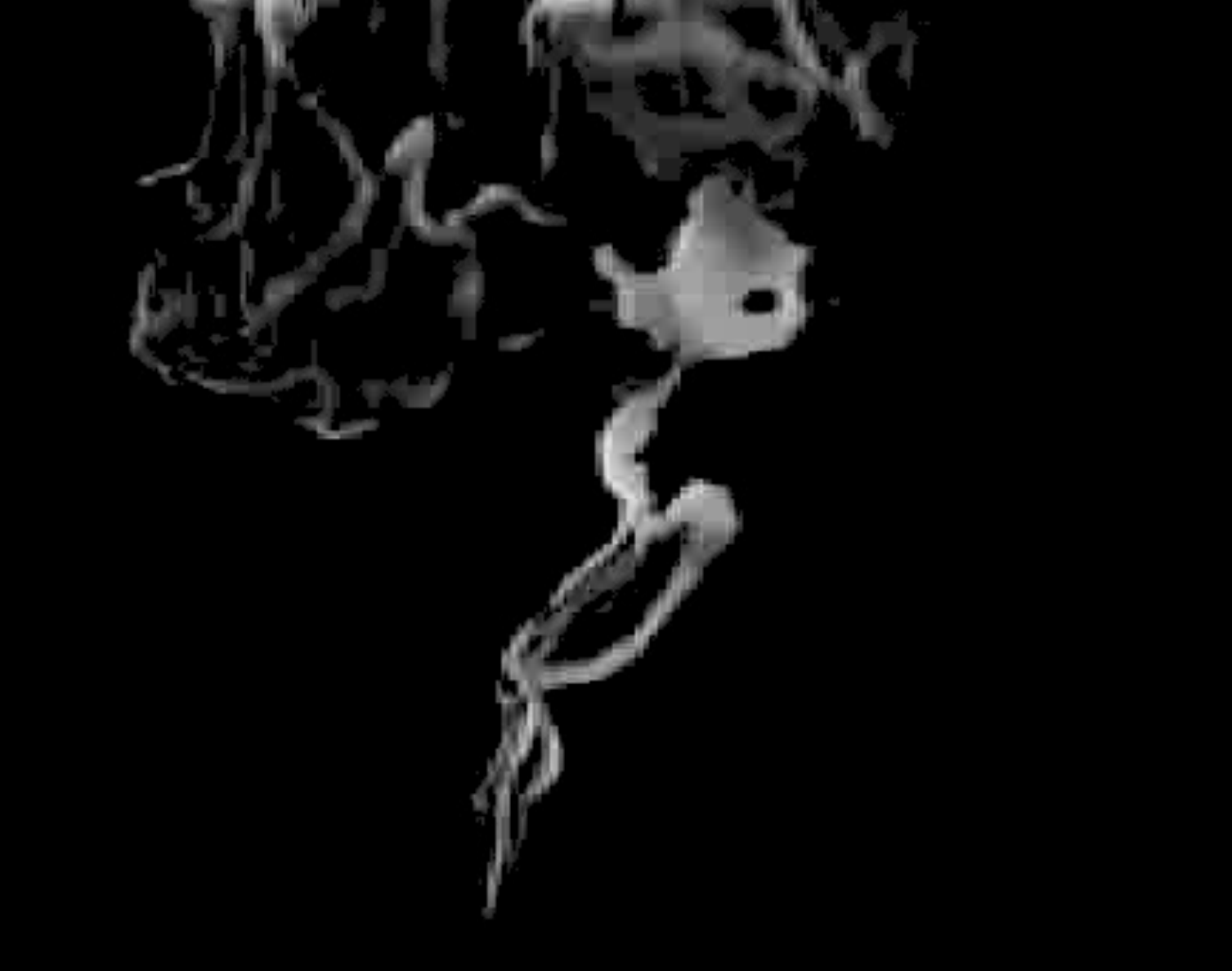}&
\includegraphics[width=0.15\textwidth]{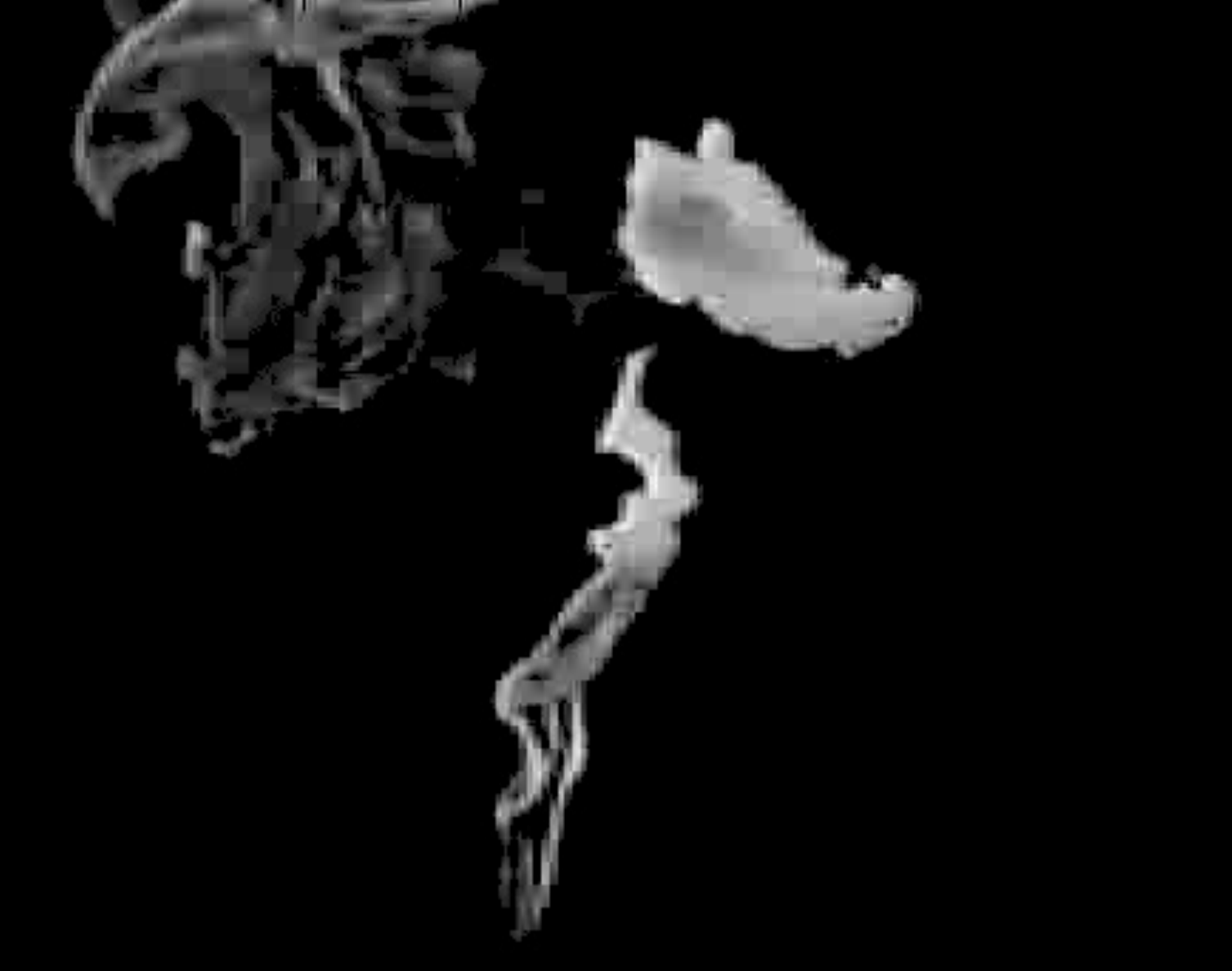}\\
(a) & (b) & (c)\\
\includegraphics[width=0.15\textwidth]{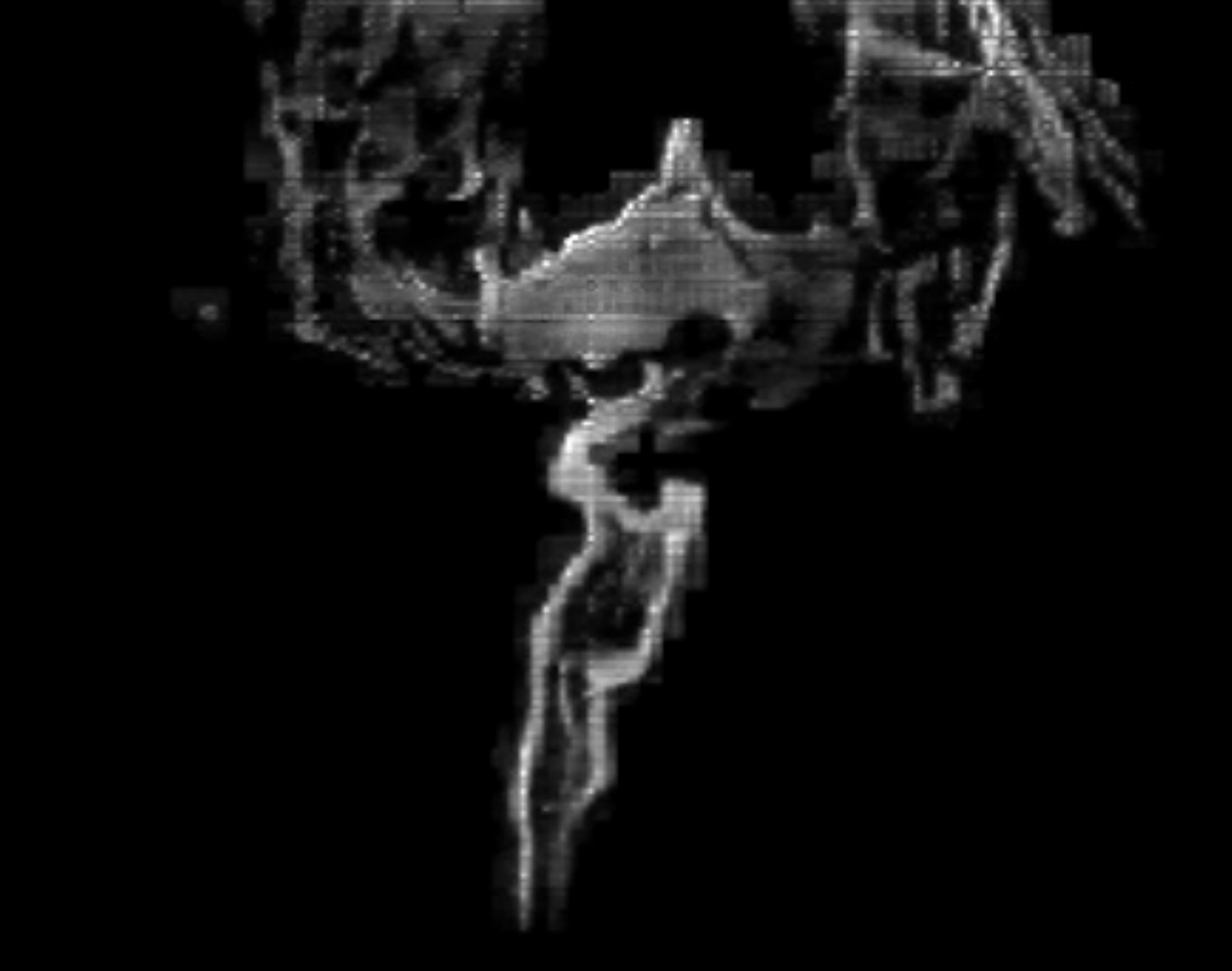}&
\includegraphics[width=0.15\textwidth]{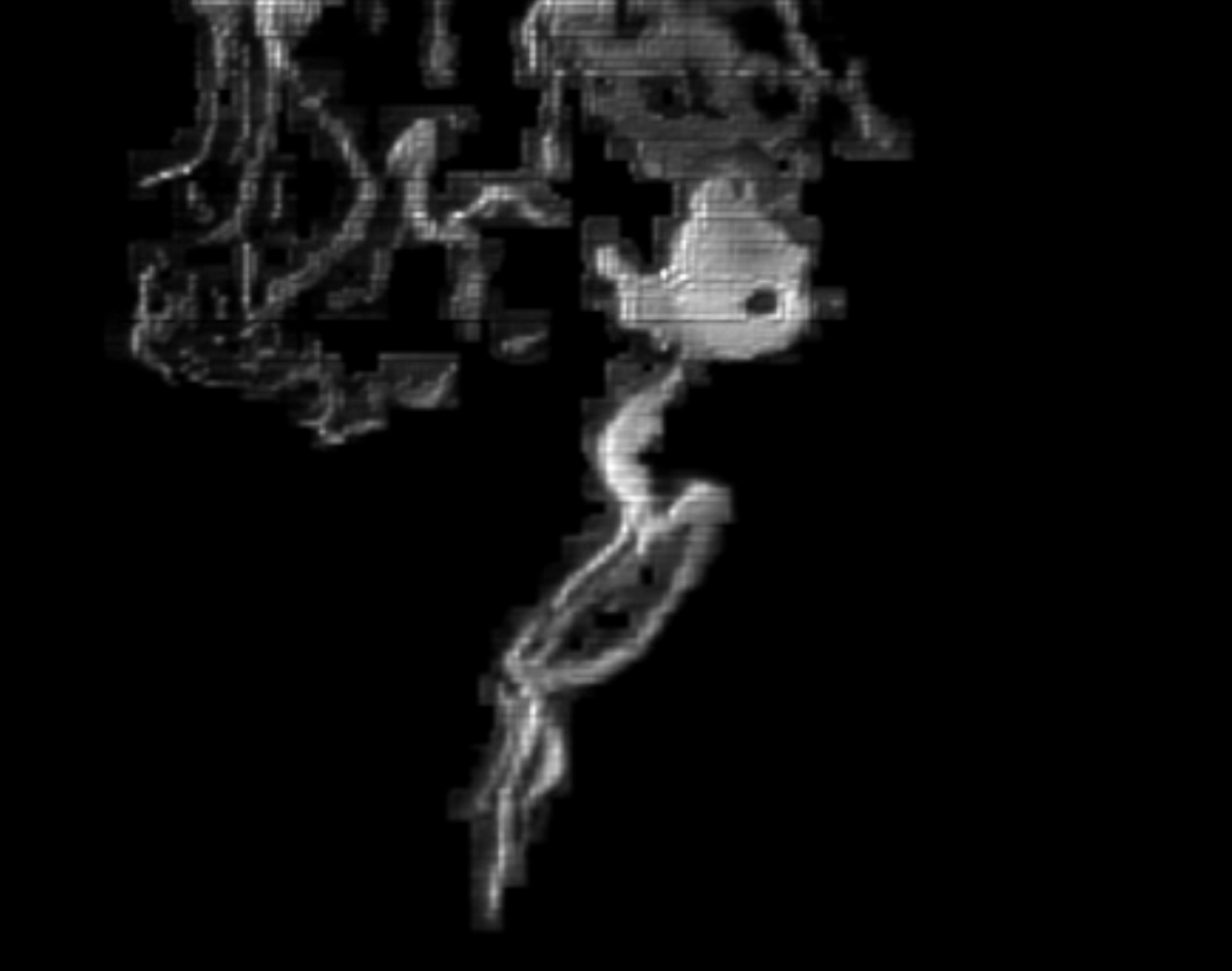}&
\includegraphics[width=0.15\textwidth]{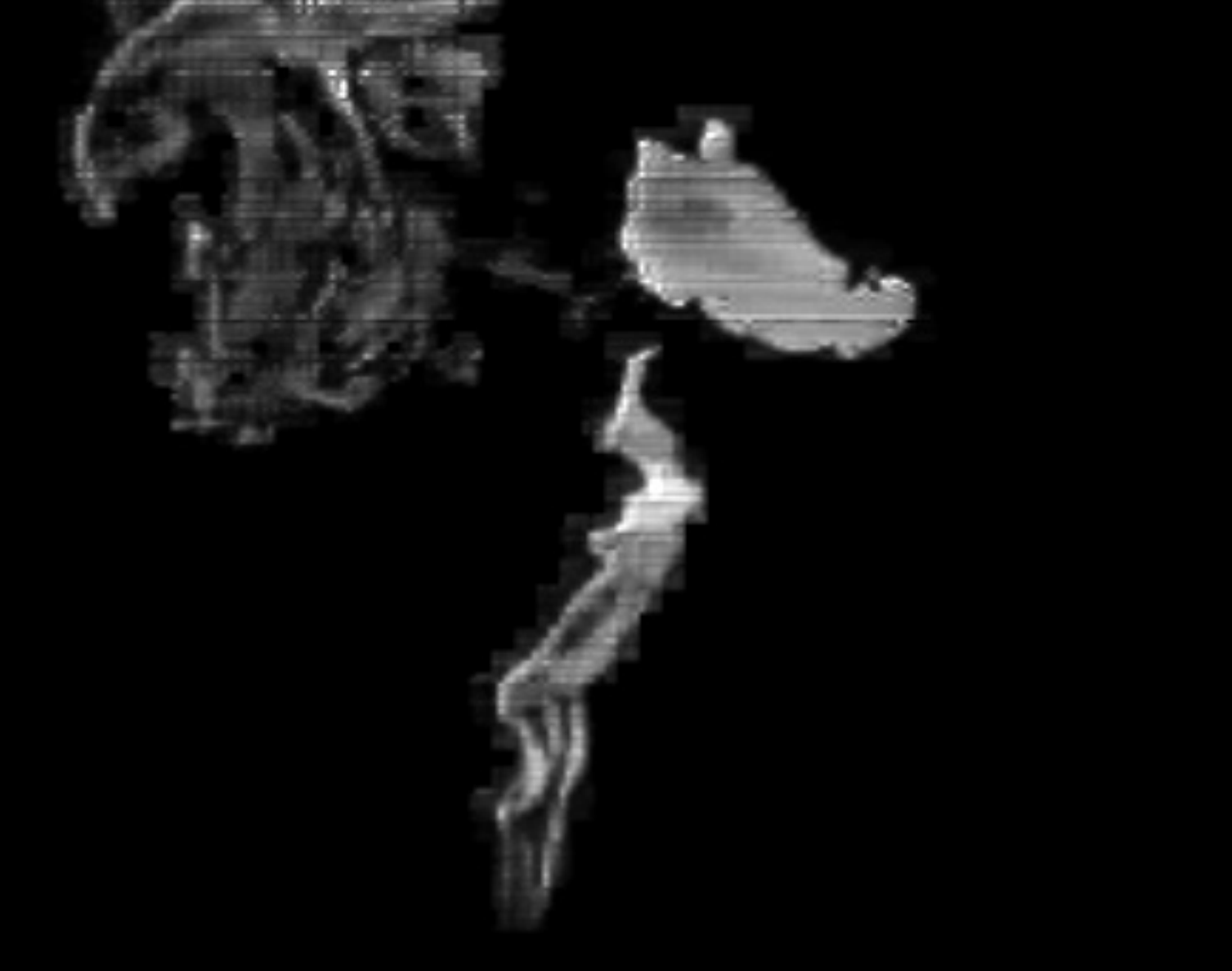}\\
(d) & (e) & (f)
\end{tabular}
\end{center}
\vspace{-3mm}
\caption{Smoke reconstruction input images (top row) and the corresponding reconstructed images (bottom row).
\label{fig_result_smoke}}
\end{figure}

Using the color intensity reconstruction workflow,
our approach can also be applied to reconstruct other semitransparent phenomena, such as smoke.
We use a smoke dataset with 47 images from different views, with 3 views as input.
The errors for the 3 input views and all the 47 views using our reconstruction approach and the sheet decomposition method \cite{hasinoff2007photo}
are shown in Fig.~\ref{fig_result_smoke} and Table~\ref{table_err_of_smoke}. 

\begin{table}[t]
\normalsize
\caption{RMSE of smoke reconstruction with 3 input views.}
\vspace{-1mm}
\label{table_err_of_smoke}
\centering
\begin{tabular}
{c|c|c}
\hline
\multirow{2}{*}{Reconstruction method} & \multicolumn{2}{c}{RMSE} \\
\cline{2-3} & Input & All \\
\hline
\hline
Flame sheet decomposition & 10.1 & 18.4 \\
\hline
Our method & 13.9 & 16.7 \\
\hline
\end{tabular}
\vspace{-2mm}
\end{table}


\section{Conclusion and Future Work}
\label{sec_conclusion}

In this paper, we have presented an algorithm for flame volume reconstruction.
We have devised a novel camera synchronization method to capture flame data using inexpensive consumer CCD cameras.
We also presented a novel reconstruction method that
enables complex flame rendering models to be used in the reconstruction process.
Moreover, we accelerated our reconstruction method using the GPU, which provides real-time performance for each iteration.
Finally, we evaluated our method on both simulated and real captured data
and demonstrated a variety of different applications for our approach.

There are two limitations of our current approach. The method currently can only deal with small-scale flames, and flames with complex internal structures cannot be accurately reconstructed. Since we currently deal with small scale flames, refraction is ignored when the rays travel through the volume and constant transparency for the volume data is used for simplicity. In the future, we will remove these constraints and create a more comprehensive picture of the flame volume data. We will also extend our approach to deal with reconstruction of large-scale fire and flames with complex internal structures.

In the case of temperature reconstruction, we assume the flame particles act as an ideal black-body,
and little is considered in terms of the effects of different chemicals on rendering flame images,
such as pink flames affected by lithium ion and green flames affected by copper ion.
Thus, the temperature reconstruction cannot be applied on all kinds of flames.
In the future, we will deal with this issue using a physically-based fire rendering approach \cite{pegoraro2006physically}.

We will also work on creating a user interface for artists and designers to edit special flame effects after every iteration of the reconstruction process.
Since our approach can achieve real-time performance, allows visualization after every iteration, and provides visually plausible results,
the artists/designers can immediately receive feedback after editing the input images and adjust the results accordingly.

\ifCLASSOPTIONcompsoc
  \section*{Acknowledgments}
\else
  \section*{Acknowledgment}
\fi

This research is supported by National High Technology Research and Development Program of China (863 Program) (Grant No.
2015AA016401), National Natural Science Foundation of China under Grant Nos. 61379085, 61173067 and 61532002, and US National Science Foundation grants CNS0959979, IIP1069147, CNS1302246, NRT1633299, and CNS1650499. The authors would like to thank the team of Prof. W. Wu at Beihang University for providing their code and helping us evaluate our method. The bonsai, teapot, and lobster volumetric datasets are courtesy IAPR-TC18 Digital Geometry repository.


\vspace{-12mm}
\begin{IEEEbiography}[{\includegraphics[width=1in,height=1.25in,clip,keepaspectratio]{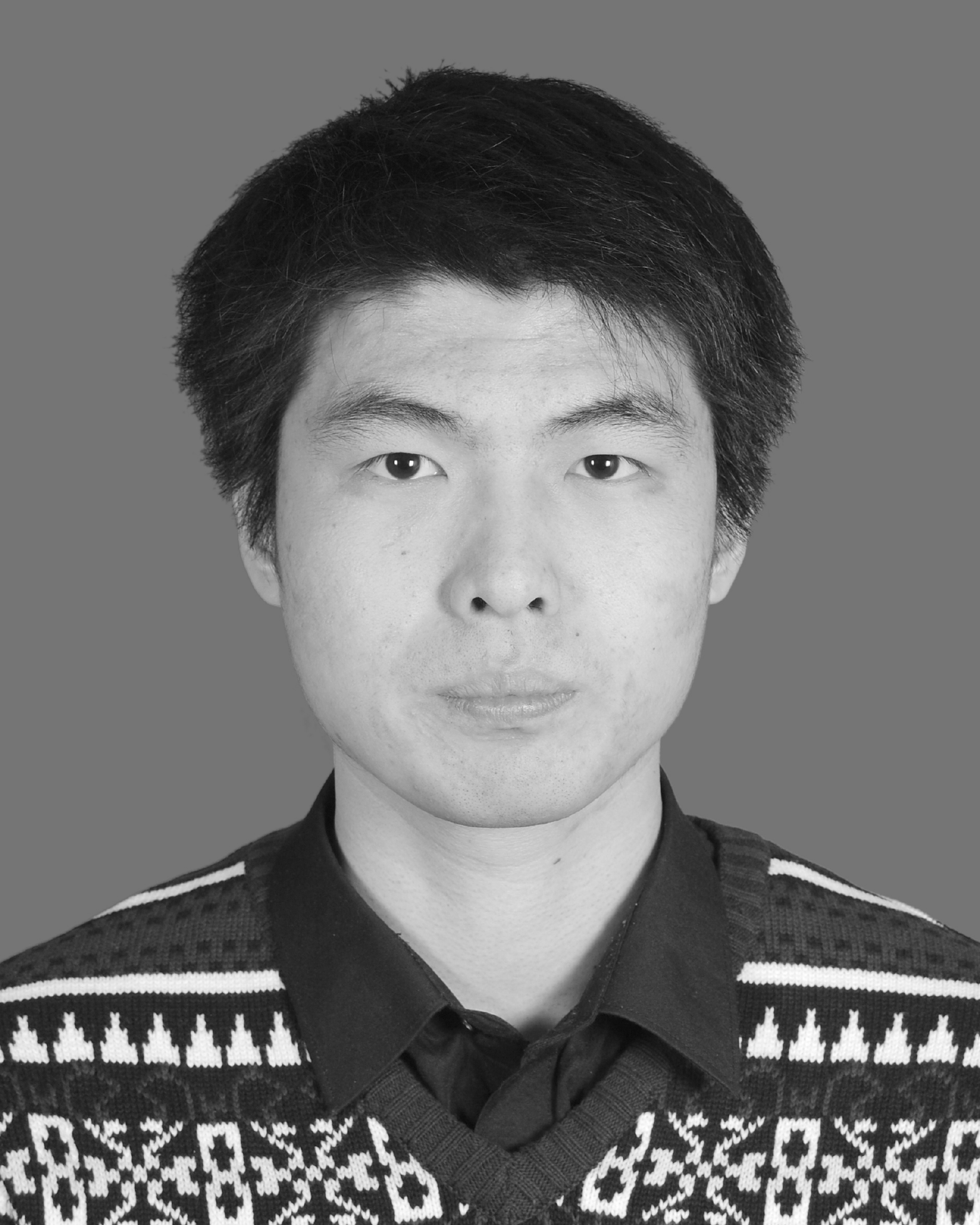}}]{Liang Shen}
is a PhD candidate at the Institute of Computing Technology, Chinese Academy of Sciences. He did his BSc in Computer Science from China University of Geosciences, Wuhan, China.
His main research interests are in fluid reconstruction and computer vision.
\end{IEEEbiography}
\vspace{-12mm}

\begin{IEEEbiography}[{\includegraphics[width=1in,height=1.25in,clip,keepaspectratio]{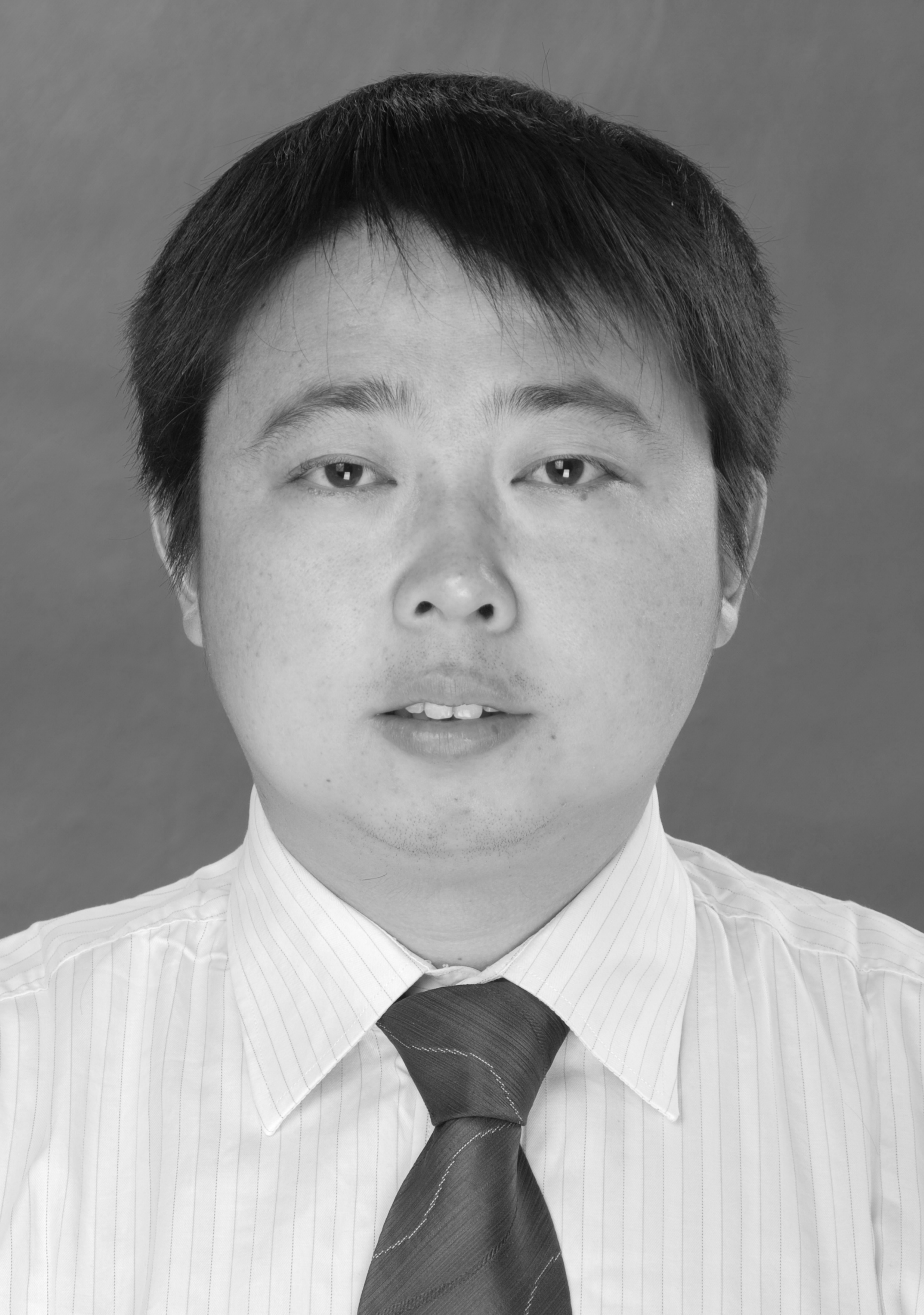}}]{Dengming Zhu}
received his PhD degree from the Institute of Computing Technology, Chinese Academy of Sciences in 2009,
where he is also working as an Associate Professor.
His main scientific interests are in the areas of fluid simulation, parallel visualization, and virtual reality.
\end{IEEEbiography}
\vspace{-12mm}

\begin{IEEEbiography}[{\includegraphics[width=1in,height=1.25in,clip,keepaspectratio]{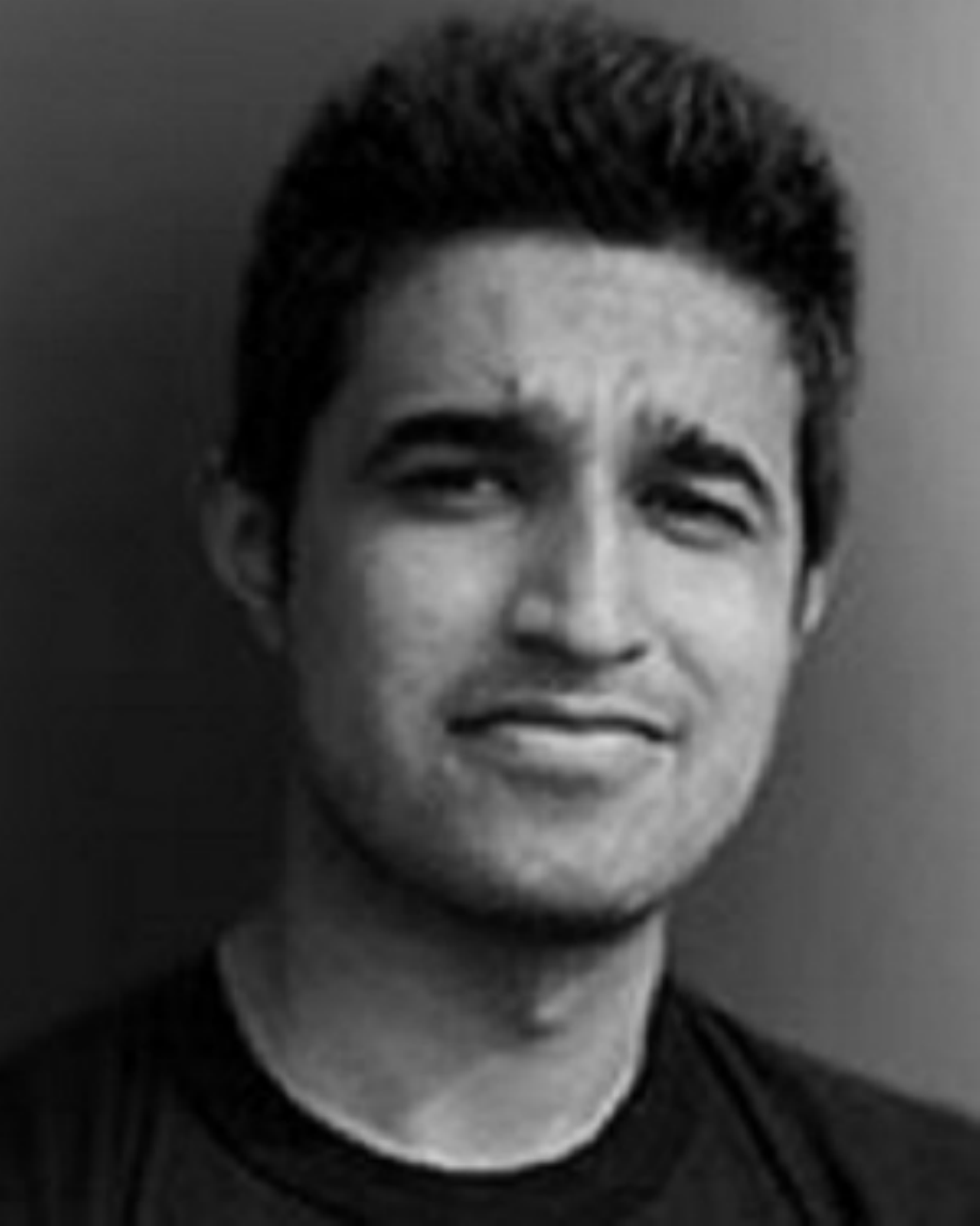}}]{Saad Nadeem}
is a PhD candidate in the Computer Science department, Stony Brook University. He did his BSc Honors in Computer Science and Mathematics from School of Science and Engineering, Lahore University of Management Sciences, Pakistan. His research interests include computer vision, computer graphics, and visualization.
\end{IEEEbiography}
\vspace{-12mm}

\begin{IEEEbiography}[{\includegraphics[width=1in,height=1.25in,clip,keepaspectratio]{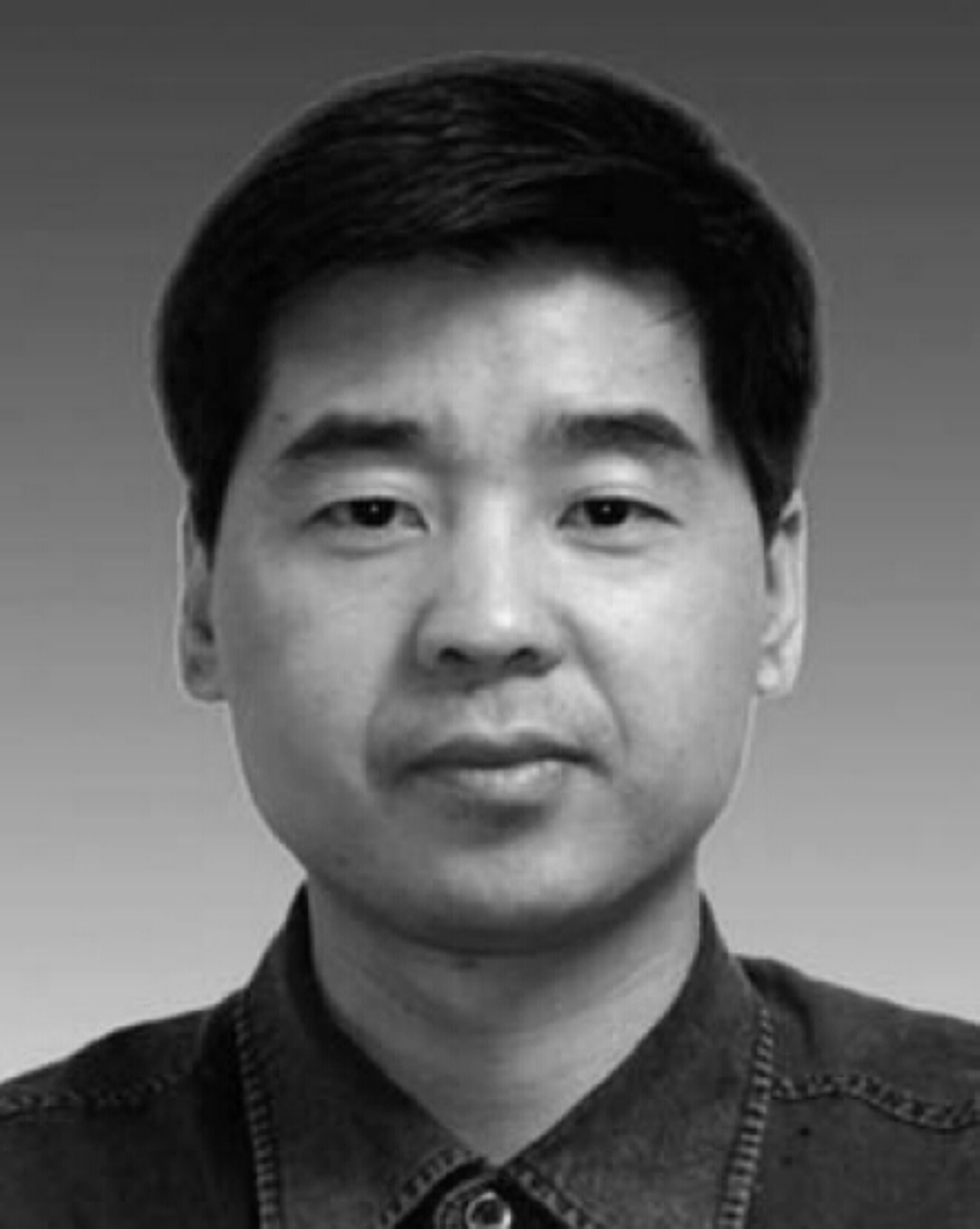}}]{Zhaoqi Wang}
is a Researcher and a Director of PhD students at the Institute of Computing Technology, Chinese Academy of Sciences.
His research interests include virtual reality and intelligent human computer interaction.
He is a senior member of the China Computer Federation.
\end{IEEEbiography}
\vspace{-12mm}

\begin{IEEEbiography}[{\includegraphics[width=1in,height=1.25in,clip,keepaspectratio]{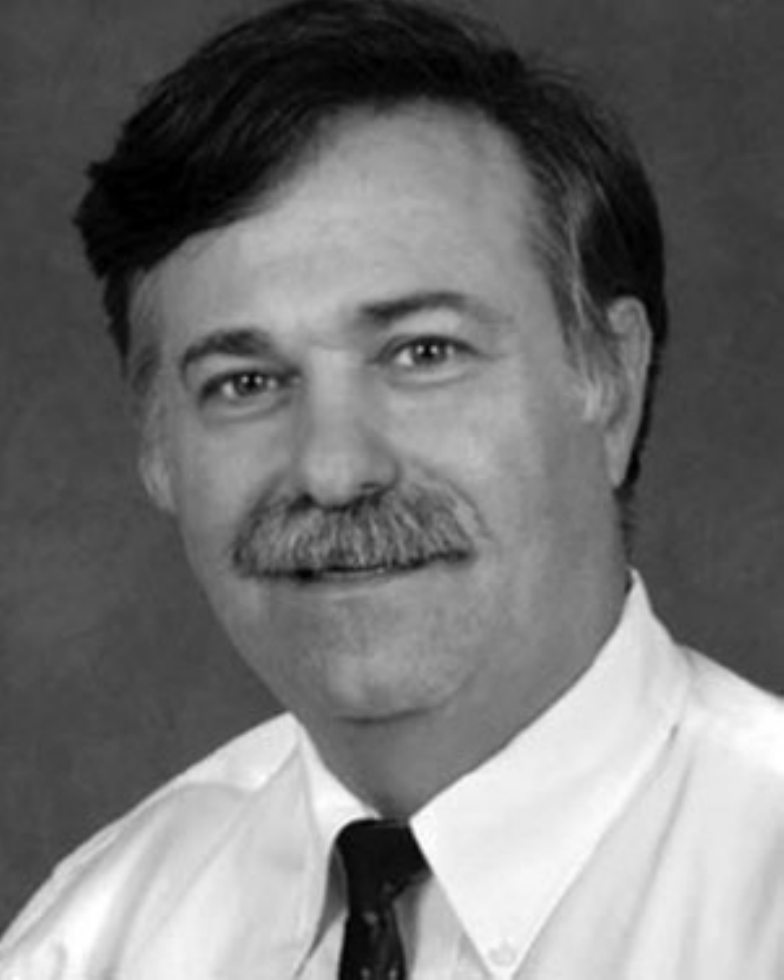}}]{Arie E. Kaufman}
is a Distinguished Professor and Chairman of the Computer Science Department, Director of the Center of Visual Computing (CVC), and Chief Scientist of the Center of Excellence in Wireless and Information Technology (CEWIT) at Stony Brook University. He has conducted research for over 40 years in visualization and graphics and their applications, has published more than 300 refereed papers, has presented more than 20 invited keynote talks, has been awarded/filed more than 40 patents, and has been PI/co-PI on more than 100 grants. He was the founding Editor-in-Chief of IEEE Transaction on Visualization and Computer Graphics (TVCG), 1995-1998. He is a Fellow of IEEE, a Fellow of ACM, the recipient of the IEEE Visualization Career Award (2005), and was inducted into the Long Island Technology Hall of Fame (2013). He received his PhD in Computer Science from Ben-Gurion University, Israel, in 1977.
\end{IEEEbiography}

\end{document}